\documentclass{article} 
\usepackage{iclr2023_conference,times}

\usepackage{amsmath,amsfonts,bm}

\def\eqref#1{equation~\ref{#1}}

\def\1{\bm{1}}

\DeclareMathAlphabet{\mathsfit}{\encodingdefault}{\sfdefault}{m}{sl}
\SetMathAlphabet{\mathsfit}{bold}{\encodingdefault}{\sfdefault}{bx}{n}

\newcommand{\E}{\mathbb{E}}

\usepackage{xurl}

\usepackage{lipsum}
\usepackage{listings}
\usepackage{extarrows}
\usepackage{subfigure}
\usepackage{xspace}
\usepackage{graphicx}
\usepackage{makecell}
\usepackage{ifthen}
\usepackage{xifthen}
\usepackage{wrapfig}
\usepackage{amsthm}
\usepackage{stmaryrd}
\usepackage{threeparttable}
\usepackage{enumerate}
\usepackage{multirow}
\usepackage{booktabs}
\usepackage{titlesec}
\usepackage{mathtools, nccmath}
\usepackage{comment}
\usepackage{amsmath,amsfonts,amssymb}
\usepackage[framemethod=TikZ]{mdframed}
\usepackage{lipsum}
\usepackage{float}
\usepackage{threeparttable}
\usepackage{caption}
\usepackage{tcolorbox}
\usepackage{listings}
\usepackage{upquote}
\usepackage[T1]{fontenc}
\usepackage{xcolor}
\usepackage[hidelinks,colorlinks=true,linkcolor=blue,citecolor=blue]{hyperref}
\usepackage{blindtext}

\usepackage{tcolorbox}
\tcbuselibrary{skins}
\usepackage{enumerate}
\usepackage{enumitem}

\usepackage{minitoc}

\usepackage[normalem]{ulem}
\useunder{\uline}{\ul}{}

\title{\icon GLM-130B: An Open Bilingual Pre-Trained Model}

\author{\\
\bf Aohan Zeng$^{\diamond \dagger*}$, Xiao Liu$^{\diamond \dagger*}$, Zhengxiao Du$^{\diamond\dagger}$, Zihan Wang$^{\diamond}$, Hanyu Lai$^\diamond$, Ming Ding$^\diamond$,\\
\bf Zhuoyi Yang$^\diamond$, Yifan Xu$^\diamond$, Wendi Zheng$^\diamond$, Xiao Xia$^\diamond$, Weng Lam Tam$^{\diamond\mathsection}$, Zixuan Ma$^\diamond$, \\
\bf Yufei Xue$^\mathsection$, Jidong Zhai$^\diamond$, Wenguang Chen$^\diamond$, Peng Zhang$^\mathsection$, Yuxiao Dong$^{\diamond\ddagger}$, Jie Tang$^{\diamond\ddagger}$ \\\\
\centerline{Tsinghua University$^\diamond$ \qquad Zhipu.AI$^\mathsection$\qquad}
}

\newcommand{\glm}[0]{GLM-130B\xspace}
\newcommand\icon{\raisebox{-3.7pt}{\includegraphics[width=1.5em]{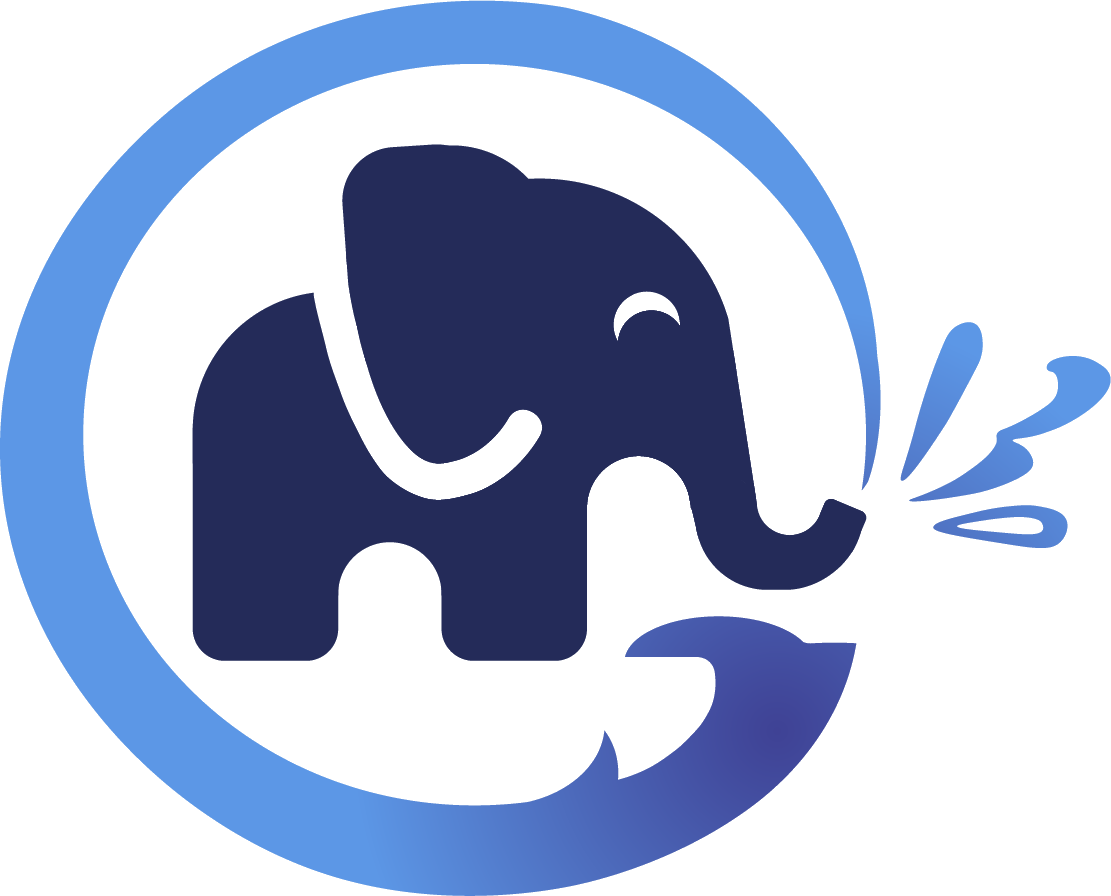}}}

\newcommand{\hide}[1]{}

\newmdtheoremenv[%
backgroundcolor=yellow!10,%
outerlinecolor=black,%
innertopmargin = \topskip,%
splittopskip = \topskip,%
ntheorem = false,%
roundcorner=4pt]
{insight}{Lesson}

\newcommand\mybar{\kern1pt\rule[-\dp\strutbox]{.8pt}{\baselineskip}\kern1pt}

\newcommand\blfootnote[1]{%
  \begingroup
  \renewcommand\thefootnote{}\footnote{#1}%
  \addtocounter{footnote}{-1}%
  \endgroup
}

\newcommand{\vvpara}[1]{\noindent\textbf{#1}\xspace} %
\newcommand{\todo}[1]{\textbf{\color{red}[(TODO: #1 )]}}
\newcommand{\aohan}[1]{\textbf{\color{blue}[(aohan: #1 )]}}
\newcommand{\yuxiao}[1]{{\color{purple}[(Yuxiao: #1 )]}}

\newcommand{\du}[1]{{\color{brown}[(Zhengxiao: #1)]}}

\iclrfinalcopy 
\begin{document}

\doparttoc
\faketableofcontents

\maketitle


\renewcommand{\thefootnote}{\fnsymbol{footnote}}
    \footnotetext[1]{The two lead authors AZ and XL contributed equally (\texttt{\{zengaohan,shawliu9\}@gmail.com})}
    \footnotetext[2]{Work partially done when AZ, XL, and ZD interned at Zhipu.AI.}
    \footnotetext[3]{Team leads: YD and JT. Corresponding author: JT (\texttt{jietang@tsinghua.edu.cn})}
\renewcommand{\thefootnote}{\arabic{footnote}}

\begin{abstract}

We introduce \glm, a bilingual (English and Chinese) pre-trained language model with 130 billion parameters. 
It is an attempt to open-source a 100B-scale model at least as good as 
GPT-3 (davinci) and unveil how models of such a scale can be successfully pre-trained. 
Over the course of this effort, we face numerous unexpected technical and engineering challenges, particularly on loss spikes and divergence. 
In this paper, we introduce the training process of \glm including its design choices, training strategies for both efficiency and stability, and engineering efforts. 
The resultant \glm model offers significant outperformance over GPT-3 175B (davinci) on a wide range of popular English benchmarks while the performance advantage is not observed in OPT-175B and BLOOM-176B. 
It also consistently and significantly outperforms ERNIE TITAN 3.0 260B---the largest Chinese language model---across related benchmarks. 
Finally, we leverage a unique scaling property of \glm to reach INT4 quantization without post training, with almost no performance loss, making it the first among 100B-scale models and more importantly, allowing its effective inference on 4$\times$RTX 3090 (24G) or 8$\times$RTX 2080 Ti (11G) GPUs, the most affordable GPUs required for using 100B-scale models. 
The \glm model weights are publicly accessible and its code, training logs, related toolkit, and lessons learned are open-sourced at \url{https://github.com/THUDM/GLM-130B/}.
\blfootnote{For detailed author contributions, please refer to Appendix~\ref{app:contribution}.}
\end{abstract}

\hide{

\vspace{-8mm}
\begin{abstract}
We introduce \glm, an open and inclusive 130-billion parameters bilingual large language model (LLM).
As an initial attempt from academia, we commit to promoting ``LLM Inclusivity'', which aims at offering highly accessible, usable, and transparent LLMs to all researchers and 
individual developers.
We share a series of 
real lessons on architecture design, training strategies choice, and inference algorithm from our 8-month trajectory for developing an LLM.
All these insights result in \glm, which achieves 
better 
performance than other LLMs including GPT-3 175B, OPT 175B, BLOOM 176B, and ERNIE Titan 3.0 on a number of English and Chinese benchmarks.
In addition to its quality, GLM's special architecture and the scaling makes \glm the first LLM to be quantized into INT4 precision, which allows it to perform inference on popularized GPUs such as 4 $\times$ RTX 3090 (24G) or 8 $\times$ RTX 2080 Ti (11G).
Together with our speeding up efforts, \glm can be up to 8.4$\times$ faster than BLOOM 176B's official implementation, making \glm a real inclusive LLM for everyone.
The model weights and code are open-sourced at \url{https://anonymous.4open.science/r/GLM-130B/}.
\end{abstract}

}

\section{Introduction}
Large language models (LLMs), particularly those with over 100 billion (100B) parameters~\citep{brown2020language,thoppilan2022lamda,rae2021scaling,chowdhery2022palm,wang2021ernie}, have presented attractive 
scaling laws~\citep{wei2022emergent}, where emergent zero-shot and few-shot capabilities suddenly arose.
Among them, GPT-3~\citep{brown2020language} with 175B parameters pioneers the study of 100B-scale LLMs by strikingly generating better performance with 32 labeled examples than the fully-supervised BERT-Large model on a variety of benchmarks. 
However, both GPT-3 (and many other closed-sourced 100B-scale ones)---the model itself---and how it can be trained, have been thus far intransparent to the public. 
It is of critical value to train a high-quality LLM of such scale with both the model and training process shared with everyone.  

We thus \textit{aim to pre-train an open and highly-accurate 100B-scale model} with ethical concerns in mind. 
Over the course of our attempt, we have come to realize that pre-training a dense LLM at such a scale raises numerous unexpected technical and engineering challenges compared to training 10B-scale models, in terms of pre-training efficiency, stability, and convergence.  
Similar difficulties have also been concurrently observed in training OPT-175B~\citep{zhang2022opt} and BLOOM-176B~\citep{scao2022what}, further demonstrating the significance of GPT-3 as a pioneer study.

\begin{figure}[t]
    \centering
    \vspace{-4mm}
    \includegraphics[width=0.98\linewidth]{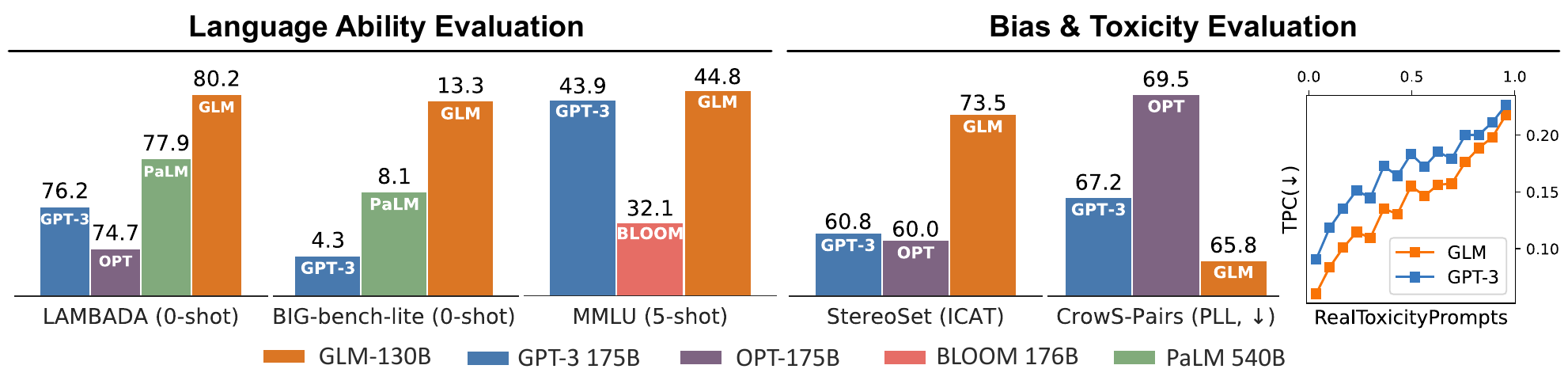}
    \vspace{-4mm}
    \caption{A summary of the performance evaluation 
    and ethical studies.}
    \label{fig:intro}
    \vspace{-4mm}
\end{figure}

\begin{table}[t]
\centering
\footnotesize
\renewcommand\tabcolsep{2.5pt}
\renewcommand\arraystretch{0.9}

\caption{A comparison between \glm and other 100B-scale LLMs and PaLM 540B. 
(LN: layer norm.; 
FPF: floating-point format; 
MIP: multi-task instruction pre-training; 
CN : Chinese)}
\vspace{-4mm}
\scalebox{0.9}{
\begin{tabular}{@{}lcccccccc@{}}
\toprule[1.2pt]
                        &                                                                          & \multicolumn{3}{c}{Architecture \& Data}                                                                                                                                                                               & \multicolumn{2}{c}{Training}                                               & \multicolumn{2}{c}{Inference}                                                                                                                                \\ \cmidrule(l){3-5} \cmidrule(l){6-7} \cmidrule(l){8-9} 
\multirow{-2}{*}{Model} & \multirow{-2}{*}{\begin{tabular}[c]{@{}c@{}}Open-\\ source\end{tabular}} & Objective                                                                                      & LN                                                   & Major Lang.                                                    & FPF  & Stabilization                                                       & Quantization                                        & GPU Needed                                                                                             \\ \midrule
GPT-3 175B              & $\times$                                                                 &                                                                                                &                                                      & English                                                        & FP16 & {\color[HTML]{9B9B9B} \textit{\small{undisclosed}}}                 & {\color[HTML]{9B9B9B} \textit{\small{undisclosed}}} & {\color[HTML]{9B9B9B} \textit{\small{undisclosed}}}                                                    \\
OPT-175B                & $\checkmark$                                                             &                                                                                                &                                                      & English                                                        & FP16 & Manual Adjusting                                                    & INT8                                                & 8 $\times$ 3090                                                                                        \\
BLOOM-176B              & $\checkmark$                                                             & \multirow{-3}{*}{GPT}                                                                          & \multirow{-3}{*}{Pre-LN}                             & Multi-lingual                                                  & BF16 & Embedding Norm                                                      & INT8                                                & 8 $\times$ 3090                                                                                        \\ \midrule
PaLM 540B               & $\times$                                                                 & GPT                                                                                            & Pre-LN                                               & English                                                        & BF16 & Manual Adjusting                                                    & {\color[HTML]{9B9B9B} \textit{\small{undisclosed}}} & {\color[HTML]{9B9B9B} \textit{\small{undisclosed}}}                                                    \\ \midrule
\glm                    & $\checkmark$                                                             & {\color[HTML]{CB0000} \begin{tabular}[c]{@{}c@{}}GLM (Blank \\ Infilling \& MIP)\end{tabular}} & \begin{tabular}[c]{@{}c@{}}Deep-\\ Norm\end{tabular} & \begin{tabular}[c]{@{}c@{}}Bilingual\\ (EN \& CN)\end{tabular} & FP16 & \begin{tabular}[c]{@{}c@{}}Embedding\\ Gradient Shrink\end{tabular} & INT4                                                & {\color[HTML]{CB0000} \begin{tabular}[c]{@{}c@{}}4 $\times$ 3090 or\\ 8 $\times$ 1080 Ti\end{tabular}} \\ \bottomrule[1.2pt]
\end{tabular}
}
\vspace{-6mm}
\label{tab:intro}
\end{table}

In this work, we introduce the pre-training of a 100B-scale model---\glm, in terms of engineering efforts, model design choices, training strategies for efficiency and stability, and quantization for affordable inference. 
As it has been widely realized that it is computationally unaffordable to empirically enumerate all possible designs for training 100B-scale LLMs, we present not only the successful part for training \glm but also many of the failed options and lessons learned. 
Particularly, the training stability is {the} decisive factor in the success of training models of such a scale. 
Different from practices such as manually adjusting learning rates in OPT-175B and using embedding norm in the sacrifice of performance in BLOOM-176B, we experiment with various options and find the strategy of embedding gradient shrink can significantly stabilize the training of \glm.  

Specifically, \glm is a bilingual (English and Chinese) bidirectional dense model with 130 billion parameters, pre-trained over 400 billion tokens 
on a cluster of 96 NVIDIA DGX-A100 (8$\times$40G) GPU nodes 
between May 6 and July 3, 2022. 
Instead of using the GPT-style architecture, we adopt the  General Language Model (GLM) algorithm~\citep{du2022glm} to leverage its bidirectional attention advantage and autoregressive blank infilling objective. 
Table~\ref{tab:intro} summarizes the comparison between \glm, GPT-3 and another two open-source efforts---OPT-175B and BLOOM-176B, as well as PaLM 540B~\citep{chowdhery2022palm}---a 4$\times$ larger model---as a reference. 

Altogether, the conceptual uniqueness and engineering efforts enable \glm to exhibit performance that surpasses the level of GPT-3 on a wide range of benchmarks (in total 112 tasks) and also outperforms PaLM 540B in many cases, while outperformance over GPT-3 has not been observed in OPT-175B and BLOOM-176B (Cf. Figure ~\ref{fig:intro} left). 
For zero-shot performance, \glm is better than GPT-3 175B (+5.0\%), OPT-175B (+6.5\%), and BLOOM-176B (+13.0\%) on LAMBADA~\citep{paperno2016lambada}, and achieves 3$\times$ better performance than GPT-3 on Big-bench-lite~\citep{srivastava2022beyond}. 
For the 5-shot MMLU~\citep{hendrycks2021measuring} tasks, it is  better than GPT-3 175B (+0.9\%) and BLOOM-176B (+12.7\%). 
As a bilingual LLM also in Chinese, it offers significantly better results than ERNIE TITAN 3.0 260B~\citep{wang2021ernie}---the largest Chinese LLM---on 7 zero-shot CLUE~\citep{xu2020clue} datasets (+24.26\%) and 5 zero-shot FewCLUE~\citep{xu2021fewclue} ones (+12.75\%). 
Importantly, as summarized in Figure~\ref{fig:intro} right, \glm as an open model is associated with \textit{significantly less bias and generation toxicity than its 100B-scale counterparts}. 

Finally, we design \glm to empower as many people as possible to conduct 100B-scale LLM studies. 
First, instead of using 175B+ parameters as OPT and BLOOM, the 130B size is decided because such a size supports inference on a single A100 (8$\times$40G) server. 
Second, to further lower the GPU requirements, we quantize \glm into INT4 precision without post training  while OPT and BLOOM can only reach INT8. 
Due to a unique property of the GLM architecture, \glm's INT4 quantization introduces negligible performance degradation, e.g., -0.74\% on LAMBADA and even +0.05\% on MMLU, making it still better than the uncompressed GPT-3. 
This enables \glm's fast inference with performance guarantee on a server of 4$\times$RTX 3090 (24G) or 8$\times$RTX 2080 Ti (11G), \textit{the most affordable GPU required for using 100B-scale LLMs to date. }

We open-source the model checkpoints, code, training logs, related toolkits, and lessons learned. 


\vspace{-3mm}
\section{The Design Choices of \glm}
\vspace{-2mm}
The architecture of a machine learning model defines its inductive bias. 
However, it has been realized that it is computationally unaffordable to explore various architectural designs for LLMs.
We introduce and explain the unique design choices of \glm. 


\vspace{-2mm}
\subsection{\glm's Architecture} \label{sec:glm-arch}
\vspace{-2mm}


\vvpara{GLM as  Backbone.}
Most recent 100B-scale LLMs, such as GPT-3, PaLM, OPT, and BLOOM, follow the traditional GPT-style~\citep{radford2019language} architecture of decoder-only autoregressive language modeling. 
In \glm, we instead make an attempt to explore the potential of a bidirectional GLM---General Language Model~\citep{du2022glm}---as its backbone.



GLM is a transformer-based language model that leverages autoregressive blank infilling as its training objective.
Briefly, for a text sequence $\boldsymbol{x}=[x_1,\cdots,x_n]$, text spans $\{\boldsymbol{s}_1,\cdots,\boldsymbol{s}_m\}$ are sampled from it, each of which $\boldsymbol{s}_i$ denotes a span of consecutive tokens $[s_{i,1},\cdots,s_{i,l_i}]$ and is replaced (i.e., corrupted) with a single mask token to form $\boldsymbol{x}_{\text{corrupt}}$. 
The model is asked to recover them autoregressively.
To allow interactions between corrupted spans, their visibility to each other is decided by a randomly sampled permutation on their order. 

GLM's bidirectional attention over unmasked (i.e., uncorrupted) contexts distinguishes \glm from GPT-style LLMs in which the unidirectional attention is used. 
To support both understanding and generation, it mixes two corruption objectives, each indicated by a special mask token:
\begin{itemize}[leftmargin=*,itemsep=0pt,parsep=0.2em,topsep=0.0em,partopsep=0.0em]
    \item \textbf{[MASK]}: short blanks in sentences whose lengths add up to a certain portion of the input.
    \item \textbf{[gMASK]}: random-length long blanks at the end of sentences with prefix contexts provided.
\end{itemize}

\begin{wrapfigure}{r}{6cm}
    \small
    \vspace{-6mm}
    \centering
    \includegraphics[width=1.0\linewidth]{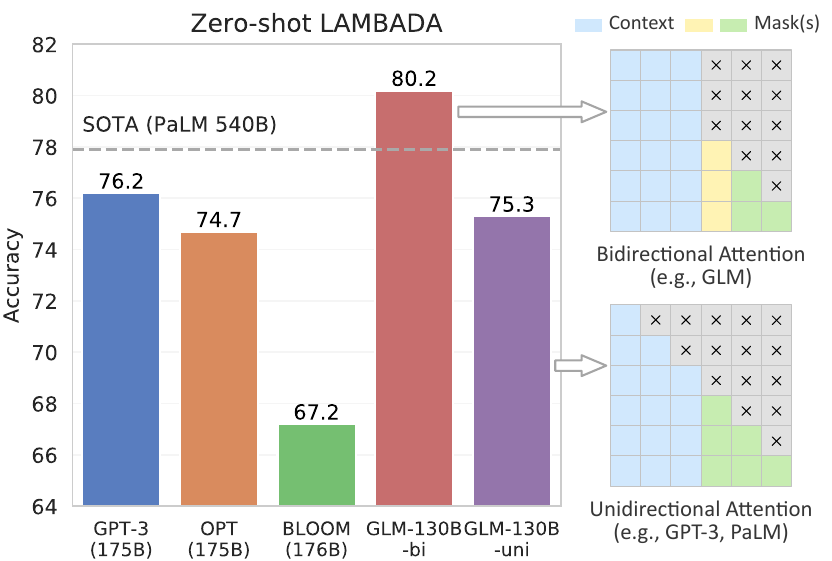}
    \vspace{-6mm}
    \caption{\glm and LLMs of similar scale on zero-shot LAMBADA language modeling. 
    Details on GLM's bidirectional attention are provided in \citet{du2022glm}.}
    \label{fig:lambada}
    \vspace{-8mm}
\end{wrapfigure}


Conceptually, the blank infilling objective with bidirectional attention enables a more effective comprehension of contexts than GPT-style models: 
when using [MASK], \glm behaves as BERT~\citep{devlin2019bert} and T5~\citep{raffel2020exploring}; 
when using [gMASK], \glm behaves similarly to PrefixLM~\citep{liu2018generating,dong2019unified}.

Empirically, 
\glm 
offers a record-high accuracy of 80.2\% on zero-shot LAMBADA  by outperforming both GPT-3 and PaLM 540B in Figure \ref{fig:lambada}. 
By setting the attention mask, \glm's  unidirectional variant is comparable to GPT-3 and OPT-175B. 
Our observations are in line with existing findings~\citep{liu2018generating,dong2019unified}. 

\begin{figure}[t]
    \centering
    \includegraphics[width=.92\linewidth]{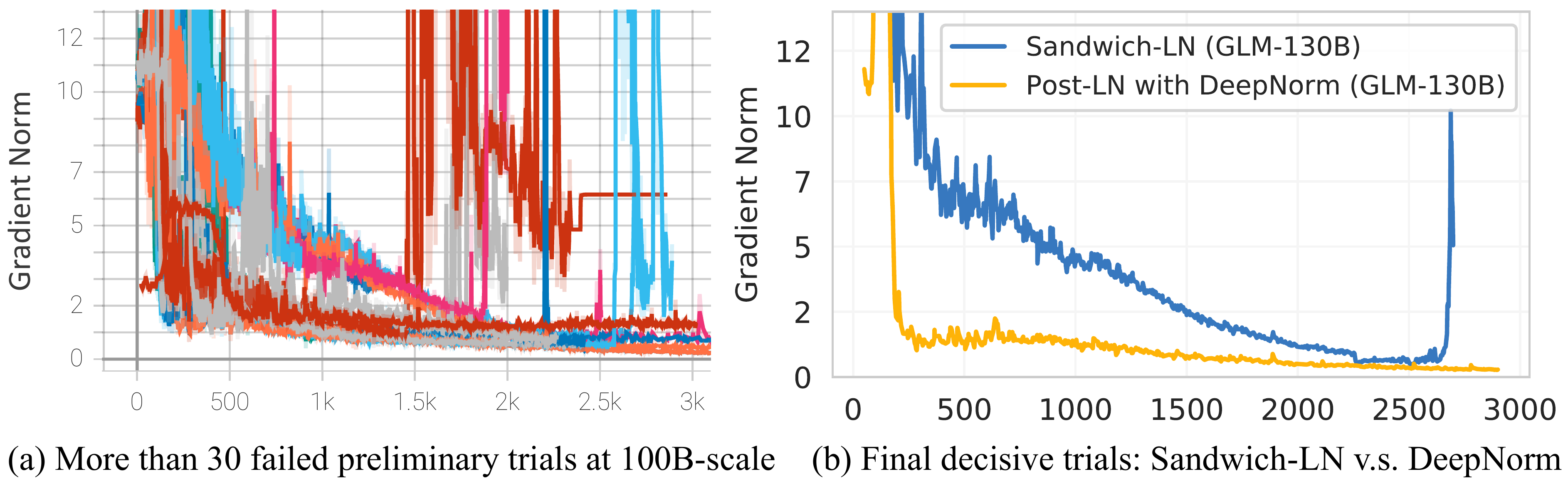}
    \vspace{-2mm}
    \caption{Trials on different LayerNorms for \glm training. It turns out that DeepNorm is the most stable one, as it has small gradient norm and does not spike in the early stage training.}
    \vspace{-6mm}
    \label{fig:layernorm}
\end{figure}

\vvpara{Layer Normalization (LN,~\cite{ba2016layer}).}
Training instability is one major challenge for training LLMs~\citep{zhang2022opt,scao2022what,chowdhery2022palm} (Cf. Figure ~\ref{fig:collapse} in Appendix for  collapses in training several 100B-scale models). 
A proper choice of LNs can help stabilize the training of LLMs. 
We experiment with existing practices, e.g., Pre-LN~\citep{xiong2020layer}, Post-LN~\citep{ba2016layer}, 
Sandwich-LN~\citep{ding2021cogview}, which are unfortunately incapable of stabilizing our \glm test runs (Cf. Figure~\ref{fig:layernorm} (a) and Appendix~\ref{app:ln} for details). 


Our search is later focused on Post-LN due to its favorable downstream results in preliminary experiments though it does not stabilize \glm. 
Fortunately, one of the attempts on Post-LN initialized with the newly-proposed DeepNorm~\citep{wang2022deepnet} generates promising training stability. 
Specifically, given the number of \glm's layers $N$, 
    we adopt $\textrm{DeepNorm}(\boldsymbol{x}) = \textrm{LayerNorm}(\alpha\cdot \boldsymbol{x} + \textrm{Network}(\boldsymbol{x}))$, where $\alpha=(2N)^\frac{1}{2}$, 
    and 
   apply the Xavier normal initialization with the scaling factor of $(2N)^{-\frac{1}{2}}$ to \texttt{ffn}, \texttt{v\_proj} and \texttt{out\_proj}. Additionally, all bias terms are initialized to zero.
Figure~\ref{fig:layernorm} shows it significantly benefits the training stability of \glm. 




\vvpara{Positional Encoding and FFNs.}
We empirically test different options for positional encoding (PE) and FFN improvements in terms of both training stability and downstream performance (Cf. Appendix \ref{app:pe-ffn} for details).  
For PEs in \glm, we adopt Rotary Positional Encoding (RoPE, \cite{su2021roformer}) rather than  ALiBi~\citep{press2021train}. 
To improve FFNs in Transformer, we pick GLU with the GeLU~\citep{hendrycks2016gaussian} activation as the replacement.

\vspace{-2mm}
\subsection{\glm's Pre-Training Setup} \label{sec:training_objective}
\vspace{-2mm}

Inspired by recent works~\citep{aribandi2022ext5,wei2022finetuned,sanh2022multitask}, the \glm pre-training objective includes not only the self-supervised GLM autoregressive blank infilling) but also multi-task learning for a small portion of tokens. 
This is expected to help boost its downstream zero-shot performance. 


\vvpara{Self-Supervised Blank Infilling (95\% tokens).}
Recall that \glm uses both [MASK] and [gMASK] for this task. 
Each training sequence is applied with one of them independently at a time.
Specifically, [MASK] is used to mask consecutive spans in 30\% of training sequences for blank infilling. The lengths of spans follow a Poisson distribution ($\lambda=3$) and add up to 15\% of the input. 
For the other 70\% sequences, the prefix of each sequence is kept as context and [gMASK] is used to mask the rest of it. 
The masked length is sampled from the Uniform distribution.


The pre-training data includes 1.2T Pile (train split)~\citep{gao2020pile} English, 1.0T Chinese WudaoCorpora~\citep{yuan2021wudaocorpora}, 
and 250G Chinese corpora (including online forums, encyclopedia, and QA) we crawl from the web, which form a balanced composition of English and Chinese contents.

\vvpara{Multi-Task Instruction Pre-Training (MIP, 5\% tokens).}
T5~\citep{raffel2020exploring} and ExT5~\citep{aribandi2022ext5} suggest that multi-task learning in pre-training can be more helpful than fine-tuning, we thus propose to include a variety of instruction prompted datasets including language understanding, generation, and information extraction in \glm's pre-training.

Compared to recent works~\citep{wei2022finetuned,sanh2022multitask} that leverage multi-task prompted fine-tuning to improve zero-shot task transfer, MIP only accounts for 5\% tokens and is set in the pre-training stage to prevent spoiling LLMs' other general ability, e.g., unconditional free generation.
Specifically, we include 74 prompted datasets from ~\citep{sanh2022multitask,wang2022deepstruct}, listed in Appendix~\ref{app:data} and Table~\ref{tab:mip}. 
\glm users are suggested to {avoid evaluating its zero-shot and few-shot capabilities on these datasets} according to the criterion illustrated in Section~\ref{sec:results}.


\subsection{Platform-Aware Parallel Strategies and Model Configurations} \label{sec:parallel_strategy}


\glm is trained on a cluster of 96 DGX-A100 GPU (8$\times$40G) servers with a 60-day access.  
The goal is to pass through as many tokens as possible, as a recent study~\citep{hoffmann2022training} suggests that most existing LLMs are largely under-trained. 

\vvpara{The 3D Parallel Strategy.}
The data parallelism~\citep{valiant1990bridging} and tensor model parallelism~\citep{shoeybi2019megatron} are the de facto practices for training billion-scale models~\citep{gpt-j,du2022glm}. 
To further handle
the huge GPU memory requirement and the decrease in overall GPU utilization resulted from applying tensor parallel between nodes---as 40G rather than 80G A100s are used for training \glm, we combine the pipeline model parallelism with the other two strategies to form a 3D parallel strategy. 

The pipeline parallelism divides the model into sequential stages for each parallel group, and to further minimize bubbles introduced by pipeline, we leverage the PipeDream-Flush~\citep{narayanan2021memory} implementation from DeepSpeed~\citep{rasley2020deepspeed} to train \glm with a relative big global batch size (4,224) to reduce time and GPU memory wasting. 
Through both numerical and empirical examinations, we adopt 4-way tensor parallelism and 8-way pipeline parallelism (Cf. Appendix~\ref{app:pipeline} for details). Following the calculation in~\citep{chowdhery2022palm}, we report hardware FLOPs utilization (HFU) of 43.3\% and model FLOPs utilization (MFU) of 32.5\% due to re-materialization.


\vvpara{\glm Configurations.}
We aim to enable our 100B-scale LLM to run a single DGX-A100 (40G) node in FP16 precision. 
Based on the hidden state dimension of 12,288 we adopt from GPT-3, the resultant model size has to be no more than 130B parameters, thus \glm.
To maximize GPU utilization, we configure the model based on the platform and its corresponding parallel strategy. 
To avoid insufficient memory utilization in the middle stages
due to the additional word embedding at both ends, we balance the pipeline partition by removing one layer from them, making 9$\times$8-2=70 transformer layers in \glm.

During the 60-day access to the cluster, we manage to train \glm for 400 billion tokens (roughly 200 billion each for Chinese and English) with a fixed sequence length of 2,048 per sample. 
For the [gMASK] training objective, we use a context window of 2,048 tokens. 
For the [MASK] and multi-task objectives, we use a context window of 512 and concatenate four samples together to cater the 2,048-sequence-length. 
We warm-up the batch size from 192 to 4224 over the first 2.5\% samples. 
We use AdamW~\citep{loshchilov2017decoupled} as our optimizer with $\beta_1$ and $\beta_2$ set to 0.9 and 0.95, and a weight decay value of 0.1. 
We warm up the learning rate from $10^{-7}$ to $8\times 10^{-5}$ over the first 0.5\% samples, then decay it by a $10\times$ cosine schedule. 
We use a dropout rate of 0.1 and clip gradients using a clipping value of 1.0 (Cf. Table~\ref{tab:config} for the full configurations).


\section{The Training Stability of \glm}

The training stability is the decisive factor in \glm's quality, which is also largely impacted by the number of tokens it passes through~\citep{hoffmann2022training}. 
Thus, given the computing usage constraint, there has to be a trade-off between efficiency and stability with regard to floating-point (FP) formats: 
low-precision FP formats (e.g., 16-bit precision---FP16) improve computing efficiency but are prone to overflow and underflow errors, resulting in training collapses. 

 \begin{wrapfigure}{r}{4.3cm}
    \centering
    \vspace{-3mm}
    \includegraphics[width=\linewidth]{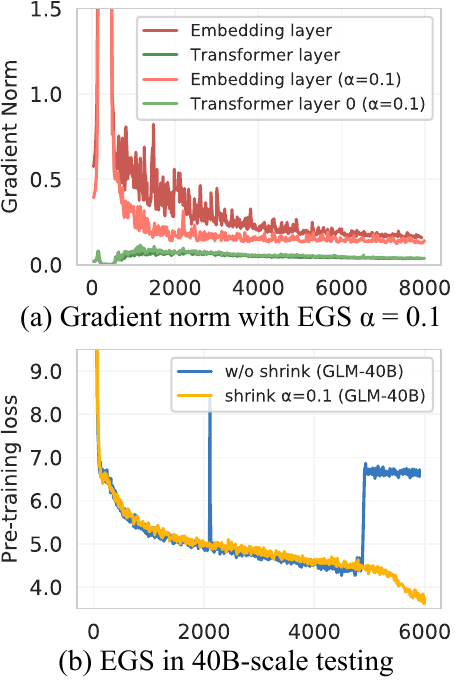}
    \vspace{-7mm}
    \caption{EGS reduces gradient scale and variance to stabilize LLMs' pre-training.}
    \label{fig:shrink}
    \vspace{-8mm}
\end{wrapfigure}

\vvpara{Mixed-Precision.}
We follow the common practice of a mixed-precision~\citep{micikevicius2018mixed} strategy (Apex O2), i.e., FP16 for forwards and backwards and FP32 for optimizer states and master weights, to reduce the GPU memory usage and improve training efficiency. 
Similar to OPT-175B and BLOOM-176B (C.f. Figure~\ref{fig:collapse} in Appendix), the training of \glm faces frequent loss spikes resulted from this choice, which tends to become increasingly frequent as the training goes on. 
The precision related spikes are often without clear reasons: 
some recover on their own; others come with a portent of suddenly soaring gradient norm and eventually a spike or even NaN in loss. 
OPT-175B attempted to fix by manually skipping data and adjusting hyper-parameters; BLOOM-176B did so via the embedding norm technique~\citep{dettmers20218}. 
We spent months to empirically investigate the spikes and realize that a few issues emerge when transformers scale up: 

First, the transformer main branch's value scale can be extremely large in deeper layers if using Pre-LN. 
This is addressed in \glm by using DeepNorm based Post-LN (Cf.  Section ~\ref{sec:glm-arch}), which makes the value scale always bounded. 

Second, the attention scores grow so large that they exceed FP16's range, as the model scales up. 
There are a few options to overcome this issue in LLMs. 
In CogView~\citep{ding2021cogview}, PB-Relax is proposed to remove bias terms and deduct extremum value in attention computation to avoid the problem, which unfortunately does not help avoid disconvergence in \glm. 
In BLOOM-176B, the BF16 format is used instead of FP16, due to its wide range of values on NVIDIA Ampere GPUs (i.e., A100). 
However, BF16 consumes $\sim$15\% more run-time GPU memory than FP16 in our experiments due to its conversion to FP32 in gradient accumulation, and more importantly it is not supported on other GPU platforms (e.g., NVIDIA Tesla V100), limiting the accessibility of produced LLMs. 
Another option from BLOOM-176B is to apply embedding norm with BF16, but in sacrifice of a significant penalty on model performance, as they notice that embedding norm can harm model's zero-shot learning (Cf. Section 4.3 in~\citep{scao2022what}). 


\vvpara{Embedding Layer Gradient Shrink (EGS).}
Our empirical search identifies that the gradient norm can serve as an informative indicator of training collapses. 
Specifically, we find that 
a training collapse usually lags behind a ``spike'' in gradient norm by a few training steps. 
Such spikes are usually caused by the embedding layer's abnormal gradients, as we observe that its gradient norm is often several magnitude larger that those of other layers in \glm's early stage training (Cf. Figure~\ref{fig:shrink} (a)). 
In addition, it tends to fluctuate dramatically in the early training. 
The problem is handled in vision models~\citep{chen2021empirical} via freezing the patch projection layer. 
Unfortunately, we cannot freeze the training of the embedding layer in language models.

Finally, we find the gradient shrink on embedding layers could overcome loss spikes and thus stabilize \glm's training. 
It is first used in the multi-modal transformer CogView~\citep{ding2021cogview}. 
Let $\alpha$ be the shrinking factor, the strategy can be easily implemented via
  $  \mathsf{word\_embedding} = \mathsf{word\_embedding} * \alpha + \mathsf{word\_embedding.detach()} * (1 - \alpha)$. 
Figure~\ref{fig:shrink} (b) suggests that empirically, setting $\alpha=0.1$ wipes out most spikes we would have met, with negligible latency.

In fact, the final \glm training run only experiences three late-stage loss divergence cases, though it fails numerous times due to hardware failures. 
For the three unexpected spikes, it turns out further shrinking the embedding gradient can still help stabilize the \glm training. 
See the training notes and Tensorboard logs in our code repository
for details. 

\hide{
\section{Training Strategy}

The root cause of  is that there is a trade-off between training efficiency and stability with regard to floating-point (FP) formats:
\begin{itemize}[leftmargin=*,itemsep=0pt,parsep=0.2em,topsep=0.0em,partopsep=0.0em]
\item \textbf{Efficiency}: low-precision FP formats (e.g., FP16) reduce memory and computation costs
\item \textbf{Stability}: low-precision FP formats are prone to overflow and underflow, resulting in collapses
\end{itemize}

\subsection{Stability: Systematic and Numerical Challenges}

 Modern computing devices such as NVIDIA A100s and TPUs support 16-bit precision computation with much higher throughput compared to 32-bit precision. 
Therefore, it has become a de facto practice to train language models with mixed-precision~\citep{micikevicius2018mixed} to reduce GPU memory usage and improve computation efficiency without hurting performance. 
However, as far as we have observed, LLMs are far more vulnerable to loss spikes in the mixed-precision setup than smaller ones, which is well-aligned with other LLMs' observations (C.f. Figure~\ref{fig:collapse} in Appendix) such as OPT-175B~\citep{zhang2022opt} and BLOOM-176B~\citep{scao2022what}.

Handling these unexpected spikes has been the most challenging part in \glm's training. 
We empirically ascribe them into structural and numerical instability:

\begin{itemize}[leftmargin=*,itemsep=0pt,parsep=0.2em,topsep=0.0em,partopsep=0.0em]
\item \textbf{Systematic Instability}:
which tends to appear in the early stage of training. 
Its appearance is often accompanied with a gradual but steady rise in gradient norm. 
Skipping data or adjusting hyper-parameters do not eliminate, but only make it happen at other random neighboring steps. 

\item \textbf{Numerical Instability}:
which instead tends to occur in the late stage of training and becomes increasingly frequent as the training goes on. 
Its resultant spikes are often without clear reasons: some recover on their own; others come with a portent of suddenly soaring gradient norm and eventually a spike or even NaN in loss, as the model has been optimized by the abnormal gradients. 
\end{itemize}

We refer to the first type of instability as \emph{systematic}, since regardless of its optimization configuration, a model just cannot converge ideally owing to its current setup (i.e., model architectures and mixed-precision training strategies). 
To eliminate consequent disconvergence, a more stable model architecture and proper mixed-precision strategies are necessary.

The second type of instability is referred to as \emph{numerical}, since we believe it is likely to derive from some unknown numerical reasons in 16-bit precision training.
Its related spikes can be mitigated via re-loading the model from a checkpoint 100 to 200 iterations ahead with data skipping, or changing hyper-parameters such as the learning rate.
Such instability may not arise from bad data samples as is previously assumed, because changing the learning rate also allows the model to go through them.
Authors of~\citep{chowdhery2022palm} also report that no bad patterns have been observed from data causing training spikes in their examination.
Thus, numerical spikes are likely to occur only when a certain model goes through certain data, and can be avoided by applying some perturbation to the training process.

Due to the large cost of LLM training, conquering these spikes are crucial: it would be a tragedy if the training diverged on the half road and turned out unrecoverable. 
We conduct plenitude of experiments spanning few months to find several effective strategies to stabilize \glm's training.

\begin{insight}
\rm It is both systematic and numerical instabilities that LLMs suffer from in training, which takes us months to identify and conquer from scratch in \glm.
\end{insight}

\vvpara{Mixed-precision strategy.} 
Mixed-precision~\citep{micikevicius2018mixed} training with FP16 has become a default in mainstream frameworks such as DeepSpeed~\citep{rasley2020deepspeed}.
A common mixed-precision training strategy (Apex O2) is as follows: forward and backward are calculated in pure 16-bit precision (FP16), but the optimizer states and master weights are FP32 are used in the optimization.
However, such a strategy is prone to meet severe numerical instability in a conventional transformer architecture when a model scales up, due to its two main bottlenecks:
\begin{itemize}[leftmargin=*,itemsep=0pt,parsep=0.2em,topsep=0.0em,partopsep=0.0em]
    \item Under Pre-LN, the transformer main branch's value scale can be extremely large in deeper layers.
    \item As a model scales up, attention scores grow so large that they exceed FP16's representation scope.
\end{itemize}

For the first bottleneck, our use of DeepNorm (a variant of Post-LN) servers as LayerNorm in the main branch, so the value scale is always bounded. 
For the second bottleneck, CogView~\citep{ding2021cogview} proposes PB-Relax to remove bias terms in attention's linear projection and deduct the largest attention score from each attention score matrix to avoid the problem. 
Nevertheless, the technique does not eliminate disconvergence in \glm, probably due to its gigantic size.
 
As a remedy for fp16's narrow representation range, NVIDIA Ampere GPUs provide BF16 floating-point format (adopted by BLOOM 176B) to mitigate the problem. 
Nevertheless, BF16's dissupport on other computing platforms (e.g., NVIDIA Tesla V100) has significantly hampered its wider applications.
In practice, we also notice that compared to FP16, BF16 requires around 15\% additional run-time GPU memory due to its necessary conversion to FP32 in gradient accumulation.

To support as many researchers and developers as possible, \glm thus chooses FP16 as its training floating-point format. 
On the other hand, it means our model is faced with more stability challenges, as is also reported in another FP16 LLM OPT-175B. 
Fortunately, after great struggle, we successfully get it through via a set new techniques derived from some following insights.

\begin{insight}
\rm FP16 is a challenging but rewarding decision: it suffers from more instability, but is a must for enabling LLMs to train and inference on inclusive ranges of platforms.
\end{insight}

\vvpara{Gradient Shrink on the Embedding Layer.}
Besides explicit precision issues caused by attention scores, there are implicit problems arouses by random noisy samples appearing along the training.
For example, in our observation these samples might consist bunches of semantically plausible repeating lines, which can be hardly cleaned in rule-guided preprocesses.
We find LLMs are so vulnerable to these noises, which likely results in unrecoverable unexpected training collapses.

\begin{figure}[t]
    \centering
    \includegraphics[width=\linewidth]{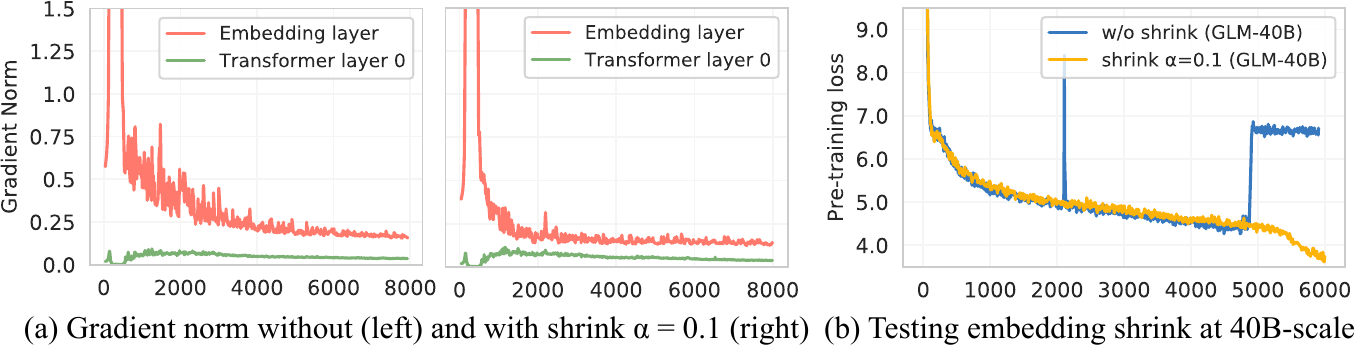}
    \vspace{-6mm}
    \caption{Gradient shrink on embedding layer stabilizes \glm's training. (a) Preliminary experiments on GLM-base show that gradient norm for embedding layer can be much larger; (b) 40B-scale testing shows that embedding gradient shrink is necessary for LLM training.}
    \label{fig:shrink}
    \vspace{-6mm}
\end{figure}

For such instability, we notice that the gradient norm serves as an effective indicator, which accords with observations from vision transformers~\citep{chen2021empirical,ding2021cogview}: the happening of a training collapse usually lags behind a ``spike'' in gradient norm by few training steps. 
Such spikes are usually caused by embedding layer's abnormal gradient, as we observe a remarkably larger gradient norm from it than other layers---often several magnitudes larger---in \glm's early stage training (Cf. Figure~\ref{fig:shrink} (a)).
It also tends to fluctuate dramatically in the early stage training of GLM compared to other layers.
The issue is handled in~\citep{chen2021empirical} via freezing the patch projection layer; however, a language model cannot freeze the training of its embedding layer.

Alternatively, we find the strategy of gradient shrink over embedding layer can overcome the problem in LLM's training, which is similarly applied in text-to-image transformer CogView~\citep{ding2021cogview}.
Let $\alpha$ be the shrinking factor, the strategy can be easily implemented via
\begin{equation}
    \mathsf{word\_embedding} = \mathsf{word\_embedding} * \alpha + \mathsf{word\_embedding.detach()} * (1 - \alpha)
\end{equation}
Specifically, we shrink to $\alpha=0.1$ and discover that it wipes out most spikes we would have met, with negligible speed loss (Cf. Figure~\ref{fig:shrink} (b)).
Actually, in our late stage training, we observe that further shrinking embedding gradient still works when unexpected spikes take place.

\begin{insight}
\rm Shrinking embedding layer's gradient to its 0.1 can solve most instability problems.
\end{insight}

We train \glm for 400 billion tokens (roughly 200 billion each for Chinese and English) with a fixed sequence length of 2,048 per sample. 
For the [gMASK] training objective, we use a context window of 2,048 tokens. 
For the [MASK] and multi-task training objectives, we use context windows of length 512 and concatenate four samples together to cater the 2,048-sequence-length. 
We warm-up the batch size from 192 to 4224 over the first 2.5\% samples. 
We use AdamW~\citep{loshchilov2017decoupled} as our optimizer with $\beta_1$ and $\beta_2$ set to 0.9 and 0.95, and a weight decay value of 0.1. 
We warm-up the learning rate from $10^{-7}$ to $8\times 10^{-5}$ over the first 0.5\% samples, then decay it by a $10\times$ cosine schedule. 
We use a dropout rate of 0.1 and clip gradients using a clipping value of 1.0.

\glm's pre-training lasts 60 days and it takes up 96 NVIDIA DGX-A100 (40G) nodes with 400G bandwidth IB network, which would cost equivalently 4.9 million dollars based on the GPU pricing on public cloud services in the same period.
On top of the basic training setup, during the long training we have experienced several crises and failures.
We successfully managed to recover the training via slight adjustments over some of configuration above (mostly about learning rate and shrinking factor), which are reported in our released training notes and Tensorboard logs\footnote{\url{https://github.com/THUDM/GLM-130B/tree/main/logs}}.

}
\section{\glm Inference on RTX 2080 Ti}


One of the major goals of \glm is to lower the hardware requirements for accessing 100B-scale LLMs without efficiency and effectiveness disadvantages. 

As mentioned, the model size of 130B is determined for running the full \glm model on a single A100 (40G$\times$8) server, rather than the high-end A100 (80G$\times$8) machine required by OPT-175B and BLOOM-176B. 
To accelerate \glm inference, we also leverage FasterTransformer~\citep{timonin2022accelerated}
to implement \glm in C++.  
Compared to the PyTorch implementation of BLOOM-176B in Huggingface, \glm's decoding inference is 7-8.4$\times$ faster on the same single A100 server. (Cf. Appendix~\ref{app:inference_acceleration} for details).

\vvpara{INT4 Quantization for RTX 3090s/2080s.}
To further support popularized GPUs,
we attempt to compress \glm as much as possible while maintaining performance superiority, particularly via  quantization~\citep{zafrir2019q8bert,shen2020q,tao2022compression}, which introduces little task-agnostic performance drops for generative language models. 

\begin{wrapfigure}{r}{6cm}
    \small
    \vspace{-6mm}
    \centering
    \includegraphics[width=1.0\linewidth]{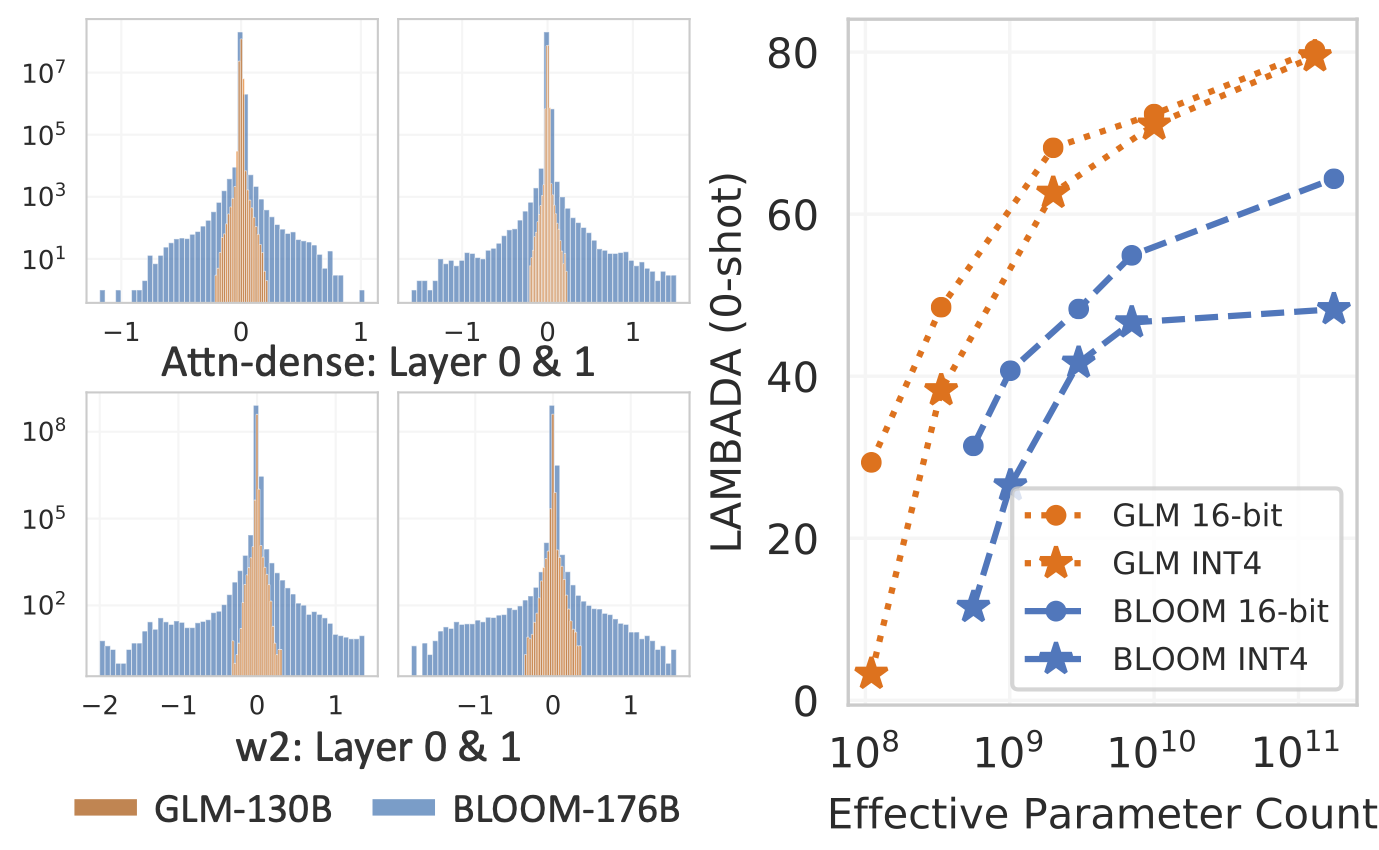}
    \vspace{-4mm}
    \caption{(Left) \texttt{attn-dense} and \texttt{w2}'s weight distributions; (Right) \glm's INT4 weight quantization scaling law.}
    \label{fig:scaling_law}
    \vspace{-5mm}
\end{wrapfigure}

Typically, the practice is to quantize both model weights and activations to INT8.
However, our analysis in Appendix~\ref{app:activation_analysis} suggests that LLMs' activations may contain extreme outliers. 
Concurrently, the emergent outliers in OPT-175B and BLOOM-176B are also discovered~\citep{dettmers2022llm}, which influence only about 0.1\% feature dimensions and are thus solved by matrix multiplication decomposition for the outlying dimensions.
Differently, there exist about 30\% outliers in GLM-130B's activations, making the technique above far less efficient. 
Thus, we decide to focus on the quantization of model weights (i.e., mostly linear layers) while keeping the FP16 precision for activations.  
The quantized model is dynamically converted to FP16 precision at runtime, introducing a small computational overhead but greatly reducing the GPU memory usage for storing model weights. 

Excitingly, we manage to reach the INT4 weight quantization for \glm while existing successes have thus far only come to the INT8. 
Memory-wise, by comparing to INT8, the INT4 version helps additionally save half of the required GPU memory to 70GB, thus allowing \glm inference on 4 $\times$ RTX 3090 Ti (24G) or 8 $\times$ RTX 2080 Ti (11G). 
Performance-wise, Table~\ref{tab:quantization} left indicates that without post-training at all, the INT4-version \glm experiences almost no performance degradation, thus maintaining the performance advantages over GPT-3 on common benchmarks. 

\vvpara{GLM's INT4 Weight Quantization Scaling Law.}
We examine the underlying mechanism of this unique INT4 weight quantization scaling law exhibited in Figure~\ref{fig:scaling_law} right. 
We plot the weight value distributions in  Figure~\ref{fig:scaling_law} left, which turns out to directly impact the quantization quality.
Specifically, a wider-distributed linear layer needs to be quantized with larger bins, leading to more precision loss.
Thus the wide-distributed \texttt{attn-dense} and \texttt{w2} matrices explain the INT4 quantization failure for GPT-style BLOOM.
Conversely, GLMs tend to have much narrower distributions than those of similar-sized GPTs, and the gap between INT4 and FP16 versions keeps further decreasing as the GLM model size scales up (Cf.  Figure~\ref{fig:quantization_appendix} in Appendix for details).

\begin{table*}[t]
\footnotesize
\centering
\caption{Left: Quantized \glm's performance on several benchmarks; Right: INT4 quantized \glm's inference speed (encode and decode) with FasterTransformer.}
\vspace{-2mm}
\begin{subtable}
    \centering
    \begin{threeparttable}
    \centering
    \renewcommand\tabcolsep{3pt}
    \renewcommand\arraystretch{0.75}
    \begin{tabular}{@{}llccc@{}}
    \toprule[1.2pt]
    \multirow{2}{*}{Model Precision} & \multicolumn{3}{c}{\glm} & GPT-3 \\ \cmidrule(l){2-4} \cmidrule(l){5-5} 
                                     & FP16   & INT8   & INT4   & FP16  \\ \midrule
    MMLU (acc, $\uparrow$)           & 44.75  & 44.71  & 44.80  & 43.9  \\
    LAMBADA (acc, $\uparrow$)        & 80.21  & 80.21  & 79.47  & 76.2  \\
    Pile (a part, BPB, $\downarrow$) & 0.634  & 0.638  & 0.641  & 0.74  \\ \bottomrule[1.2pt] 
    \end{tabular}
    \end{threeparttable}
\end{subtable}%
\hspace{.02\linewidth}%
\begin{subtable}
    \centering
    \begin{threeparttable}
    \centering
    \renewcommand\tabcolsep{2pt}
    \renewcommand\arraystretch{0.95}
    \begin{tabular}{@{}lcccc@{}}
    \toprule[1.2pt]
    GPU Type                     & \multicolumn{2}{c}{128 Enc./Dec.} & \multicolumn{2}{c}{512 Enc./Dec,} \\ \midrule
    8 $\times$ A100 (40G)        & 0.15s           & 4.29s           & 0.18s           & 17.7s           \\
    8 $\times$ V100 (32G)        & 0.31s           & 6.97s           & 0.67s           & 28.1s           \\
    4 $\times$ RTX 3090 (24G)    & 0.37s           & 8.16s           & 1.30s           & 32.3s           \\
    8 $\times$ RTX 2080 Ti (11G) & 0.39s           & 6.77s           & 1.04s           & 27.3s           \\ \bottomrule[1.2pt]
    \end{tabular}
    \end{threeparttable}
    \end{subtable}
    \vspace{-4mm}
\label{tab:quantization}
\end{table*}





\hide{

\section{Inference for Inclusivity}
In terms of LLM inclusivity, we think its scope should not be limited to accessibility, but also usability.
It is particularly a fact that, even with released LLM checkpoints such as OPT-175B~\citep{zhang2022opt} and BLOOM-176B~\citep{scao2022what}, individuals and academic researchers can hardly afford the cost for inference, which have been reported to depend on an 8 $\times$ A100 (80G) server.
Thus, a considerable amount of sweat we spend in \glm has been on supporting its high usability and fast inference speed on popularized GPUs, such as NVIDIA RTX 3090 Ti (24G).

Interestingly, such efforts actually deeply associate with our \textit{algorithm designs}, rather than mere engineering undertakings.
We will show that GLM series presents \textit{a unique scaling law to allow INT4 quantization}, while other existing public LLMs (OPT and BLOOM series) would fail.

\vvpara{Inference Acceleration.}
A model's plain PyTorch implementation is easy to read and run; but it can be intolerably slow for LLMs. 
Based on NVIDIA's FasterTransformer\footnote{\url{https://github.com/NVIDIA/FasterTransformer}} we spend two months \aohan{no need to mention?} implementing \glm into C++ to speed up its inference (Cf. Appendix~\ref{app:inference_acceleration} for details).
Compared to a naive implementation (e.g., BLOOM-176B by Huggingface Transformers), \glm's decoding inference on the same-length sequences can be up to $\times$8.4 faster.

\vvpara{GLM's INT4 Weight Quantization Scaling Law.}
Notwithstanding \glm's accelerated inference speed, its demand for at least a DGX-A100 (40G) server can still discourage individual users.
In light of the situation, we endeavor to compress \glm to fit into 4 $\times$ RTX 3090 Ti (24G)'s capacity, particularly via the quantization~\citep{zafrir2019q8bert,shen2020q,tao2022compression}, which presents little task-agnostic performance drop for generative language models.

\begin{wrapfigure}{r}{4.5cm}
    \small
    \vspace{-5mm}
    \centering
    \includegraphics[width=1.0\linewidth]{figures/quantization_scaling_law.pdf}
    \vspace{-4mm}
    \caption{\glm's INT4 weight quantization scaling law.}
    \label{fig:scaling_law}
    \vspace{-8mm}
\end{wrapfigure}

Typical methods quantize both model weights and activation to INT8.
But in our analysis, we find LLMs' activations may contain extreme outliers (Cf. Appendix~\ref{app:activation_analysis}).
Concurrent to our efforts, \citep{dettmers2022llm} also discovers the emergent outliers in OPT-175B and BLOOM-176B, which influence only about 0.1\% feature dimensions and are thus solved by matrix multiplication decomposition for outlying dimensions.
However, we find substantial outliers (30\%) in GLM-130B's activations, making it far less efficient to apply due to the slow high-precision computation.

After a few weeks of trial, we finally decide to keep activations' FP16 precision and only consider the symmetric quantization of model weights (majorly consist of linear layers). 
The quantized model is dynamically converted to FP16 precision at runtime, introducing a small computational overhead but greatly reducing GPU memory usage for storing model weights.

Surprisingly we notice \glm's unusual talent for INT4 weight quantization, which saves half of required GPU memory to 70GB compared to INT8 and thus allows \glm's inference on even 4 $\times$ RTX 3090 Ti (24G) or 8 $\times$ RTX 2080 Ti (11G).
More surprising, it requires no post-training at all and shows little quality degradation (Cf. Table~\ref{tab:quantization}).
However, such a nice property does not apply to BLOOM-176B and OPT series models (Cf. Figure~\ref{fig:scaling_law}).
To understand it deeper, we dive into LLMs' weight value distribution (Cf. Figure~\ref{fig:weight_dist},~\ref{fig:quantization_appendix}), which directly impact the quantization quality.
We find language models' skewed \texttt{w2} matrix is the major cause for INT4 quantization failure, especially for GPTs.
On the contrary, GLMs usually present more symmetrical \texttt{w2} value distributions than similar-sized GPTs, and it seems to keep improving as they scale up.
We summarize the discovery as \textbf{GLM's Scaling Law for INT4 Weight Quantization}, which demonstrates GLM's unique advantage to be quantized into INT4 precision as the architecture scales up.

\begin{figure}[t]
    \centering
    \includegraphics[width=\linewidth]{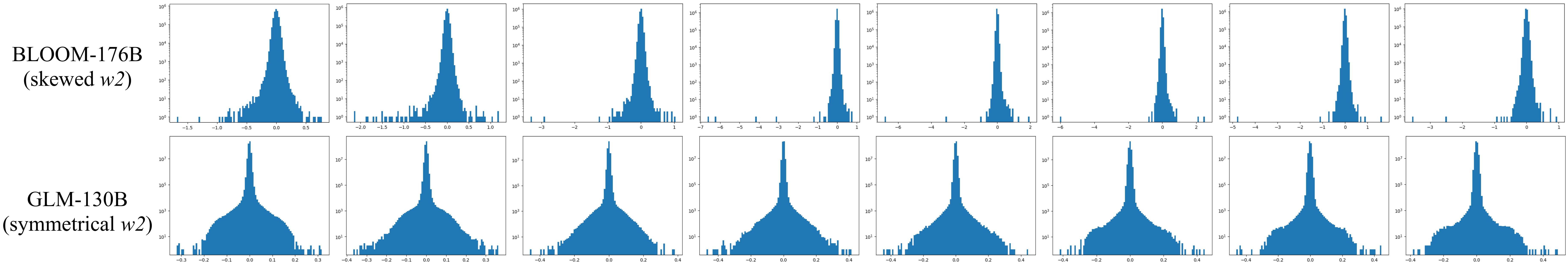}
    \caption{Weight value distribution of \texttt{w2} from BLOOM-176B and \glm's first-8 layers' FFN (or GLU). A more skewed distribution with outliers is likely to suffer from more quantization loss.}
    \label{fig:weight_dist}
    \vspace{-5mm}
\end{figure}


\section{Inference for Inclusivity}
In terms of LLM inclusivity, we think its scope should not be limited to accessibility, but also usability.
It is particularly a fact that, even with released LLM checkpoints such as OPT-175B~\citep{zhang2022opt} and BLOOM-176B~\citep{scao2022what}, individuals and academic researchers can hardly afford the cost for inference, which have been reported to depend on an 8 $\times$ A100 (80G) server.
Thus, a considerable amount of sweat we spend in \glm has been on supporting its high usability and fast inference speed on popularized GPUs, such as NVIDIA RTX 3090 Ti (24G).

Interestingly, such efforts actually deeply associate with our \textit{algorithm designs}, rather than mere engineering undertakings.
We will show in the following part that \glm is the only LLM so far being able to be quantized into INT4 precision, while other existing public LLMs (OPT-175B, BLOOM-176B, and even those 10B-scale smaller language models) would fail.

\subsection{Inference Acceleration.}
\du{This section can be mostly moved to appendix, leaving only the resutls.}
A model's plain Pytorch implementation is easy to read and run; but it can be intolerably slow for LLMs. 
Based on NVIDIA's FasterTransformer\footnote{\url{https://github.com/NVIDIA/FasterTransformer}} we spend two months implementing \glm into C++ to speed up inference, including the following main optimizations: 
\begin{itemize}[leftmargin=*,itemsep=0pt,parsep=0.2em,topsep=0.0em,partopsep=0.0em]
    \item Optimize time-costing operations such as GeGLU, Layer Normalization and SoftMax.
    \item Reduce number of GPU kernel calls (e.g., fuse MultiheadAttention into one computation kernel).
    \item Specify the algorithm of the best performance when calling cuBLAS.
    \item Improve the computing efficiency by transposing the model parameters in advance.
    \item Use half2 in FP16 computation to double the half's access bandwidth and computing throughput.
\end{itemize}

We currently pack up the full FasterTransformer implementation for \glm into a plug-and-play docker image for users' convenience, and we are still working on adapting it to our Pytorch implementation
by only changing one line of code.
A comparison between our speeding up \glm implementation and the so far default available BLOOM-176B implementation in Huggingface Transformers\footnote{\url{https://huggingface.co/docs/transformers/model_doc/bloom}} is shown in Table~\ref{tab:fastertransformer}.
Our implementation for \glm can be 7.0 to 8.4 times faster than BLOOM-176B's Pytorch implementation.
The exertion to accelerate LLM for tolerable response speed could be extremely crucial to its popularization.

\subsection{Compression via Quantization}
We have improved \glm's inference speed substantially, but its demand for at least a DGX-A100 (40G) server can still discourage individual developers and researchers.
In light of the situation, we endeavor to compress \glm to fit into 4 $\times$ RTX 3090 Ti (24G)'s capacity, particularly via the quantization~\citep{zafrir2019q8bert,shen2020q,tao2022compression}, which presents little task-agnostic performance drop for generative language models.

\vvpara{INT4 Weight Quantization.}
Typical methods quantize both model weights and activation to INT8, enabling the INT8 matrix multiplication kernel for efficiency. 
However, we find substantial outliers in GLM-130B's activations, making it hard for precision reduction. 
Concurrent to our efforts and observations in \glm, researchers~\citep{dettmers2022llm} also discover the issue of emergent outliers in LLMs. 
In their in-depth analysis on OPT-175B and BLOOM-176B, they find the outliers make up only about 0.1\% of all feature dimensions.
Therefore, they propose to decompose the matrix multiplication for high-precision multiplication in outlying dimensions.

But our detailed analysis demonstrates that the outliers in GLM-130B can sometimes make up at most 30\% of the feature dimension, which is different from GPT-based LLMs. 
Therefore, a mixed-precision decomposition for \texttt{matmul} can be far less efficient than a single FP16 \texttt{matmul}. 
After a few weeks of trial, we finally decide to keep the FP16 precision of activation and only consider the quantization of model weights. 
In this case, the quantized \glm is dynamically converted to FP16 precision at runtime, introducing a small computational overhead but greatly reducing GPU memory usage for storing model weights.
We quantize all linear layers, which take up most of the transformer parameters, via vector-wise symmetric quantization, and leave input/output embedding, layer normalization, and bias terms unchanged. 

Surprisingly we note \glm's unusual talent for INT4 weight quantization.
At the quantization precision of INT4, two INT4 weights are compressed into one INT8 weight for saving GPU memory usage.
Thus approximately only 70GB GPU memory is required to serve INT4 \glm weights, which allow \glm's inference on even 4 $\times$ RTX 3090 Ti (24G) or 8 $\times$ RTX 2080 Ti (11G).
More surprising, it requires no post-training at all and shows almost no quality degradation.
On the contrary, it does not apply to BLOOM-176B in our experiments.

\vvpara{Empirical Understanding.}
Such nice property for \glm is fascinating but mysterious.
We try our best to understand it and empirically demonstrate that the INT4 weight quantization for \glm may be a compound result of multiple factors, rather than a single one.

In terms of whether a model can be properly quantized, the most fundamental reason lies in the skewness of its weight distribution.
Since existing methods on weight mostly rely on symmetrical quantization, if weight distributions in certain linear layers are too skewed, the quantization would lose much precision.
\begin{itemize}[leftmargin=*,itemsep=0pt,parsep=0.2em,topsep=0.0em,partopsep=0.0em]
    \item \textbf{GLM pre-training objective}: 
    \item \textbf{Gated Linear Unit (GLU)}:
    \item \textbf{Scaling}:
\end{itemize}

\begin{insight}
\rm Backbone architecture matters in LLMs' post quantization. While GLM scaled up to \glm can be quantized to as low as INT4, other LLMs only support INT8.
\end{insight}

} 

\section{The Results}
\label{sec:results}

We follow the common settings in LLMs such as GPT-3 and PaLM to evaluate \glm for English~\footnote{\small Results in OPT-175B's paper are reported as applications to access it have not been approved for months.}. 
As a bilingual LLM with Chinese, \glm is also evaluated on Chinese benchmarks.

\vvpara{Discussion on the Scope of Zero-Shot Learning in \glm.} 
Since \glm has been trained with MIP, here we clarify its scope of zero-shot evaluation.
In fact, ``zero-shot'' seems to have controversial interpretations without a consensus in the community. 
We follow one of the influential related surveys~\citep{xian2018zero}, which says 
\textit{``At test time, in zero-shot learning setting, the aim is to assign a test image to an unseen class label''}
where involving unseen class labels is a key. 
Therefore, we derive our criterion to pick \glm's zero-shot (and few-shot) datasets as:
\begin{itemize}[leftmargin=*,itemsep=0pt,parsep=0.2em,topsep=0.0em,partopsep=0.0em]
    \item \textbf{English}: 1) For tasks with fixed labels (e.g., \textit{natural language inference}): no datasets in such tasks should be evaluated on; 2) For tasks without fixed labels (e.g., \textit{(multiple-choice) QA, topic classification}): only datasets with an obvious domain transfer from those in MIP should be considered. 
    \item \textbf{Chinese}: All datasets can be evaluated as there exists a zero-shot cross-lingual transfer.
\end{itemize}


\vvpara{Filtering Test Datasets.}
Following prior practices~\citep{brown2020language,rae2021scaling} and our criterion mentioned above, we filter and refrain to report potentially contaminated datasets' evaluation results.
For LAMBADA and CLUE, we find minimal overlap under the 13-gram setting.
Pile, MMLU, and BIG-bench are either held-out or released later than the crawling of corpora.


\subsection{Language Modeling} \label{sec:language_modeling}

\vvpara{LAMBADA.}
LAMBADA~\citep{paperno2016lambada} is a dataset to test the last word language modeling capability.
The results previously shown in Figure~\ref{fig:lambada} suggest \glm achieves a zero-shot accuracy of 80.2 with its bidirectional attention, setting up a new record on LAMBADA.

\begin{wraptable}{r}{5.5cm}
	\centering
	\footnotesize
	\vspace{-6mm}
    \renewcommand\tabcolsep{2pt}
\renewcommand\arraystretch{1}
	\caption{\glm's average BPB on Pile evaluation (18 sub-datasets).}
	\vspace{-3mm}
	\scalebox{0.97}{
	\begin{tabular}{@{}lccc@{}}
    \toprule[1.2pt]
                       & Jurassic-1     & GPT-3          & \glm           \\ \midrule
    Avg. BPB      & 0.650          & 0.742          & \textbf{0.634} \\ \bottomrule[1.2pt]
    \end{tabular}
    }
    \vspace{-4mm}
\end{wraptable}

\vvpara{Pile.} 
The Pile test-set~\citep{gao2020pile} includes a series of benchmarks for language modeling.
On average, \glm performs the best on its 18 shared test sets in terms of weighted BPB when compared to GPT-3 and Jurassic-1~\citep{lieber2021jurassic} whose results are directly adopted from the latter, demonstrating its strong language capability (Cf. Appendix~\ref{app:pile} for details).

\subsection{Massive Multitask Language Understanding (MMLU)} \label{sec:mmlu}


MMLU~\citep{hendrycks2021measuring} is a diverse benchmark including 57 multi-choice question answering tasks concerning human knowledge ranging from high-school-level to expert-level. 
It is released after the crawling of Pile and serves as an ideal test-bed for LLMs' few-shot learning.
The GPT-3 result is adopted from MMLU and BLOOM-176B is tested by using the same prompts as \glm's 
(Cf. Appendix~\ref{app:mmlu} and Table~\ref{tab:mmlu} for details).

\glm's few-shot (5-shot) performance on MMLU approaches GPT-3 (43.9) after viewing about 300B tokens in Figure~\ref{fig:mmlu}. 
It continues moving up as the training proceeds, achieving an accuracy of 44.8 when the training has to end (i.e., viewing 400B tokens in total).  
This aligns with the observation~\citep{hoffmann2022training} that most existing LLMs are far from adequately trained.

\subsection{Beyond the Imitation Game Benchmark (BIG-bench)} \label{sec:big-bench}


\begin{figure}[t]
\begin{minipage}{0.33\linewidth}
    \small
    \vspace{-5mm}
    \centering
    \includegraphics[width=1.0\linewidth]{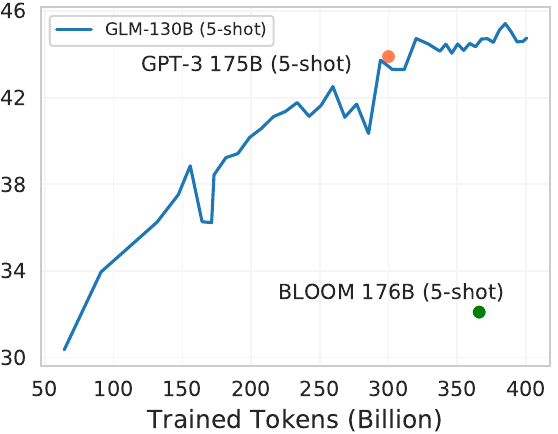}
    \vspace{-4mm}
    \caption{\glm on MMLU (57 tasks) along training steps.}
    \label{fig:mmlu}
    \vspace{-6mm}
\end{minipage}
\hspace{.03\linewidth}%
\begin{minipage}{0.34\linewidth}
    \centering
	\includegraphics[width=0.99\columnwidth]{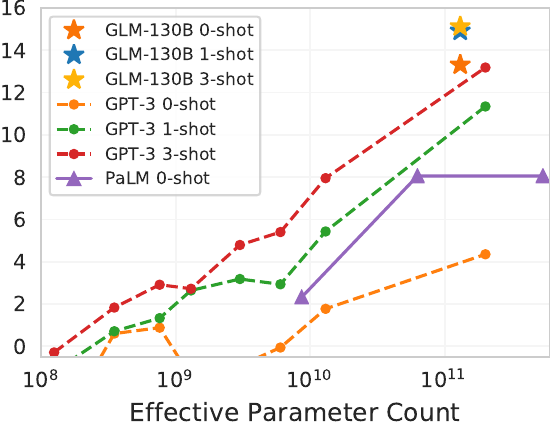}
	\vspace{-5mm}
	\caption{BIG-bench-lite evaluation (24 tasks) across scales.}
	\label{fig:big-bench}
\end{minipage}
\hspace{.002\linewidth}%
\begin{minipage}{0.27\linewidth}
    \centering
    \renewcommand\tabcolsep{1pt}
    \renewcommand\arraystretch{1.2}
    \scalebox{0.9}{
    \footnotesize
    \begin{tabular}{@{}lccc@{}}
    \toprule[1.2pt]
               & 0-shot         & 1-shot         & 3-shot         \\ \midrule
    GPT-3 2.6B & 0.60           & 0.71           & 1.83           \\
    GPT-3 6.7B & -0.06          & 2.93           & 5.40           \\
    GPT-3 13B  & 1.77           & 5.43           & 7.95           \\
    GPT-3 175B & 4.35           & 11.34          & 13.18          \\ \midrule
    PaLM 540B  & 8.05           & \textbf{37.77} & -              \\ \midrule
    \glm       & \textbf{13.31} & 14.91          & \textbf{15.12} \\ \bottomrule[1.2pt]
    \end{tabular}}
    \vspace{1mm}
    \captionof{table}{Details on BIG-bench-lite (24 tasks).}
    \label{tab:big-bench}
\end{minipage}  
\vspace{-5mm}
\end{figure}

BIG-bench~\citep{srivastava2022beyond} benchmarks challenging tasks concerning models' ability on reasoning, knowledge, and commonsense. 
Given evaluating on its 150 tasks is time-consuming for LLMs, we report the BIG-bench-lite---an official 24-task sub-collection---for now.
Observed from Figure~\ref{fig:big-bench} and Table~\ref{tab:big-bench}, 
\glm outperforms GPT-3 175B and even PaLM 540B (4$\times$ larger) in zero-shot setting. 
This is probably owing to \glm's bidirectional context attention and MIP, which has been proved to improve zero-shot results in unseen tasks~\citep{wei2022finetuned,sanh2022multitask}. 
As the number of shots increases, \glm's performance keeps going up, maintaining its outperformance over GPT-3  
(Cf. Appendix~\ref{app:big-bench} and Table~\ref{tab:big-bench-details} for details on each model and task).

\vvpara{Limitations and Discussions.}
In the experiments above, we observe that \glm's performance growth (13.31 to 15.12) with the increase of few-shot samples is not as significant as GPT-3's (4.35 to 13.18). 
Here is our intuitive attempt to understand the phenomenon. 

First, the bidirectional nature of \glm could lead to strong zero-shot performance (as is indicated in zero-shot language modeling), thus getting closer to the few-shot ``upper-bound'' for models of similar scale (i.e., 100B-scale) than unidirectional LLMs.
Second, it may be also attributed to a deficit of existing MIP paradigms~\citep{wei2022finetuned,sanh2022multitask}, which only involve zero-shot prediction in the training and will be likely to bias \glm for stronger zero-shot learning but relatively weaker in-context few-shot performance.
To correct the bias, a potential solution we came up with would be to employ MIP with varied shots of in-context samples rather than only zero-shot samples.

Finally, despite almost the same GPT architecture as GPT-3, PaLM 540B's relative growth with few-shot in-context learning is substantially more significant than GPT-3's. 
We conjecture this further acceleration in performance growth is a source of PaLM's high-quality and diverse private-collected training corpora. 
By combining our experiences with~\citep{hoffmann2022training}'s insights, we came to realize that better architectures, better data, and more training FLOPS should be further invested. 

\subsection{Chinese Language Understanding Evaluation (CLUE)} \label{sec:clue}

\begin{figure}[t]
    \centering
    \includegraphics[width=\linewidth]{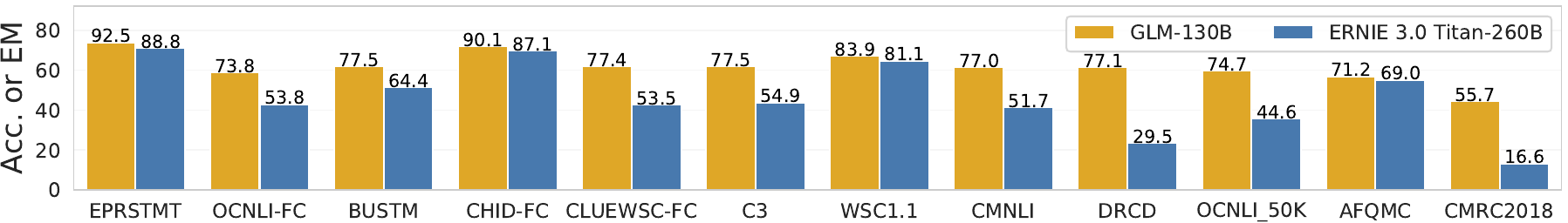}
    \vspace{-6mm}
    \caption{\glm and ERNIE Titan 3.0 260B evaluated on zero-shot CLUE and FewCLUE.}
    \label{fig:clue}
    \vspace{-4mm}
\end{figure}

We evaluate \glm's Chinese zero-shot performance on established Chinese NLP benchmarks, CLUE~\citep{xu2020clue} and FewCLUE~\citep{xu2021fewclue}.
Note that we do not include any Chinese downstream tasks in MIP. 
To date, we have finished testing on part of the two benchmarks, including 7 CLUE and 5 FewCLUE datasets (Cf. Appendix~\ref{app:clue} for details). 
We compare GLM-130B to the largest existing Chinese monolingual language model---the 260B ERNIE Titan 3.0~\citep{wang2021ernie}. 
We follow its setting to report zero-shot results on dev datasets. 
\glm consistently outperforms ERNIE Titan 3.0 across 12 tasks (Cf. Figure~\ref{fig:clue}). 
Interestingly, \glm performs at least 260\% better than ERNIE on two abstractive MRC datasets (DRCD and CMRC2018), possibly due to \glm's pre-training objective that naturally resonates to abstractive MRC's form.

\section{Related Work}
In this section, we review related work to \glm on topics of pre-training, transferring, and inference of pre-trained LLMs~\citep{qiu2020pre,bommasani2021opportunities}.

\vvpara{Pre-Training.}
Vanilla language modeling refers to decoder-only autoregressive models (e.g., GPT~\citep{radford2018improving}), but it also recognizes any forms of self-supervised objectives on texts. 
Recently, transformer-based~\citep{vaswani2017attention} language models present a fascinating scaling law: new abilities~\citep{wei2022emergent} arise as models scale up, from 1.5B~\citep{radford2019language}, 10B-scale language models~\citep{raffel2020exploring,shoeybi2019megatron,black2022gpt}, to 100B-scale GPT-3~\citep{brown2020language}.
Later, despite many 100B-scale LLMs~\citep{lieber2021jurassic,thoppilan2022lamda,rae2021scaling,smith2022using,chowdhery2022palm,wu2021yuan,zeng2021pangu,wang2021ernie} in both English and Chinese, they are not available to public or only accessible via limited APIs.
The closeness of LLMs severely stymies its development.
\glm's efforts, along with recent ElutherAI, OPT-175B~\citep{zhang2022opt}, and BLOOM-176B~\citep{scao2022what}, aim to offer high-quality open-sourced LLMs to our community.

\vvpara{Transferring.}
Though fine-tuning has been a \textit{de facto} way for transfer learning, the evaluation for LLMs has been focused on prompting and in-context learning due to their tremendous sizes~\citep{brown2020language,liu2021pre}.
Nevertheless, some recent attempts has been on parameter-efficient learning on language models
~\citep{houlsby2019parameter} and prompt tuning (i.e., P-tuning, ~\cite{li2021prefix,liu2021gpt,lester2021power,liu2022p}). 
For now we do not focus on them and will leave the comprehensive testing of them on \glm in future study.

\vvpara{Inference.}
Most public-accessible LLMs nowadays are providing their services via limited APIs.
In this work, an important part of our endeavor has been on LLMs' efficient and fast inference.
Related work may include distillation~\citep{sanh2019distilbert,jiao2020tinybert,wang2020minilm}, quantization~\citep{zafrir2019q8bert,shen2020q,tao2022compression}, and pruning~\citep{michel2019sixteen,fan2019reducing}.
Very recent work~\citep{dettmers2022llm} shows that LLMs such as OPT-175B and BLOOM-176B can be quantized to 8 bit due to special distribution of outlier dimensions.
In this work, we demonstrate GLM's scaling law for INT4 weight quantization, which allows \glm to inference on as few as 4$\times$RTX 3090 (24G) GPUs or 8$\times$RTX 2080 Ti (11G) GPUs.

\section{Conclusion and Lessons}

We introduce \glm, a bilingual pre-trained language model that aims to facilitate open and inclusive LLM research. 
\glm's technical and engineering undertakings generate insight into LLMs' architectures, pre-training objectives, training stability and efficiency, and affordable inference. Altogether, it contributes to the high quality of \glm in terms of both language performance on 112 tasks and ethical results on bias and toxicity benchmarks. 
Our experiences of both success and failure are condensed into the lessons for training 100B-scale LLMs, attached in the Appendix \ref{sec:lessons}.

\section*{Acknowledgement}
This research was supported by Natural Science Foundation of China (NSFC) 61825602, 62276148 and Zhipu.AI.
We thank all our collaborators and partners from the Knowledge Engineering Group (KEG), Parallel Architecture \& Compiler technology of Mobile, Accelerated, and Networked systems Group (PACMAN), Natural Language Processing Group (THUNLP) at Tsinghua University, and Zhipu.AI.

\section*{Ethics Statement}
We hereby acknowledge that all of the co-authors of this work are aware of the provided ICLR Code of Ethics and honor the code of conduct.
This work introduces an open-source Large Language Model (LLM), which could be used to generate synthetic text for harmful applications, such as telemarketing fraud, political propaganda, and personal harassment as is discussed in~\citep{weidinger2021ethical,Sheng2021SocietalBI,Dev2021HarmsOG}. 
We do not anticipate any hazardous outputs, especially towards vulnerable and historically disadvantaged groups of peoples, after using the model.

And to better collaborate with our community to prevent and ultimately eliminate the risks technically, we make the following crucial open efforts in this work:

\vvpara{Open-Sourced LLMs for Ethical Risk Study.}
While some people think that restricting the access of LLMs can prevent such harmful applications, we argue that promoting LLM inclusivity can lead to better defense against potential harms caused by LLMs. 
Currently, only governments and large corporations can afford the considerable costs of pre-training LLMs. 
There is no guarantee that organizations having the the substantial financial resources will not do harm using a LLM. 
Without access to such LLMs, individuals cannot even realize the role of LLMs in the harm. 

Conversely, releasing an open LLM can provide access and transparency to all the researchers and promote the research to reduce the potential harm of LLMs, like algorithms to identify the synthetic text~\cite{gehrmann2019gltr}. 
Also, it is known that LLMs can suffer from problems in fairness, bias, privacy, and truthfulness~\cite{abs-2112-12938,lin2022truthfulqa,Liang2021SocialBias,Bender2021Danger}. 
An open LLM can reveal the model parameters and internal states corresponding to specific inputs instead of providing APIs to black-box models. 
In conclusion, researchers can conduct analysis of LLMs' flaws in depth and propose improved algorithms to solve the problems. 

\vvpara{Ethical Evaluation and Improvements.}
We also evaluate our model over a wide range of English ethical evaluation benchmarks, including bias measurement~\citep{nadeem2021stereoset,nangia2020crows}, hate speech detection~\citep{mollas2020ethos}, and toxic generation estimation~\citep{gehman2020realtoxicityprompts}.
Notwithstanding their deficiency~\citep{blodgett2021stereotyping,jacobs2021measurement}, these datasets serve as a meaningful initial step towards an open quantitative evaluation LLMs.

Our evaluation implies that our algorithm designs, especially the bilingual pre-training of a LLM, can significantly mitigate the biases and toxicity an LLM may present while keeping its strong language performance compared to other LLMs~\citep{brown2020language,zhang2022opt} trained with monolingual English corpora (Cf. Appendix~\ref{app:ethics} for more details).

\section*{Reproducibility}
Compared to mainstream closed-sourced LLMs including GPT-3 175B\citep{brown2020language}, PaLM 540B~\citep{chowdhery2022palm}, Gopher~\citep{rae2021scaling}, Chinchilla~\citep{hoffmann2022training}, LaMDA~\citep{thoppilan2022lamda}, FLAN~\citep{wei2022finetuned}, and many others, \glm is open-sourced and devotes to promote openness and inclusivity in LLM research from the very beginning.

We have paid great effort to ensure the reproducibility of our evaluation.
For pre-training section, despite the unaffordable costs it needs to reproduce at present, we still make our best efforts to disclose the code, details, and the whole process of \glm's pre-training. 
Our endeavor to allow \glm inference on few popularized GPUs such as 3090/2080 Ti also aligns with the reproducibility undertaking, as it allows most academic researchers to reproduce \glm's results on their offline machines.
We also provide free APIs for individual users to test \glm's ability.

\vvpara{Pre-Training.}
We provide the complete training notes, Tensorboard logs, and code for our pre-training in our repository (Cf. Abstract).
The pre-training hyper-parameters and cluster configuration are provided in Section~\ref{sec:parallel_strategy} and Table~\ref{tab:config}.
The training corpora composition and details for Multi-task Instruction Pre-training are provided in Section~\ref{sec:training_objective} and Appendix~\ref{app:mip_description} and~\ref{app:mip_dataset}. 

\vvpara{Evaluation.}
We organize all the evaluation, including language benchmarks (LAMBADA, Pile, MMLU, BIG-bench, CLUE, and FewCLUE) and ethical benchmarks (CrowS-Pairs, StereoSet, ETHOS, RealToxicPrompts), into one-command-to-run bash scripts in our code repository.
Data processing details for language modeling benchmarks are provided in Section~\ref{sec:language_modeling} and Appendix~\ref{app:pile}, for MMLU are provided in Section~\ref{sec:mmlu} and Appendix~\ref{app:mmlu}, for BIG-bench are provided in Section~\ref{sec:big-bench} and Appendix~\ref{app:big-bench}, for CLUE and FewCLUE are provided in ~\ref{sec:clue}.
For all ethical evaluation, please refer to Appendix~\ref{app:ethics} for details.

\bibliography{ref}
\bibliographystyle{iclr2023_conference}

\clearpage
\appendix
\part{Appendix} 
\parttoc 

\section{Ethics: Evaluation on Biases and Toxicity} \label{app:ethics}
Albeit LLMs' strong abilities in language and beyond, which could bring substantial welfare to human beings, they can potentially produce toxic and illegal contents for evil use~\citep{weidinger2021ethical,Sheng2021SocietalBI,Dev2021HarmsOG,bommasani2021opportunities}.
In \glm, before granting model weight to applicants, in the model license we demand them to agree that they will not use it for any deeds that may be harmful to society and human beings.

Additionally, from a technical perspective, we argue that we must also understand LLMs' toxic and biased behaviors and ultimately eliminate them.
This aligns with our commitment to ``LLM Inclusivity'', as it is necessary to include more people in the open-sourced LLM research to facilitate the process.
Moreover, if an LLM is shown to be good at identifying toxic and biased content, techniques such as self-diagnoses~\citep{schick2021self} can help to reduce the harmful generation in a self-consistent post-processing procedure.
Therefore, as an initial step, we evaluate \glm over a variety of related benchmarks to shed light on the challenging topic.
Despite their limitations~\citep{blodgett2021stereotyping,jacobs2021measurement} which should be addressed in future work, they still serve as a good start to arouse the community's awareness of the problem.


\subsection{Bias Measurement: CrowS-Pairs}

\begin{wraptable}{r}{6.8cm}
\vspace{-10mm}
\caption{CrowS-Pairs~\citep{nangia2020crows} Bias Measurement. The lower scores the better.}
\footnotesize
\vspace{-2mm}
\renewcommand\tabcolsep{1.5pt}
\begin{tabular}{@{}lccc@{}}
\toprule[1.2pt]
Category             & GPT-3         & OPT-175B      & \glm          \\ \midrule
Gender               & 62.6          & 65.7          & \textbf{55.7} \\
Religion             & 73.3          & \textbf{68.6} & 73.3          \\
Race/Color           & 64.7          & 68.6          & \textbf{58.5} \\
Sexual orientation   & 76.2          & 78.6          & \textbf{60.7} \\
Age                  & 64.4          & 67.8          & \textbf{63.2} \\
Nationality          & \textbf{61.6} & 62.9          & 64.1          \\
Disability           & 76.7          & 76.7          & \textbf{71.6} \\
Physical appearance  & \textbf{74.6} & 76.2          & \textbf{74.6} \\
Socioeconomic status & 73.8          & 76.2          & \textbf{70.9} \\ \midrule
Overall              & 67.2          & 69.5          & \textbf{65.8} \\ \bottomrule[1.2pt]
\end{tabular}
\label{tab:crows-pairs}
\vspace{-4mm}
\end{wraptable}
CrowS-Pairs~\citep{nangia2020crows}, or namely Crowdsourced Stereotype Pairs benchmark, is widely used for measuring biases for masked language models.
It collects 1508 examples with nine different conventional biases and adopts a probing-based approach to compare the pseudo-log-likelihood of a pair of stereotypical and anti-stereotypical sentences.
Since \glm is pre-trained with autoregressive blanking infilling, CrowS-Pairs evaluation is directly applicable.
We compare the GPT-3 Davinci and OPT-175B's results on CrowS-Pairs reported in~\citep{zhang2022opt} with \glm.

Our results are presented in Table~\ref{tab:crows-pairs}.
\glm shows fewer biases on almost all kinds of stereotypes except for religion and nationality.
We speculate that it is because \glm is a bilingual pre-trained LLM that learns the semantics for certain content from both English and Chinese corpora.
Since CrowsS-Pairs' stereotypes mainly draw from the US Equal Employment Opportunities Commission's list\footnote{\url{https://www.eeoc.gov/prohibited-employment-policiespractices}}, 
the bias distributions in two different cultures and languages may be different and consequently reconcile social biases in \glm on a benchmark originally designed for English-language society.
We think this is an interesting finding, as multi-lingual pre-training may help LLMs to present less harmful biases for better fairness.
Finally, we also admit that \glm may in turn presents some special Chinese biases which currently lack testing benchmarks and require considerable future efforts to detect and prevent.

\subsection{Bias Measurement: StereoSet}
Another widely used bias and stereotype evaluation benchmark is StereoSet~\citep{nadeem2021stereoset}, which is also adopted in~\citep{lieber2021jurassic,artetxe2021efficient,zhang2022opt}.
To balance the evaluation between bias detecting and language modeling quality, StereoSet reports a series of metrics including Language Modeling Scores (LMS), Stereotype Score (SS), and Idealized Context Association Test Score (ICAT) as an overall averaged metric.
For example, given the premise ``\textit{She is the twin's mother}'', StereoSet provides three candidate hypothesis: 1) ``\textit{the water is deep}'', 2) ``\textit{she is a lazy, unkind person}'', and 3) ``\textit{she is a kind, caring woman}''.
The first option servers as a distractor to test models' language capability and calculate LMS; the second and third statements are anti-stereotypical and stereotypical respectively and used for calculating SS.
A widely-adopted technique here is to calibrate the likelihood of an option according to its length~\citep{lieber2021jurassic,zhang2022opt}, as the distractor term is particularly short.

Following~\citep{zhang2022opt}, we normalize scores over tokens rather than characters~\citep{lieber2021jurassic} to yield model predictions for calculating the metrics.
The results are shown in Table~\ref{tab:stereoset}.
As we observe, \glm exceedingly outperforms GPT-3 Davinci and OPT-175B on all metrics.
Such results accurately align with our discoveries in language modeling experiments and CrowS-Pairs bias evaluation, that \glm has a high quality in both language modeling and social fairness.

\begin{table}[h]
\vspace{-3mm}
\footnotesize
\centering
\renewcommand\tabcolsep{2.7pt}
\caption{StereoSet~\citep{nadeem2021stereoset} Bias Measurement with LMS ($\uparrow$), SS ($\downarrow$), and ICAT ($\uparrow$).}
\begin{tabular}{@{}lccccccccccccccc@{}}
\toprule[1.2pt]
\multirow{2}{*}{Category} & \multicolumn{3}{c}{Profession}                           & \multicolumn{3}{c}{Gender}                               & \multicolumn{3}{c}{Religion}                             & \multicolumn{3}{c}{Race}                                 & \multicolumn{3}{c}{Overall}                              \\ \cmidrule(l){2-4} \cmidrule(l){5-7} \cmidrule(l){8-10} \cmidrule(l){11-13} \cmidrule(l){14-16} 
                          & LMS & SS & ICAT & LMS & SS & ICAT & LMS & SS & ICAT & LMS & SS & ICAT & LMS & SS & ICAT \\ \midrule 
GPT-3                     & 78.4             & 63.4              & 57.5              & 75.6             & 66.5              & 50.6              & 80.8             & 59.0              & 66.3              & 77.0             & 57.4              & 65.7              & 77.6             & 60.8              & 60.8              \\
OPT-175B                  & 74.1             & 62.6              & 55.4              & 74.0             & 63.6              & 53.8              & 84.0             & 59.0              & 68.9              & 74.9             & 56.8              & 64.8              & 74.8             & 59.9              & 60.0              \\
\glm                      & \textbf{86.5}    & \textbf{59.6}     & \textbf{69.9}     & \textbf{83.9}    & \textbf{63.5}     & \textbf{61.2}     & \textbf{91.0}    & \textbf{53.5}     & \textbf{84.6}     & \textbf{85.7}    & \textbf{54.1}     & \textbf{78.7}     & \textbf{86.0}    & \textbf{57.3}     & \textbf{73.5}     \\ \bottomrule[1.2pt]
\end{tabular}
\label{tab:stereoset}
\vspace{-3mm}
\end{table}

\subsection{Hate Speech Detection: ETHOS}
Social media corpus may contain hate speeches, and to investigate to what extent LLMs know and can help to identify them is crucial.
We adopt the ETHOS dataset originally proposed in~\citep{mollas2020ethos} to detect sexism and racism speech on zero-shot or few-shot datasets created by~\citep{chiu2021detecting}.
GPT-3 Davinci (a public-accessible variant of GPT-3 175B) and OPT 175B are also tested on the benchmark (whose results are reported in~\citep{zhang2022opt}).
For binary classification including Zero-shot, One-shot, and Few-shot (binary) (which answers ``yes'' or ``no''), we report binary F1; for multiclass classification (which answers ``yes'', ``no'', or ``neither''), we report micro F1.
We adopt almost the same prompts as in~\citep{chiu2021detecting}, except aligning the Few-shot (binary) prompt to the form used in One-shot and adding the word ``\texttt{Classification}'' before the colon in the original Few-shot (multiclass) prompt.

\begin{wraptable}{r}{6.8cm}
\vspace{-5mm}
\centering
\footnotesize
\renewcommand\tabcolsep{4pt}
\caption{ETHOS~\citep{mollas2020ethos} Hate speech detection. ``(bi)'' and ``(mul)'' denote binary and multiclass classification respectively. All scores are F1 and the higher the better.}
\begin{tabular}{@{}lccc@{}}
\toprule[1.2pt]
                      & GPT-3         & OPT-175B & \glm          \\ \midrule
Zero-shot             & 62.8          & 66.7     & \textbf{68.8} \\
One-shot              & 61.6          & 71.3     & \textbf{79.1} \\
Few-shot (bi)         & 35.4          & 75.9     & \textbf{79.7} \\
Few-shot (mul)        & 67.2          & 81.2     & \textbf{85.8} \\ \bottomrule[1.2pt]
\end{tabular}
\label{tab:ethos}
\vspace{-5mm}
\end{wraptable}

Results are shown in Table~\ref{tab:ethos}.
We find that \glm outperforms two other LLMs among four different settings.
On one hand, \glm's pre-training over unsupervised diverse corpora from online forums and social media including sections such as ``hackernews'', ``stackexchange'', and ``pile\_cc'' can endow our model with the background knowledge to identify those speeches.
On the other hand, the MIP training may also improve \glm's zero-shot and few-shot capabilities.

\subsection{Toxic Genearation: RealToxicPrompts}

\begin{wrapfigure}{r}{6cm}
    \small
    \vspace{-13mm}
    \centering
    \includegraphics[width=1.0\linewidth]{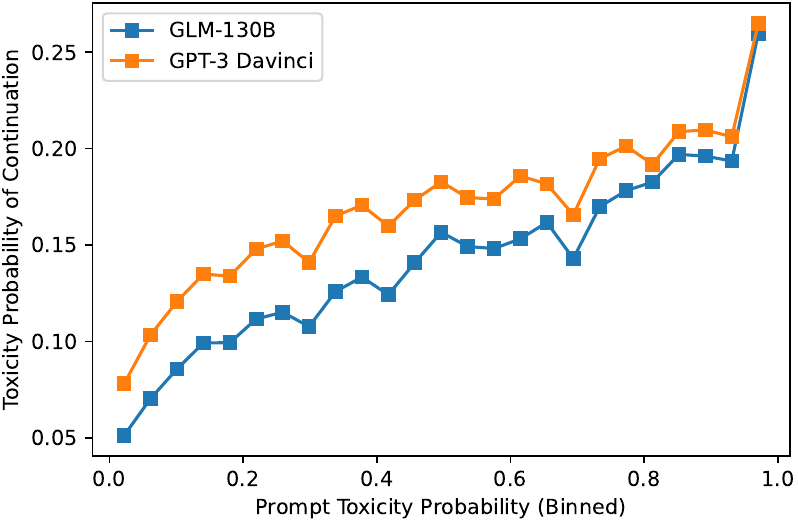}
    \caption{RealToxicPrompts~\citep{gehman2020realtoxicityprompts} evaluation. Lower continuation toxicity probability is better.}
    \label{fig:rtp}
    \vspace{-6mm}
\end{wrapfigure}

Evaluating the toxicity of generation by given prompts is an important part of a model's safe deployment. We evaluate the toxic generation of \glm on the RealToxicPrompts~\citep{gehman2020realtoxicityprompts} dataset. Following its settings, we use nucleus sampling ($p = 0.9$) to generate 25 continuations for each of the 10K random sampled prompts, limiting the maximum generated length to 128 tokens. Then we report the mean toxicity probabilities of 25 continuations evaluated by Perspective API\footnote{\url{https://www.perspectiveapi.com/}}. In order to make a fair comparison under different tokenization methods, we only report the toxicity score of the first complete sentence of a continuation as we found that the score returned by the Perspective API seems to increase with sentence length.

Results are shown in Figure~\ref{fig:rtp}. Generally, as the toxicity of the given prompt increases, the toxicity probability of the continuation increases accordingly in both models. Compared to GPT-3 Davinci, \glm has a lower toxicity rate in all cases, indicating that \glm is less prone to generating toxic content.

\begin{figure}[t]
    \centering
    \includegraphics[width=\linewidth]{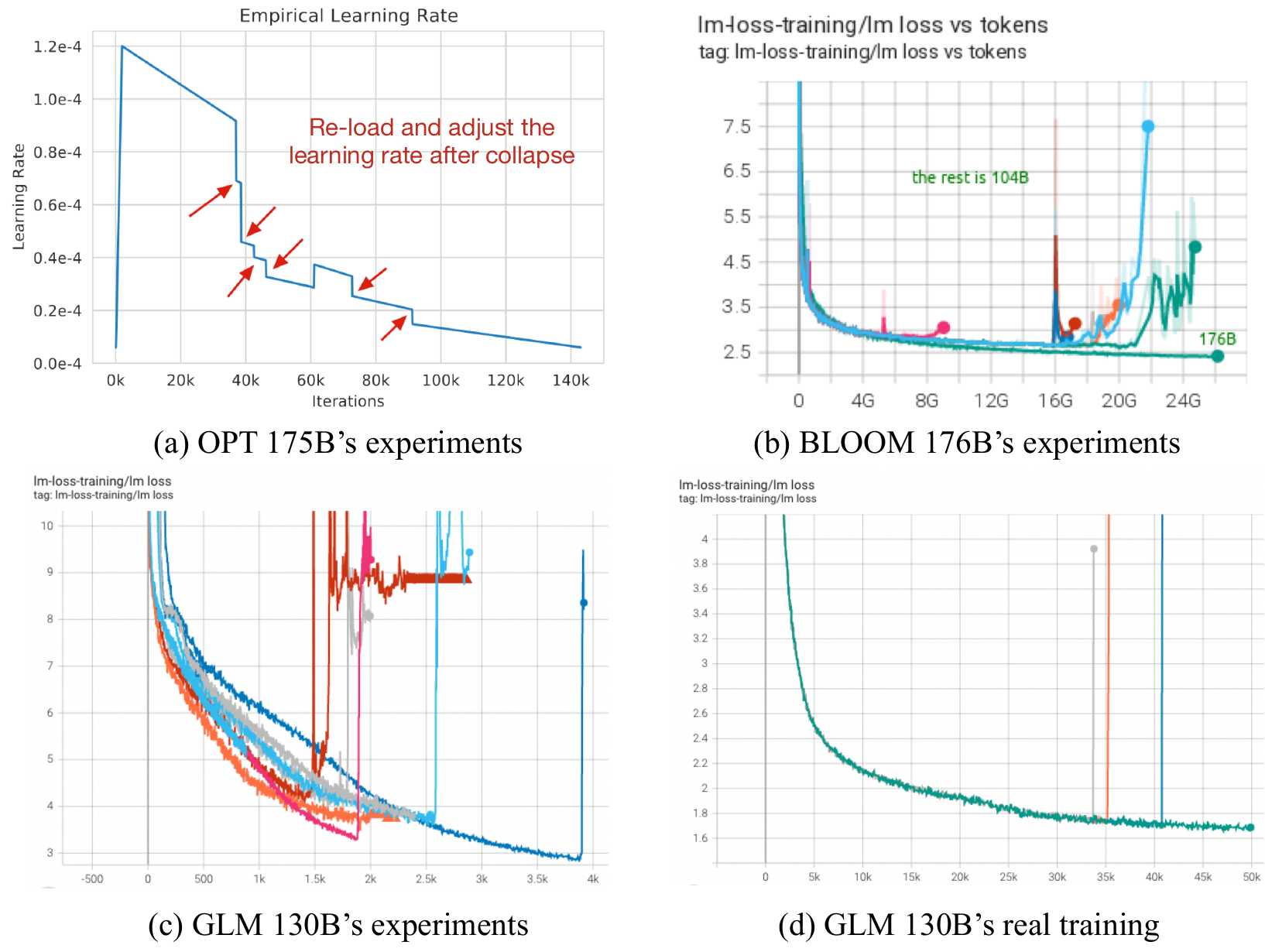}
    \caption{Handling training collapses and instability is the first priority when training LLMs.}
    \label{fig:collapse}
\end{figure}

\section{Technical Details}

In this section, we introduce additional details about the technical issues we have identified and solved throughout the \glm training.
Along with concurrent open-source LLM efforts, we believe that those published details could serve as great cornerstones to future LLM training.

\subsection{Tokenization}
For the tokenization of the corpus, we implement a text tokenizer based on the package \textit{icetk} with several adjustments. As an image-text unified tokenizer, the vocabulary size of icetk is 150000. The first 20000 tokens are image tokens and the rest are text tokens. The text tokenizer of icetk is formulated and trained by sentencepiece\footnote{\url{https://github.com/google/sentencepiece}}, on a 25GB bilingual corpus equally distributed with English and Chinese contents. We divide tokens recognized by the tokenizer into four categories. The common tokens are assigned from No.20000 to No.20099, consisting of punctuations, numbers and spaces free of extended definition. No.20100 to No.83822 are English tokens and No.83823 to No.145653 are Chinese tokens. Tokens after No.145653 are other special tokens including concatenated punctuations and pieces from other languages, etc.

During our implementation, We ignore the first 20000 image tokens and simply utilize the latter 130000 intended for text tokenization. we disable the ignoring of linebreak to tokenize the linebreak mark 
\texttt{\textbackslash n} 
into No. 20004 token \texttt{<n>}. On the basis of inherent tokens, we add special tokens \texttt{[MASK]} and \texttt{[gMASK]} for model prediction. We also add special tokens \texttt{<sop>}, \texttt{<eop>}, \texttt{<eos>} for sentence and passage separation.

\subsection{Layer Normalization} \label{app:ln}
Here we briefly introduce the history of layer normalization in language modeling problems, and how its variants perform in recent LLMs including our experiments for them on \glm.

\vvpara{Post-LN~\citep{vaswani2017attention}.}
Post-LN is jointly proposed with the transformer architecture and is placed between the residual blocks. 
It is then adopted by BERT~\citep{devlin2019bert} for bidirectional language model pre-training.
Nevertheless, Post-LN was later accused of transformers' slow and vulnerable converging~\citep{xiong2020layer} and the Pre-LN emerged as a substitute.

\vvpara{Pre-LN~\citep{xiong2020layer}.}
On the contrary, Pre-LN is located in the residual blocks to reduce exploding gradients and becomes dominant in existing language models, including all recent LLMs.
However, OPT-175B~\citep{zhang2022opt}, BLOOM~\citep{scao2022what}, and text-to-image model CogView~\cite{ding2021cogview} later observe that Pre-LN is still unable to handle the vulnerable training when models scale up to 100B or meet multi-modal data.
This is also justified in \glm's preliminary experiments, where Pre-LN consistently crashes in its early stage training.

Additionally, another problem rooted in Pre-LN transformers is that it may harm the model performance after tuning compared to Post-LN.
This is observed in~\citep{he2021realformer}.

\vvpara{Sandwich-LN~\citep{ding2021cogview}.}
As a remedy, on top of Pre-LN, CogView (later in Normformer~\citep{shleifer2021normformer}) develops Sandwich-LN which appends extra normalization to the end of each residual branch.
Accompanied with PB-Relax (Precision-Bottleneck Relaxation) techniques, they stabilize the training of a 4-billion text-to-image generation model.
Despite its superiority over Pre-LN, sadly Sandwich-LN is also proved to collapse in \glm training; let alone the potential consequent weaker tuning performance caused by its Pre-LN nature.

\subsection{Positional Encoding and Feed-forward Network} \label{app:pe-ffn}

\paragraph{Positional Encoding} Vanilla transformer adopts absolute (or sinuous) position encoding, and is later evolved into relative positional encoding~\citep{dai2019transformer}.
Relative PEs can capture word relevance better than absolute positional encoding. 
Rotary Positional Embedding (RoPE) \citep{su2021roformer} is a relative position encoding implemented in the form of absolute position encoding, and its core idea is shown in the following equation.
\begin{equation}
(\boldsymbol{R}_m q)^{\top}(\boldsymbol{R}_n k) = q^{\top} \boldsymbol{R}_m^{\top}\boldsymbol{R}_n k = q^{\top} \boldsymbol{R}_{n-m} k
\end{equation}

The product of $q$ at position $m$ and $k$ at position $n$ is related to their distance $n - m$, which reflects the relativity of the position encoding. The definition of $\boldsymbol{R}$ in the above equation is
\begin{equation}
\boldsymbol{R}_{\theta, m}^{d}=\left(\begin{array}{ccccccc}
\cos m \theta_{1} & -\sin m \theta_{1} & 0 & 0 & \cdots & 0 & 0 \\
\sin m \theta_{1} & \cos m \theta_{1} & 0 & 0 & \cdots & 0 & 0 \\
0 & 0 & \cos m \theta_{2} & -\sin m \theta_{2} & \cdots & 0 & 0 \\\
0 & 0 & \sin m \theta_{2} & \cos m \theta_{2} & \cdots & 0 & 0 \\\
\vdots & \vdots & \vdots & \vdots & \ddots & \vdots & \vdots \\
0 & 0 & 0 & 0 & \cdots & \cos m \theta_{d / 2} & -\sin m \theta_{d / 2} \\
0 & 0 & 0 & 0 & \cdots & \sin m \theta_{d / 2} & \cos m \theta_{d / 2}
\end{array}\right)
\end{equation}

To allow its value to decay as the distance increases, $\theta$ takes the value
\begin{equation}
\theta = \left\{\theta_i = 10000^{\frac{-2(i-1)} d},\quad i \in \left[1, 2, \cdots, \frac d 2\right]\right\}
\end{equation}

A two-dimensional absolute position encoding method is proposed in vanilla GLM for modeling both intra- and inter-span position information.
In \glm, different from the two-dimensional positional encoding used in vanilla GLM, we turn back to conventional one-dimensional positional encoding.
However, we originally thought that two-dimensional form cannot be directly applied to RoPE\footnote{We later found the instructions to implement two-dimensional RoPE from its author's blog \url{https://kexue.fm/archives/8397}, but our training has proceeded for weeks.}.
As a substitute plan, in \glm we simply remove the second dimension used in the original GLM as we find that the unidirectional attention mask sub-matrices for [MASK] generation indicate the token order as well.
This observation results in our transforming \glm's positional encoding into a one-dimensional one according to the following strategies:

\begin{itemize}[leftmargin=*,itemsep=0pt,parsep=0.2em,topsep=0.0em,partopsep=0.0em]
    \item For sequences corrupted by short spans, we discard the second-dimensional position encoding.
    \item For sequences corrupted by a long span at the end, we change the positional ids to one-dimensional $0, 1, \cdots, s - 1$, and generated tokens will just prolong the first-dimensional positional encoding from the last context token $s - 1$.
\end{itemize}

\paragraph{Feed-forward Network}
Some recent efforts to improve transformer architecture have been on the FFN, including replacing it with GLU (adopted in PaLM). Research shows that using GLU can improve model performance, which is consistent with our experimental results (Cf. Table~\ref{tab:pe-ffn-ablation}). Specifically, we use GLU with the GeLU~\citep{hendrycks2016gaussian} activation. as
\begin{equation}
\operatorname{FFN}_{\mathrm{GeGLU}}\left(\boldsymbol{x}; \boldsymbol{W}_1, \boldsymbol{V}, \boldsymbol{W}_{2}\right)=\left(\mathrm{GeLU}(\boldsymbol{x} \boldsymbol{W}_1) \otimes \boldsymbol{x} \boldsymbol{V}\right) \boldsymbol{W}_{2}
\end{equation}

In order to keep the same parameter as the vanilla FFN, the feed-forward size $d_{\mathrm{ffn}}$ (which is usually $4 d_{\mathrm{H}}$, where $d_{\mathrm{H}}$ is the hidden dimension) is reduced to $\frac 8 3 d_{\mathrm{H}}$ as the $\boldsymbol{V}$ is additionally introduced.

\begin{wraptable}{r}{4.5cm}
	\centering
	\footnotesize
	\vspace{-4mm}
    \renewcommand\tabcolsep{2pt}
    \renewcommand\arraystretch{1.1}
	\caption{Ablation Study for PE and FFN on GLM\textsubscript{Base}}
	\vspace{-1mm}
	\scalebox{1.00}{
	\label{tab:pe-ffn-ablation}
	\begin{tabular}{@{}lc@{}}
    \toprule[1.2pt]
    Model           & Test PPL        \\ \midrule
    GLM\textsubscript{Base}        & 24.58 \\
    \quad + ALiBi         & 24.14 \\
    \quad + RoPE          & 22.95 \\
    \quad + RoPE + GeGLU  & \textbf{22.31} \\
    \bottomrule[1.2pt]
    \end{tabular}
    }
\end{wraptable}

\paragraph{Ablation Study on PE and FFN} In order to validate our PE and FFN choices, we test them in our experiments by pre-training GLM\textsubscript{Base} (110M) over a random 50G Chinese and English mixed corpus. We compare absolute PE with two recent popular relative PE variants, RoPE~\citep{chowdhery2022palm} and ALiBi~\citep{press2021train}. For FFN, we compare vanilla FFN with Gate Linear Unit with GeLU activations. Results from Table \ref{tab:pe-ffn-ablation} show that both ALiBi and RoPE improve perplexity on the test set, and the improvement is more significant with RoPE while using GeGLU can further improve the model's performance.

\subsection{Pipeline Parallel Analysis} \label{app:pipeline}

\begin{figure}[t] 
\centering    
\subfigure[Naive pipeline implementation, which can be extremely inefficient.] {
 \label{fig:pipeline-naive}
\vspace{-4mm}
\includegraphics[width=\linewidth]{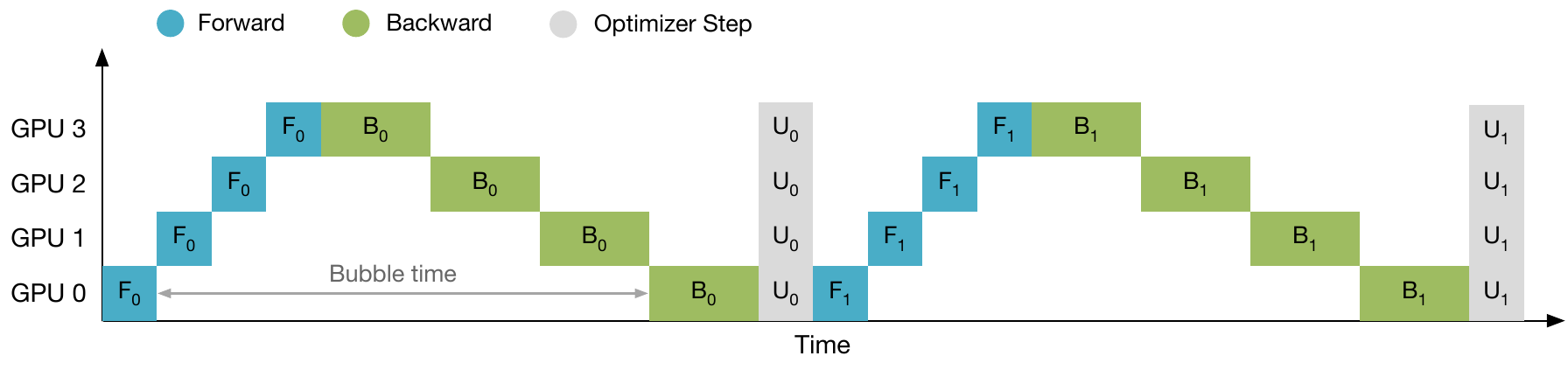}  
}     
\subfigure[GPipe~\citep{huang2019gpipe} implementation.] { 
\label{fig:gpipe}
\includegraphics[width=\linewidth]{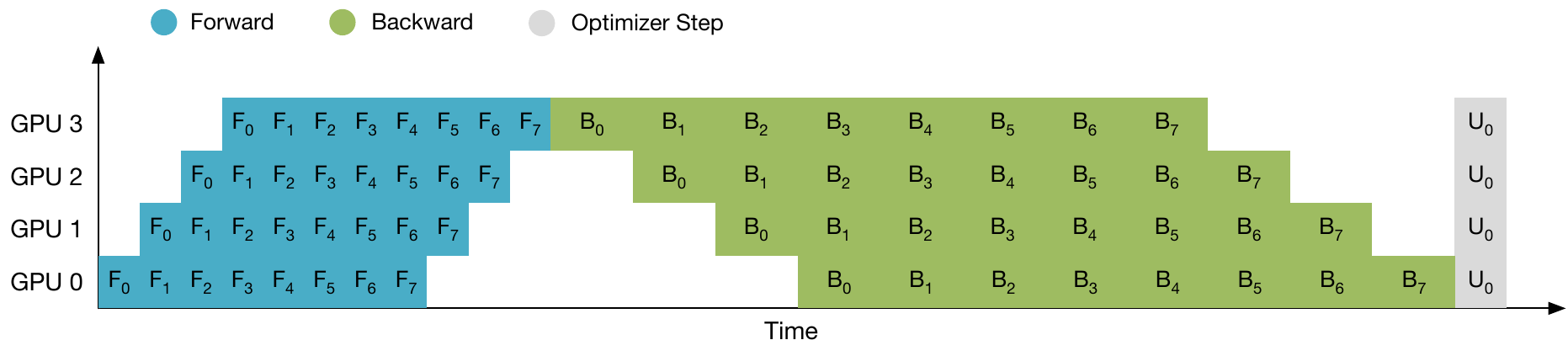} 
}    
\subfigure[Pipedream~\citep{narayanan2021memory} implementation (used in \glm).] { 
\label{fig:pipedream}
\vspace{-4mm}
\includegraphics[width=\linewidth]{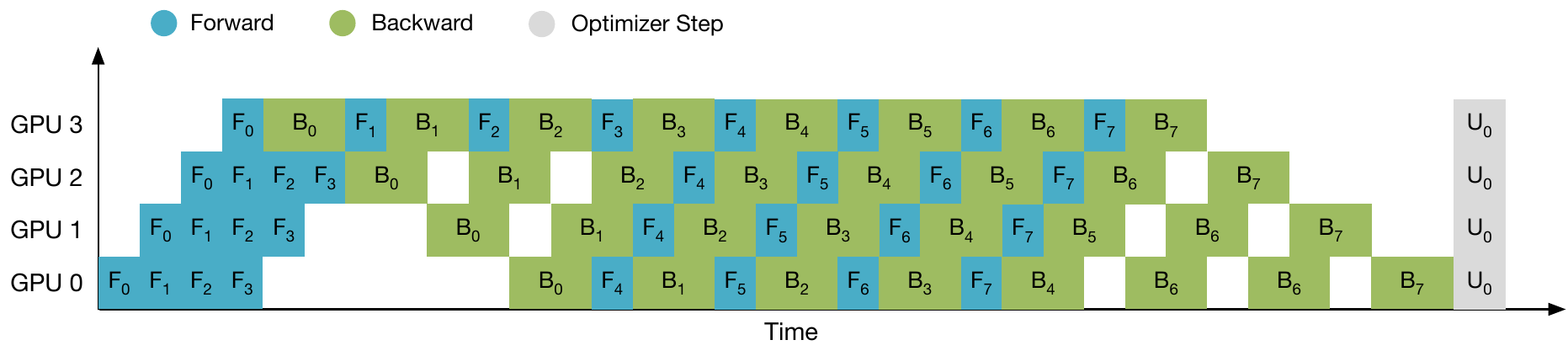} 
}
\vspace{-4mm}
\caption{Different pipeline strategies and their conceptual comparison.}     
\label{fig}     
\end{figure}

In pipeline parallelism, each stage consists of three operations (Cf. Figure~\ref{fig:pipeline-naive}): forward (denoted as F), backward (denoted as B), and optimizer step (denoted as U).
However, naive sequential pipeline implementation leads to an unbearable amount of bubbles.
The improved Gpipe~\citep{huang2019gpipe} (Cf. Figure~\ref{fig:gpipe}) strategy reduces bubbles drastically via splitting data into micro-batches; the more micro-batches there are, the more stages can compute simultaneously in an iteration. 
The recent PipeDream-Flush~\citep{narayanan2021memory} (Cf. Figure~\ref{fig:pipedream}) additionally optimizes the GPU memory usage by interweaving forward and backward from different stages to reduce forward activation's memory occupation.

We analyze the bubble share in \glm's pre-training by assuming that the number of pipeline segments is $p$, the number of micro-batches is $m$, and the time for forward and backward per micro-batch are $t_f$ and $t_b$. 
In ideal case, forward and backward take $t_{\mathrm{ideal}} = m(t_f + t_b)$. 
But in practice, the default pipeline delivery strategy causes $p - 1$ forward propagation and $p - 1$ backward propagation bubbles, respectively, for a total time of $t_{\mathrm{bubble}} = (p - 1)( t_f + t_b)$, so that the bubble occupancy is
\begin{equation}
    \text{bubble-ratio} = \frac {t_{\mathrm{bubble}}} {t_{\mathrm{ideal}} + t_{\mathrm{bubble}}} = \frac {p - 1} {m + p - 1}
\end{equation}

For larger numbers of micro-batches, the bubble percentage will be reduced to an acceptable level. In particular, experiments in GPipe \cite{huang2019gpipe} show that when $m \ge 4p$, the total percentage of pipeline bubble time is reduced to a negligible level due to the forward recomputation technique in backpropagation that allows some overlap in computational communication, thus showing that the bubbles introduced in parallel by the pipeline model do not seriously deplete the training efficiency. 

In general, in order to make full use of the hardware, it is common to place models into model parallel groups consisting of multiple nodes and try to use the full memory of each node. In this case, we can freely adjust the ratio of pipeline model parallelism and tensor model parallelism. Since data parallelism hardly affects the computation time, we assume that the scale of data parallelism is $d = 1$, the total number of nodes is $n$, the scale of tensor model parallelism is $t$, and the scale of pipeline model parallelism is $p$, and satisfies $n = t\times p$, the bubble share in this case is
\begin{equation}
    \text{bubble-ratio} = \frac {n/t - 1} {m + n/t - 1} 
\end{equation}

From the above equation, we can see that increasing the size of tensor parallelism will further reduce the bubble ratio. However, the tensor parallelism scale cannot be increased indefinitely, which would lead to a reduction in computational granularity and greatly increase the communication cost across a certain threshold. Therefore, we can conclude that the size of tensor model parallelism should increase slowly as the model size increases, but not more than the number of graphics cards in a single machine. In the training of \glm, the experiments show that the optimal tensor parallelism scale is $t = 4$ and does not scale up to the scale of $t = 8$ in the DGX-A100 system. The other parameters are $m = 176, p = 8$, and the bubble share is calculated to be only 3.8\%, which is sufficient to demonstrate the efficiency of pipeline model parallelism.

\subsection{Inference Acceleration} \label{app:inference_acceleration}


\begin{table}[t]
    \footnotesize
    \centering
    \caption{Decoding speed in our real trials between BLOOM-176B~\citep{scao2022what} (from Huggingface Transformers) and \glm's implementation in 16-bit precision with 8 $\times$ A100 (80G).}
    \vspace{-1mm}
    \begin{threeparttable}
    \centering
    \renewcommand\tabcolsep{8pt}
    \begin{tabular}{@{}lcccc@{}}
    \toprule[1.2pt]
    Decode Tokens & 128                                  & 512                                   & 1024                                  & 2048                                  \\ \midrule
    BLOOM-176B    & 36.76s                               & 137.91s                               & 287.93s                               & 631.81s                               \\
    \glm          & 4.40s \textcolor{red}{($\times$8.4)} & 18.77s \textcolor{red}{($\times$7.3)} & 39.81s \textcolor{red}{($\times$7.2)} & 89.88s \textcolor{red}{($\times$7.0)} \\ \bottomrule[1.2pt]
    \end{tabular}
    \end{threeparttable}
    \label{tab:fastertransformer}
    \vspace{-3mm}
\end{table}
A model's plain PyTorch implementation is easy to read and run, but it can be intolerably slow for LLMs. 
Based on NVIDIA's FasterTransformer\footnote{\url{https://github.com/NVIDIA/FasterTransformer}} we spend two months implementing \glm into C++ to speed up inference, including the following main optimizations: 
\begin{itemize}[leftmargin=*,itemsep=0pt,parsep=0.2em,topsep=0.0em,partopsep=0.0em]
    \item Optimize time-costing operations such as GeGLU, Layer Normalization, and SoftMax.
    \item Reduce the number of GPU kernel calls (e.g., fuse MultiheadAttention into one computation kernel).
    \item Specify the algorithm of the best performance when calling cuBLAS.
    \item Improve the computing efficiency by transposing the model parameters in advance.
    \item Use half2 in FP16 computation to double the half's access bandwidth and computing throughput.
\end{itemize}

We currently pack up the full FasterTransformer implementation for \glm into a plug-and-play docker image for users' convenience, and we are still working on adapting it to our Pytorch implementation
by only changing one line of code.
A comparison between our speeding up \glm implementation and the so far default available BLOOM-176B implementation in Huggingface Transformers\footnote{\url{https://huggingface.co/docs/transformers/model_doc/bloom}} is shown in Table~\ref{tab:fastertransformer}.
Our implementation for \glm can be 7.0 to 8.4 times faster than BLOOM-176B's Pytorch implementation.
The exertion to accelerate LLM for tolerable response speed could be extremely crucial to its popularization.

\begin{figure}[t]
    \centering
    \includegraphics[width=\linewidth]{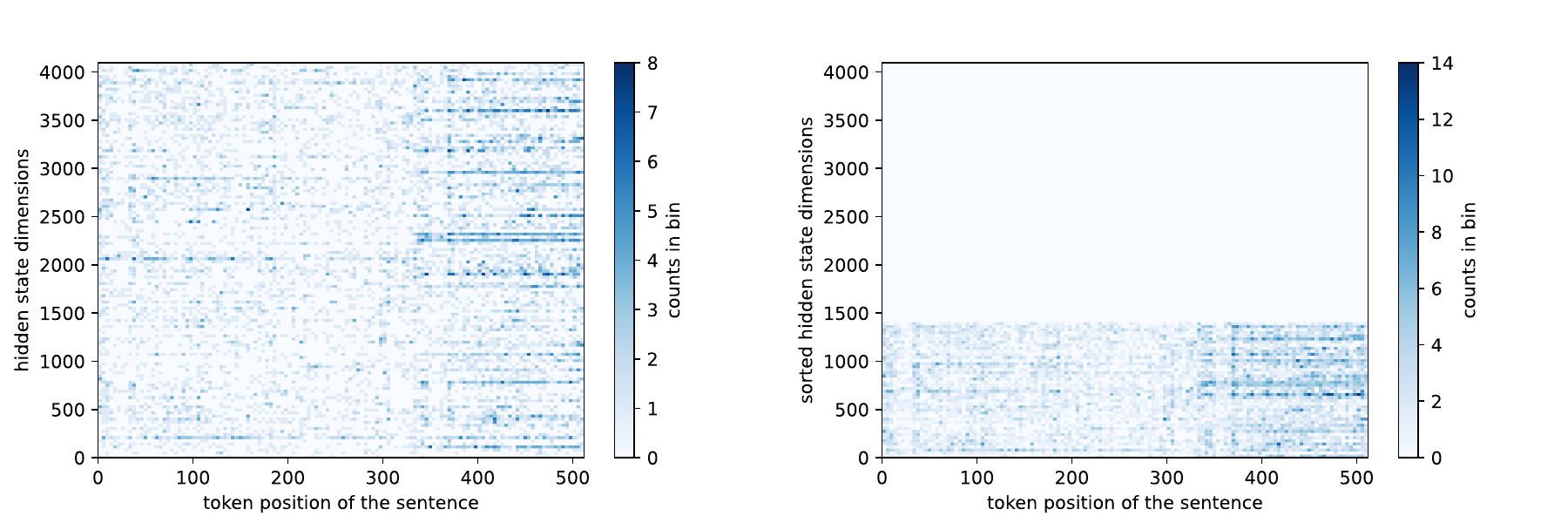}
    \caption{Distribution of outliers
    in \glm's activations. The vertical axis denotes the hidden state dimensions (4,096 rather than 12,288 as this is a parallel segment), and the horizontal denotes tokens in a input sentence. Using a 128$\times$128 2D histogram to get a better view of the distribution of outliers. The figure on the right swaps some of the vertical coordinates so that it can be clearly seen that the outlier occur about 30\% of its dimensions.
    }
    \label{fig:activation_outliers}
    \vspace{-4mm}
\end{figure}

\subsection{Activation Outlier Analysis} \label{app:activation_analysis}
As is described in prior sections, \glm's weight can be quantized into INT4 to drastically cut down parameter redundancy in the inference.
However, we also find that \glm's activations (i.e., hidden states between layers) cannot be properly quantized, as they contain value outliers as is also suggested in concurrent literature~\citep{dettmers2022llm}.

What is special in \glm is that 30\% of its dimensions may present value outliers (Cf. Figure~\ref{fig:activation_outliers}), while other GPT-based LLMs (e.g., OPT-175B and BLOOM 176B) only has very few outlying dimensions~\citep{dettmers2022llm}.
Therefore, the solution to decompose matrix multiplication for higher-precision computation in outlying dimensions proposed in~\citep{dettmers2022llm} is not applicable to \glm.

\begin{wrapfigure}{r}{4cm}
    \small
    \vspace{-10mm}
    \centering
    \includegraphics[width=1.0\linewidth]{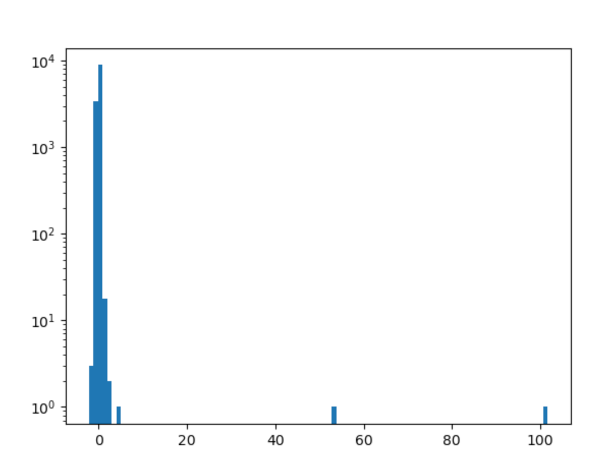}
    \vspace{-6mm}
    \caption{\glm's activation outliers' absolute value scale.}
    \label{fig:outlier_scale}
    \vspace{-4mm}
\end{wrapfigure}

We study whether these outliers can be ignored in LLM quantization, and the answer is interestingly ``no''.
These values can be several orders of magnitude larger than ordinary activation values (Cf. Figure~\ref{fig:outlier_scale}).
While most values (accounts for 99.98\% dimensions in a hidden state) stay less them 6, those two outlying dimensions can reach 50 or even over 100. 
They are speculated to be some important clues for \glm and potentially other LLMs to memorize some fixed world or language knowledge, and thus removing or omitting them in quantization can lead to significant performance degradation.

\subsection{Weight Quantization} \label{app:quantization_analysis}

\subsubsection{Preliminaries}

\paragraph{Absmax Quantization} is a symmetric quantization that a range of $[-\mathrm{absmax}(x), \mathrm{absmax}(x)]$ is mapped to $[-(2^b - 1), 2^b - 1]$ for $x$.
\begin{align}
    s_{x} &= \frac {\mathrm{absmax}(x)} {2^{b - 1} - 1} \\
    x_{q} &= \mathrm{round}(x / s_{x})
\end{align}
where $s_x$ is the scaling factor, $x_q$ is the quantization result and $b$ is the bit width.
\vspace{-0.5em}
\paragraph{Zeropoint Quantization} is an asymmetric quantization that a range of $[\min(x), \max(x)]$ is mapped to $[-(2^b - 1), 2^b - 1]$.
\begin{align}
    s_{x} &= \frac {\max(x) - \min(x)} {2^b - 2} \\
    z_{x} &= \mathrm{round}(\min(x) / s_{x}) + 2^{b - 1} - 1\\
    x_{q} &= \mathrm{round}(x / s_{x}) - z_{x}
\end{align}
where $z_x$ is the zero point.
\vspace{-0.5em}
\paragraph{Col/Row-wise Quantization} Using a single scaling factor for the weight matrix often leads to more quantization errors because one single outlier leads to a decrease in the quantization precision of all other elements. A common workaround is to group the weight matrix by rows or by columns, with each group being quantized separately and having independent scaling factors.


\subsection{Quantization settings}

Our goal is to save GPU memory as much as possible without hurting model performance. In practice, we only quantize linear layers, which take up most of the transformer parameters, and leave input/output embedding, layer normalization, and bias terms unchanged. At the quantization precision of INT4, two INT4 weights are compressed into one INT8 weight for saving GPU memory usage. Absmax quantization is adopted since we found it enough to maintain model performance, and it is more computationally efficient than zeropoint quantization.
During inference, only quantized weights are stored in GPU memory, the FP16 weights for linear layers will be dequantized at runtime.

\subsubsection{Quantization Results at Scales}

\begin{table}[t]
    \footnotesize
    \centering
    \caption{Accuracy on LAMBADA dataset for GLM and BLOOM family at 100M to 176B scales across different quantization precision.}
    \vspace{-1mm}
    \begin{threeparttable}
    \centering
    \renewcommand\tabcolsep{2.5pt}
    \begin{tabular}{@{}lccccc@{}}
    \toprule

                         & \textbf{BLOOM-560M} & \textbf{BLOOM-1B1}     & \textbf{BLOOM-3B} & \textbf{BLOOM-7B} & \textbf{BLOOM-176B} \\
            \midrule
Original                 & 31.40\%             & 40.68\%                & 48.30\%           & 54.91\%           & 64.37\%             \\
Absmax INT8, col-wise    & 26.12\%             & 40.69\%                & 48.83\%           & 55.33\%           & 65.03\%             \\
Absmax INT4, col-wise    & 9.30\%              & 17.43\%                & 37.88\%           & 38.04\%           & 34.83\%             \\
Absmax INT4, row-wise    & 21.37\%             & 35.80\%                & 40.95\%           & 46.75\%           & NaN                 \\
Zeropoint INT4, col-wise & 11.51\%             & 26.51\%                & 41.65\%           & 46.63\%           & 48.26\%             \\
Zeropoint INT4, row-wise & 24.95\%             & 33.05\%                & 43.63\%           & 49.41\%           & NaN                 \\ \midrule
                         & \textbf{GLM-110M}   & \textbf{GLM-335M}      & \textbf{GLM-2B}   & \textbf{GLM-10B}  & \textbf{GLM-130B}   \\ \midrule
Original                 & 29.36\%             & 48.51\%                & 68.19\%           & 72.35\%           & 80.21\%             \\
Absmax INT8, row-wise    & 29.25\%             & 48.69\%                & 68.12\%           & 72.37\%           & 80.21\%             \\
Absmax INT4, row-wise    & 3.26\%              & 38.25\%                & 62.62\%           & 71.03\%           & 79.47\%             \\
Zeropoint INT4, row-wise & 5.45\%              & 42.64\%                & 64.74\%           & 70.50\%           & 80.63\%             \\ \bottomrule
    \end{tabular}
    \end{threeparttable}
    \label{tab:weight-quantization}
\end{table}

GLM models at 110M to 10B scale are from GLM's original paper\citep{du2022glm}. Although the architecture of smaller scale GLMs are not the same as \glm, we believe that the training objective is the key factor for quantization. Table~\ref{tab:weight-quantization} shows the performance of GLM and BLOOM family models at different scales on the LAMBADA dataset with different quantization methods. Almost all models maintain performance at INT8 precision. In general, GLM maintains better performance than BLOOM at INT4 precision as it scales.


\subsubsection{Weight Distribution Analysis}

\begin{figure}[t]
    \centering
    \includegraphics[width=0.88\linewidth]{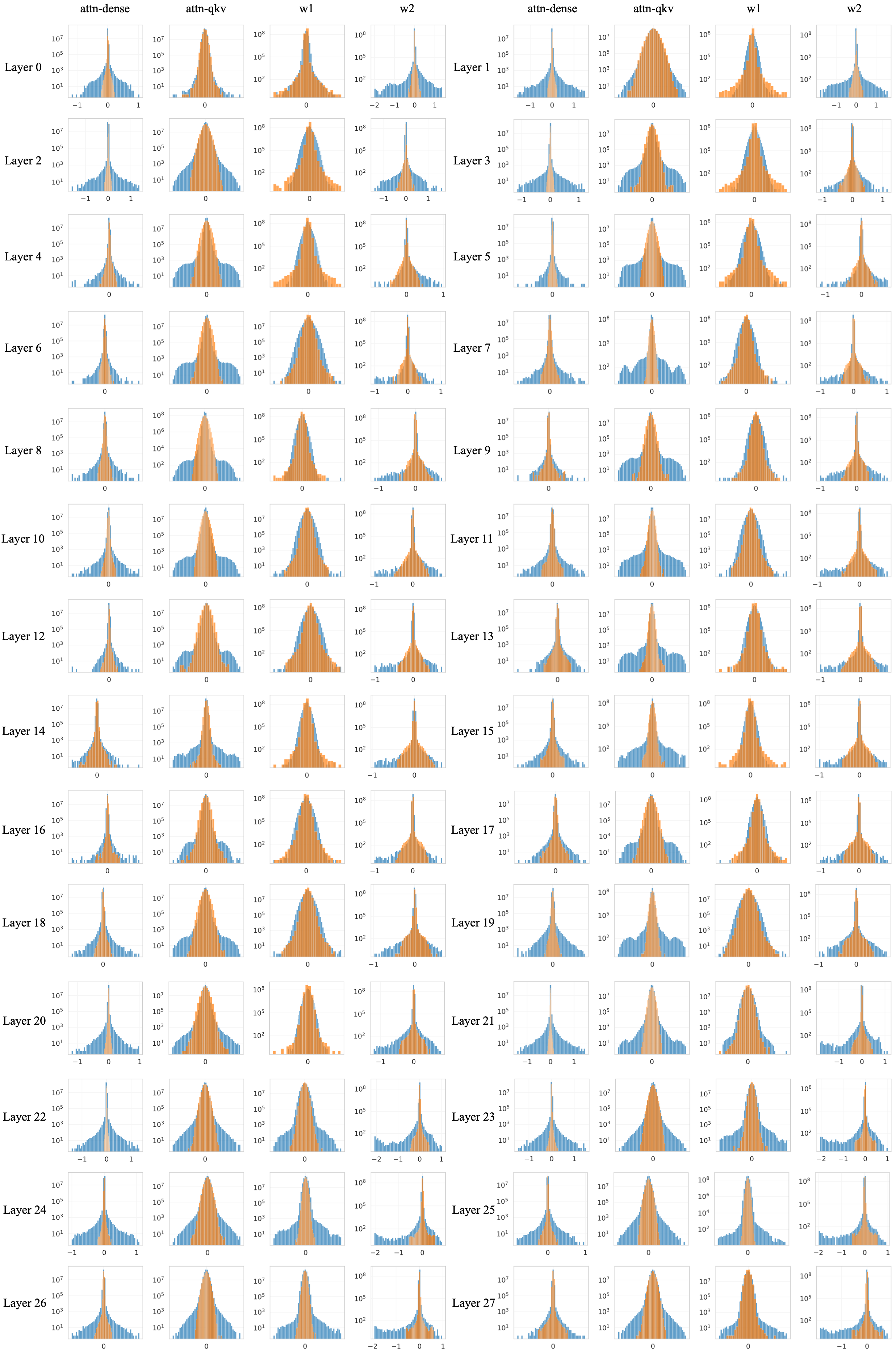}
    \caption{Weight value distribution of linear layers in \glm (in orange, \texttt{attn-dense}, \texttt{attn-qkv}, \texttt{glu-w1}, \texttt{glu-w2}) and BLOOM-176B (in blue, \texttt{attn-dense}, \texttt{attn-qkv}, \texttt{ffn-w1}, \texttt{ffn-w2})'s first 28 transformer layers. Generally for \glm it is \texttt{attn-dense} and \texttt{w2} that may present narrow value distributions. \texttt{attn-qkv} and \texttt{w1} may also be a reason for enabling INT4 quantization in middle layers of \glm.}
    \label{fig:quantization_appendix}
\end{figure}

To achieve INT4 weight quantization, we analyze the weight value distribution of major linear layers in \glm and a counterpart BLOOM-176B in a histogram (Cf. Figure~\ref{fig:quantization_appendix}).
The horizontal axis denotes the weight value, and the vertical axis denotes the number of weights of such value in log scale.
As we can see, it is majorly the \texttt{w2} linear layers in BLOOM-176B that present skewed distributions, which would hinder the symmetrical quantization.
On the contrary, \glm's \texttt{w2} is well-shaped without many outliers and skewed distribution, and thus paces the way for its INT4 quantization with little performance loss.

\subsection{Ablation on Contribution Attribution}
\begin{figure}[t]
    \centering
    \includegraphics[width=\linewidth]{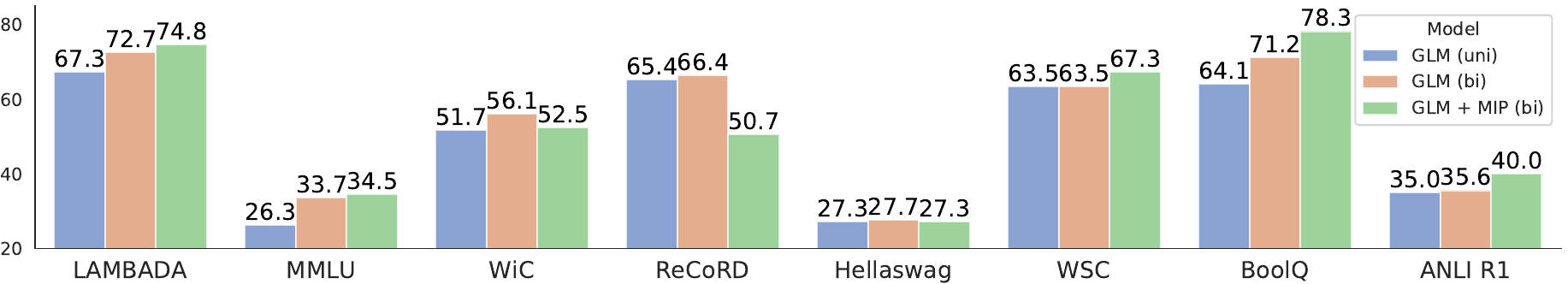}
    \caption{Contribution attribution analysis on GLM objective and MIP training. We take GLM-10B (English only) as an example in the ablation. Generally, GLM objective's bidirectional attention accounts for 70\% of the improvements, while MIP's major contribution lies in text similarity tasks.}
    \label{fig:ablation_analysis}
    \vspace{-5mm}
\end{figure}

We analyze the contribution attribution of techniques leveraged in \glm.
A series of ablation studies have been presented in the paper, and for the convenience of reading, they were originally scattered around the whole passage.
Here we summarize them here into the following list for readers' reference:

\begin{itemize}[leftmargin=*,itemsep=0pt,parsep=0.2em,topsep=0.0em,partopsep=0.0em]
    \item \textbf{Ablation on ordinary PostLN and DeepNorm}: Figure~\ref{fig:layernorm}.
    \item \textbf{Ablation on Bidirectional/Unidirectional Attention}: Figure~\ref{fig:lambada} (LAMBADA), Table~\ref{tab:nlg} (Conditional NLG), Figure~\ref{fig:superglue} (SuperGLUE). 
    \item \textbf{Ablation on Embedding Layer Gradient Shrink (EGS)}: Figure~\ref{fig:shrink}.
    \item \textbf{Ablation on Positional Encodings and FFN}: Appendix~\ref{app:pe-ffn} Table~\ref{tab:pe-ffn-ablation}.
\end{itemize}

Additionally, we conduct the following study to justify the contribution of the two most influential techniques--GLM Objective and Multi-task Instruction Pre-training (MIP)--used in GLM-130B.

\vvpara{GLM Objective and MIP.}
Ablating a 100B-scale LLM from scratch can be too expensive.
As a substitute, we try our best to conduct the comparison between GLM objective and MIP on GLM-10B (an English-only version released in~\citep{du2022glm}, without MIP).
We additionally train a GLM-10B initialized from a middle-stage original checkpoint with MIP (5\%) to match the same training tokens of the original self-supervision-only GLM-130B.
The MIP, this time, follows the exact dataset setting in T0~\citep{sanh2022multitask} and the information extraction datasets in GLM-130B to allow the correct evaluation on some types of tasks (e.g., NLI).

Figure~\ref{fig:ablation_analysis} shows the ablation results.
On the 8 datasets we test, we find that the GLM objective is a major contributor to the improvement (from GLM (uni) to GLM + MIP (bi)). 
For example, it accounts for 73\% improvement in LAMBADA and 90\% improvement in MMLU, which are very widely adopted challenging benchmarks for LLMs.
As for MIP, on some datasets (e.g., WiC, ReCoRD, Hellaswag), MIP may even harm the performance.
While for datasets related to text similarity and coreference (e.g., WSC, BoolQ, ANLI R1), MIP is the main contributor.
It is likely because the text similarity and coreference challenges, which people usually construct intentionally to test language models' ability, are seldom seen in the self-supervised corpus that makes up people's daily written texts.
Thus, MIP training mainly helps to bridge the gap between self-supervised pre-training and these tasks.

\subsection{Lessons Learned} \label{sec:lessons}

\begin{insight}[\bf Bidirectional Architecture]
\rm The bidirectional-attention GLM is a strong  architecture alternative, in  addition to GPTs.
\end{insight}

\begin{insight}[\bf Platform-aware Configuration]
\rm Configure LLMs based on the cluster and parallel strategy used to squeeze hardware potential.
\end{insight}

\begin{insight}[\bf Improved Post-LN]
\rm Counter-stereotypically, DeepNorm, a type of Post-LN, is the option to stabilize \glm.
\end{insight}

\begin{insight}[\bf Training Stability Categorization]
\rm Unexpected training instability that LLMs suffer from arouses systematically and numerically.
\end{insight}

\begin{insight}[\bf Systematical Instability: FP16]
\rm Though FP16 induces more instability, it enables training and inference on diverse platforms.
\end{insight}

\begin{insight}[\bf Numerical Instability: Embedding Gradient Shrink]
\rm Shrinking embedding layer's gradient to its 0.1 can solve most numerical instability problems.
\end{insight}

\begin{insight}[\bf GLM's INT4 Quantization Scaling Law]
\rm GLM has a unique INT4 weight quantization scaling law unobserved in GPT-style BLOOM.
\end{insight}

\begin{insight}[\bf Future Direction]
\rm To create powerful LLMs, the main focus can be on 1) more and better data, 2) better architectures and pre-training objectives, and 3) more sufficient training.
\end{insight}

\clearpage 

\begin{table}[h]
\vspace{-2em}
\centering
\footnotesize
\caption{Full configurations for \glm training}
\label{tab:config}
\renewcommand\arraystretch{0.95}
\scalebox{0.85}{
\begin{tabular}{lr}
\toprule
Configuration Key                            & Value                        \\ \midrule
adam\_beta1                               & 0.9                          \\
adam\_beta2                               & 0.95                         \\
adam\_eps                                 & 1e-08                        \\
aggregated\_samples\_per\_sequence        & 4                            \\
attention\_dropout                        & 0.1                          \\
attention\_softmax\_in\_fp32              & True                         \\
average\_block\_length                    & 3                            \\
bias\_dropout\_fusion                     & True                         \\
checkpoint\_activations                   & True                         \\
checkpoint\_in\_cpu                       & False                        \\
checkpoint\_num\_layers                   & 1                            \\
clip\_grad                                & 1.0                          \\
contigious\_checkpointing                 & False                        \\
cpu\_optimizer                            & False                        \\
data\_parallel\_size                      & 24                           \\
deepnorm                                  & True                         \\
distributed\_backend                      & nccl                         \\
eval\_interval                            & 1000                         \\
eval\_iters                               & 3                            \\
ffn\_hidden\_size                         & 32768                        \\
fp16                                      & True                         \\
global\_batch\_size                       & 4224                         \\
glu\_activation                           & geglu                        \\
gpt\_prob                                 & 0.7                          \\
hidden\_dropout                           & 0.1                          \\
hidden\_size                              & 12288                        \\
hysteresis                                & 2                            \\
init\_method\_std                         & 0.0052                       \\
init\_method\_xavier\_uniform             & False                        \\
initial\_loss\_scale                      & 65536                        \\
layernorm\_epsilon                        & 1E-05                        \\
learnable\_rotary\_embedding              & False                        \\
length\_per\_sample                       & 2000                         \\
log\_interval                             & 1                            \\
loss\_scale                               & 0                            \\
loss\_scale\_window                       & 2000                         \\
lr                                        & 8e-05                        \\
lr\_decay\_iters                          & None                         \\
lr\_decay\_samples                        & 197753905                    \\
lr\_decay\_style                          & cosine                       \\
lr\_warmup\_samples                       & 1098632                      \\
make\_vocab\_size\_divisible\_by          & 768                          \\
mask\_prob                                & 0.15                         \\
masked\_softmax\_fusion                   & True                         \\
micro\_batch\_size                        & 1                            \\
min\_gmask\_ratio                         & 0.2                          \\
min\_loss\_scale                          & 1.0                          \\
min\_lr                                   & 8e-06                        \\
multitask\_ratio                          & 0.05                         \\
num\_attention\_heads                     & 96                           \\
num\_layers                               & 70                           \\
onnx\_safe                                & None                         \\
optimizer                                 & adam                         \\
partition\_activations                    & True                         \\
pipeline\_model\_parallel\_size           & 8                            \\
position\_embedding\_type                 & rotary                       \\
rampup\_batch\_size                       & 192, 24, 5493164             \\
save\_interval                            & 250                          \\
seed                                      & 1234                         \\
seq\_length                               & 2048                         \\
short\_seq\_prob                          & 0.02                         \\
shrink\_embedding\_gradient\_alpha        & 0.1                          \\
single\_span\_prob                        & 0.02                         \\
split                                     & 949,50,1                     \\
tensor\_model\_parallel\_size             & 4                            \\
tokenizer\_type                           & IceTokenizer                 \\
weight\_decay                             & 0.1                          \\
zero\_contigious\_gradients               & False                        \\
zero\_reduce\_bucket\_size                & 500000000                    \\
zero\_reduce\_scatter                     & False                        \\
zero\_stage                               & 1                            \\
zero-optimization.allgather\_bucket\_size & 500000000                    \\
tokenizer\_type                           & IceTokenizer                 \\
weight\_decay                             & 0.1                          \\
world\_size                               & 768                          \\
zero\_contigious\_gradients               & FALSE                        \\
zero\_reduce\_bucket\_size                & 500000000                    \\
zero\_reduce\_scatter                     & FALSE                        \\
zero\_stage                               & 1                            \\
zero-optimization.allgather\_bucket\_size & 500000000                    \\ 
\bottomrule
\end{tabular}
}
\end{table}

\begin{table}[t]
\caption{
The 74 datasets involved in Multi-task Instruction Pre-training (MIP). 
Datasets from T0-PromptSource~\citep{sanh2022multitask,bach2022promptsource} are named in their Hugging Face datasets identifiers.
Datasets from DeepStruct~\citep{wang2022deepstruct} are described in Appendix~\ref{app:mip_dataset}.
}
\vspace{2mm}
\renewcommand\tabcolsep{2pt}
\renewcommand\arraystretch{1.1}
\scalebox{0.94}{
\begin{tabular}{@{}llll@{}}
\toprule[1.2pt]
Task                       & Dataset                   & Task                      & Dataset             \\ \midrule
Coreference Resolution     & super\_glue/wsc.fixed     & Multi-choice QA           & cos\_e/v1.11        \\
Coreference Resolution     & winogrande/winogrande\_xl & Multi-choice QA           & cosmos\_qa          \\
Natural Language Inference & super\_glue/cb            & Multi-choice QA           & dream               \\
Natural Language Inference & super\_glue/rte           & Multi-choice QA           & openbookqa/main     \\
Natural Language Inference & anli                      & Multi-choice QA           & qasc                \\
Paraphrase Identification  & glue/mrpc                 & Multi-choice QA           & quail               \\
Paraphrase Identification  & glue/qqp                  & Multi-choice QA           & quarel              \\
Paraphrase Identification  & paws/labeled\_final       & Multi-choice QA           & quartz              \\
Closed-Book QA             & ai2\_arc/ARC\_Challenge   & Multi-choice QA           & race/high           \\
Closed-Book QA             & ai2\_arc/ARC\_Easy        & Multi-choice QA           & race/middle         \\
Closed-Book QA             & kilt\_tasks/hoptpotqa     & Multi-choice QA           & sciq                \\
Closed-Book QA             & trivia\_qa/unfiltered     & Multi-choice QA           & social\_i\_qa       \\
Closed-Book QA             & web\_questions            & Multi-choice QA           & super\_glue/boolq   \\
Closed-Book QA             & wiki\_qa                  & Multi-choice QA           & super\_glue/multirc \\
Extractive QA              & adversarial\_qa/dbidaf    & Multi-choice QA           & wiki\_hop/original  \\
Extractive QA              & adversarial\_qa/dbert     & Multi-choice QA           & wiqa                \\
Extractive QA              & adversarial\_qa/droberta  & Multi-choice QA           & piqa                \\
Extractive QA              & duorc/SelfRC              & Topic Classification      & ag\_news            \\
Extractive QA              & duorc/ParaphraseRC        & Topic Classification      & dbpedia\_14         \\
Extractive QA              & ropes                     & Topic Classification      & trec                \\
Extractive QA              & squad\_v2                 & Word Sense Disambiguation & super\_glue/wic     \\
Extractive QA              & super\_glue/record        & Dialogue State Tracking   & multiwoz\_2.1       \\
Extractive QA              & quoref                    & Event Extraction          & ace05               \\
Sentiment                  & amazon\_polarity          & Named Entity Recognition  & conll03             \\
Sentiment                  & app\_reviews              & Named Entity Recognition  & genia               \\
Sentiment                  & imdb                      & Named Entity Recognition  & ontonotes5.0        \\
Sentiment                  & rotten\_tomatoes          & Named Entity Recognition  & ace2005             \\
Sentiment                  & yelp\_review\_full        & Named Entity Recognition  & conll04             \\
Sentence Completion        & super\_glue/copa          & Named Entity Recognition  & nyt29               \\
Sentence Completion        & hellaswag                 & Relation Extraction       & conll04             \\
Structure-to-Text          & common\_gen               & Relation Extraction       & nyt29               \\
Structure-to-Text          & wiki\_bio                 & Relation Extraction       & ace2005             \\
Summarization              & cnn\_dailymail/3.0.0      & Relation Extraction       & kelm                \\
Summarization              & gigaword                  & Relation Classification   & tacred              \\
Summarization              & multi\_news               & Semantic Role Labeling    & conll05             \\
Summarization              & samsum                    & Semantic Role Labeling    & conll12             \\
Summarization              & xsum                      & Semantic Role Labeling    & propbank            \\ \bottomrule[1.2pt]
\end{tabular}}
\label{tab:mip}
\vspace{-4mm}
\end{table}
\section{Dataset and Evaluation Details} \label{app:data}
\subsection{Multi-task Instruction Pre-training (MIP)} \label{app:mip_description}
Following practices in~\citep{raffel2020exploring,wei2022finetuned,sanh2022multitask,aribandi2022ext5}, we include a number of prompted instruction datasets in \glm's MIP training, which accounts for 5\% of the training tokens.
All prompts for T0 datasets are from PromptSource~\citep{bach2022promptsource} and prompts for DeepStruct datasets are newly created.
Their composition is shown in Table~\ref{tab:mip}, which makes up natural language understanding and generation datasets from T0~\citep{sanh2022multitask} and promptsource~\citep{bach2022promptsource}, and information extraction datasets from DeepStruct~\citep{wang2022deepstruct}.
In \glm's training, we calculate that approximately 36\% of the samples in each dataset has been seen.

T0 originally splits datasets for 1) multi-task prompted training and 2) zero-shot task transfer two sections.
We initially planed to only include training sets of T0's multi-task prompted training section and DeepStruct~\citep{wang2022deepstruct}, but by a mistake we included both multi-task prompted training and zero-shot task transfer sections' datasets in MIP and excluded DeepStruct datasets. 
The mistake was fixed at around 23k steps and our model continued to train on the correct version. 

\vvpara{Natural Language Understanding and Generation.}
We adopt datasets and corresponding prompts from promptsource~\citep{bach2022promptsource}.
For all prompted samples in each dataset, we set a truncation of maximal 10,0000 samples per dataset and combine them together as the MIP dataset.
Details of the prompted samples and datasets are provided in promptsource's GitHub repository\footnote{\url{https://github.com/bigscience-workshop/promptsource}}.

\vvpara{Information Extraction.}
Based on the datasets from DeepStruct~\citep{wang2022deepstruct}, a multi-task language model pre-training approach for information extraction tasks, we create instructions and prompts for part of its datasets (as is shown in Table~\ref{tab:mip}).
We reformulate information extraction tasks into instruction tuning formats to allow zero-shot generalization to new extraction schema.
For all prompted samples in each dataset, we set a truncation of maximal 20,0000 samples per dataset as there are fewer information extraction datasets than common language understanding and generation ones.
For KELM~\citep{agarwal2021knowledge} and PropBank~\citep{kingsbury2002treebank} datasets, since their original size is gigantic, we sample 50,0000 samples for each of them from their prompted samples.

\subsection{Data and prompts in MIP for DeepStruct} \label{app:mip_dataset}
Prompts and instructions for all datasets in DeepStruct~\citep{wang2022deepstruct} are newly created by authors manually.
The introduction, task description, and full prompts for each dataset are attached in the following sections.
To allow template infilling, all prompts are written into Jinja\footnote{\url{https://github.com/pallets/jinja}} templates.
When a dataset sample is provided in our format, Joinja engine will render it into a prompted sample with instruction.

A more systematic evaluation on \glm's information extraction ability is left for a future work, as the concentration in this work is on the training and designing details of an LLM.

\subsubsection{Dialogue State Tracking}
We adopt Multiwoz 2.1~\citep{eric2020multiwoz} dialogue state tracking dataset.
The dataset is reformulated into two tasks, each with one prompt correspondingly:
\begin{itemize}[leftmargin=*,itemsep=0pt,parsep=0.2em,topsep=0.0em,partopsep=0.0em]
    \item \textbf{Dialogue state tracking}: which asks the model to extract information from dialogues given a list of certain slots, e.g., \texttt{taxi\_arrival\_time} and \texttt{destination}.
    \item \textbf{Slot filling}: which model should fill in one provided slot and identify situations without answer.
\end{itemize}

\begin{tcolorbox}[left=0mm,right=0mm,top=0mm,bottom=0mm,boxsep=1mm,arc=0mm,boxrule=0pt, frame empty]
\textbf{(Dialogue State Tracking, Prompt 0)}
\small
\begin{lstlisting}
Read the dialogues between "[User]" and "[Agent]",

{{text}}

identify and extract the information related to the following categories (from top to down):

- {{allowed_relations | join("\n- ")}}

in the form of "( [User] ; Y ; Z )": ||| {{format_triple(relations, allowed_relations) | join(" ")}}
\end{lstlisting}
\end{tcolorbox}

\begin{tcolorbox}[left=0mm,right=0mm,top=0mm,bottom=0mm,boxsep=1mm,arc=0mm,boxrule=0pt, frame empty]
\textbf{(Slot Filling, Prompt 0)}
\small
\begin{lstlisting}
Given the following dialogue:

{{text}}

please answer the question: has "[User]" mentioned "{{allowed_relations[relation_idx].split(': ') | join("'s ")}}" ? If yes, please write down the answer from the dialogue; if not, please answer "not given".

Answer: ||| {% if filter_relation(relations, allowed_relations[relation_idx]).__len__() > 0 %}{{filter_relation(relations, allowed_relations[relation_idx])[0]['tail']}}{% else %}not given{% endif %}
\end{lstlisting}
\end{tcolorbox}

\subsubsection{Event Extraction}
We adopt ACE05~\citep{walker2005ace} event extraction datasets following the setting in~\citep{wadden2019entity}.
The dataset is reformulated into two tasks with three prompts as follows:

\begin{itemize}[leftmargin=*,itemsep=0pt,parsep=0.2em,topsep=0.0em,partopsep=0.0em]
    \item \textbf{Event Argument Extraction}: given a trigger in text and a list of its argument roles, the model is asked to extract the arguments from the provided text.
    \item \textbf{Argument Identification}: given a trigger and a certain argument role, the model is asked to extract the argument if it exists in the provided text; otherwise, the model should generate nothing.
\end{itemize}

\begin{tcolorbox}[left=0mm,right=0mm,top=0mm,bottom=0mm,boxsep=1mm,arc=0mm,boxrule=0pt, frame empty]
\textbf{(Event Argument Extraction, Prompt 0)}
\small
\begin{lstlisting}
For the task of "Event Extraction", given a trigger one should extract its related arguments conditioned on a list of potential roles.

Given the following list of roles:

- {{shuffle(allowed_arguments[trigger['event_type']].values()) | join("\n- ")}}

extract related arguments of the trigger "{{trigger['text']}} ({{allowed_triggers[trigger['event_type']]}})" in the following sentence:

{{text}}

Extractions: ||| {{format_triple(relations, "") | join(" ")}}
\end{lstlisting}
\end{tcolorbox}

\begin{tcolorbox}[left=0mm,right=0mm,top=0mm,bottom=0mm,boxsep=1mm,arc=0mm,boxrule=0pt, frame empty]
\textbf{(Event Argument Extraction, Prompt 1)}
\small
\begin{lstlisting}
TEST

1. (Event Extraction) {{text}}

Please write down ALL event arguments related to the trigger "{{trigger['text']}} ({{allowed_triggers[trigger['event_type']]}})" marked with "[ ]", given the following categories:

- {{shuffle(allowed_arguments[trigger['event_type']].values()) | join("\n- ")}}

Answer: ||| {{format_triple(relations, "") | join(" ")}}
\end{lstlisting}
\end{tcolorbox}

\begin{tcolorbox}[left=0mm,right=0mm,top=0mm,bottom=0mm,boxsep=1mm,arc=0mm,boxrule=0pt, frame empty]
\textbf{(Argument Identification, Prompt 0)}
\small
\begin{lstlisting}
Let extract event related arguments!

In the following passage, an argument with the type "{{query_arg}}" is related to the event trigger "{{trigger['text']}} ({{allowed_triggers[trigger['event_type']]}})":

{{text}}

The argument should be (copy from the context if you find it; if not, do not generate): ||| {{filter_type(relations, query_arg) | join(" ")}}
\end{lstlisting}
\end{tcolorbox}

\subsubsection{Joint Entity and Relation Extraction}
Joint entity and relation extraction aims to recognize named entities in a piece of text and judge the relationships between them.
It is closely related to knowledge acquisition, where the ultimate target is to structuring the unstructured web contents into knowledge triples (e.g., \texttt{(London, capital\_of, Britain)}).
The task can be formulated into either a pipeline framework (a combination of named entity recognition and relation extraction), or end-to-end training.

In this work, we adopt three classical joint entity and relation extraction datasets: CoNLL04~\citep{roth-yih-2004-linear}, NYT~\citep{10.1007/978-3-642-15939-8_10}, and ACE2005~\citep{walker2005ace}.
In \glm, we follow~\citep{wang2022deepstruct} to formulate such challenges into sequence-to-sequence generation, where our inputs are raw texts and outputs are triples.
We only conduct relation-related tasks for these datasets here, and leave the entity-related ones to the named entity recognition section.
\begin{itemize}[leftmargin=*,itemsep=0pt,parsep=0.2em,topsep=0.0em,partopsep=0.0em]
    \item \textbf{Relation Extraction}: here we extract knowledge triples consisting of ``head entity'', ``relation'', and ``tail entity'', given a list of relation candidates. For example, given the input ``\textit{In Kunming the 800-some faculty and student established the National Southwestern Associated University.}'', the model output could be \texttt{(National Southwestern Associated University, location of formation, Kunming)}.
    \item \textbf{Conditional Relation Extraction}: given a single relation candidate, judge if the input text contains the relation. If so, extraction all related triples; if not, do not generate.
    \item \textbf{Knowledge Slot Filling}: assign a certain entity from text, and ask the model to extract all triples that takes the entity as the head.
    \item \textbf{Relation Classification}: given two entities from texts, ask the model to judge the relation between them based on a list of candidate relations.
\end{itemize}

\begin{tcolorbox}[left=0mm,right=0mm,top=0mm,bottom=0mm,boxsep=1mm,arc=0mm,boxrule=0pt, frame empty]
\textbf{(Relation Extraction, Prompt 0)}
\small
\begin{lstlisting}
Can you figure out all triples regarding the relations of "{{shuffle(allowed_relations) | join('", "')}}" from the sentence? List them in the shape of "( X ; Y ; Z )":

{{text}} => ||| {{format_triple(relations, allowed_relations) | join(" ")}}
\end{lstlisting}
\end{tcolorbox}

\begin{tcolorbox}[left=0mm,right=0mm,top=0mm,bottom=0mm,boxsep=1mm,arc=0mm,boxrule=0pt, frame empty]
\textbf{(Conditional Relation Extraction, Prompt 0)}
\small
\begin{lstlisting}
Conditioned on the relation "{{allowed_relations[relation_idx]}}", what knowledge triples can be extracted from:

{{text}}

Please write them down here: ||| {{format_triple(relations, [allowed_relations[relation_idx]]) | join(" ")}}
\end{lstlisting}
\end{tcolorbox}

\begin{tcolorbox}[left=0mm,right=0mm,top=0mm,bottom=0mm,boxsep=1mm,arc=0mm,boxrule=0pt, frame empty]
\textbf{(Knowledge Slot Filling, Prompt 0)}
\small
\begin{lstlisting}
{% if entity_types.__len__() > 0 %}
In the sentence

{{text}}

the X = "{{entities[entity_idx]}}" is an entity of the type "{{entity_types[entity_idx]}}". Extract all possible triples contains "{{entities[entity_idx]}}" in the form of ( X ; Y ; Z ), given the following candidate properties Y:

{% for r in allowed_relations %}- {{r}}
{% endfor %}
Answer: ||| {% for r in relations %}{% if r['head'][0] == entities[entity_idx] %}{{format_triple([r], allowed_relations) | join(" ")}}{% endif %}{% endfor %}
{% endif %}
\end{lstlisting}
\end{tcolorbox}

\begin{tcolorbox}[left=0mm,right=0mm,top=0mm,bottom=0mm,boxsep=1mm,arc=0mm,boxrule=0pt, frame empty]
\textbf{(Relation Classification, Prompt 0)}
\small
\begin{lstlisting}
QUIZ

1. Given the candidate relations:

- {{shuffle(allowed_relations) | join("\n- ")}}

what is the relation between "{{relations[triple_idx]['head'][0]}}" and "{{relations[triple_idx]['tail'][0]}}" in the following sentence?

{{text}}

Answer: ||| {{relations[triple_idx]['relation']}}
\end{lstlisting}
\end{tcolorbox}

Nevertheless, existing joint entity and relation extraction datasets have very limited relation schema.
For example, CoNLL04 only contains five different relations; the most diverse NYT dataset contains 24 Freebase predicates.
To allow the model to capture a diverse range of potential verbalized predicates, we extend the task with automatically generated knowledge-text aligned data from KELM~\citep{agarwal2021knowledge}.
We do not include other distantly supervised dataset (e.g., T-Rex~\citep{elsahar2018t}) since they can be extremely noisy.

For KELM data, since it is based on the full Wikidata schema (which contains too many relations to be enumerated), we create two KELM-specific prompts for the task of \textbf{Relation Extraction} and \textbf{Knowledge Slot Filling}:

\begin{tcolorbox}[left=0mm,right=0mm,top=0mm,bottom=0mm,boxsep=1mm,arc=0mm,boxrule=0pt, frame empty]
\textbf{(Relation Extraction, Prompt 1, KELM ONLY)}
\small
\begin{lstlisting}
{# kelm #}
Can you figure out all knowledge triples regarding whole Wikidata properties from the sentence? List them in the shape of "( X ; Y ; Z )":

{{text}} => ||| {{format_triple(relations, "") | join(" ")}}
\end{lstlisting}
\end{tcolorbox}

\begin{tcolorbox}[left=0mm,right=0mm,top=0mm,bottom=0mm,boxsep=1mm,arc=0mm,boxrule=0pt, frame empty]
\textbf{(Knowledge Slot Filling, Prompt 1, KELM ONLY)}
\small
\begin{lstlisting}
{# kelm #}
Given the entity "{{entities[entity_idx]}}" marked with "[" and "]" in the context:

{{text}}

please list all triples related to it (do not generate if there is no answer): ||| {% for r in relations %}{% if r['head'][0] == entities[entity_idx] %}{{format_triple([r], "") | join(" ")}}{% endif %}{% endfor %}
\end{lstlisting}
\end{tcolorbox}

\subsubsection{Named Entity Recognition}
Named entity recognition is a task which targets identifying named entities from raw text corpus and assign them with proper entity types.
For example, in the sentence ``\textit{In 1916 GM was reincorporated in Detroit as "General Motors Corporation".}'', \texttt{General Motors Corporation} could be of entity type \texttt{organization}.
We design two different types of tasks based on named entity recognition datasets CoNLL03~\citep{tjong-kim-sang-de-meulder-2003-introduction}, OntoNotes 5.0~\citep{pradhan-etal-2013-towards}, and GENIA~\citep{10.5555/1289189.1289260}. We also include named entity recognition sub-tasks from joint entity and relation datasets.
\begin{itemize}[leftmargin=*,itemsep=0pt,parsep=0.2em,topsep=0.0em,partopsep=0.0em]
    \item \textbf{Named Entity Recognition}: given a certain list of possible entity types (e.g., \texttt{location}, \texttt{person}, \texttt{organization}), extract all related entities from the provided text content.
    \item \textbf{Entity Typing}: entity typing is one of the important derivative tasks from named entity recognition. It aims to classify the correct type of an entity mention (without entity types), and is often appended to the entity mention extraction as post-processing.
\end{itemize}

\begin{tcolorbox}[left=0mm,right=0mm,top=0mm,bottom=0mm,boxsep=1mm,arc=0mm,boxrule=0pt, frame empty]
\textbf{(Named Entity Recognition, Prompt 0)}
\small
\begin{lstlisting}
Given the following list of entity types:

Z = {{shuffle(allowed_types) | join(", ")}}

please extract all mentioned entities from left to right in the sentence, in the form of "( X ; instance of ; Z )".

{{text}} => ||| {% for entity, type in zip(entities, entity_types) %}( {{entity}} ; instance of ; {{type}} ) {% endfor %}
\end{lstlisting}
\end{tcolorbox}

\begin{tcolorbox}[left=0mm,right=0mm,top=0mm,bottom=0mm,boxsep=1mm,arc=0mm,boxrule=0pt, frame empty]
\textbf{(Entity Typing, Prompt 0)}
\small
\begin{lstlisting}
Extract all entity mentioned in the sentence with entity type "{{allowed_types[type_idx]}}" in the form of "( X ; instance of ; {{allowed_types[type_idx]}} )"

{{text}} => ||| {% for entity, type in zip(entities, entity_types) %}{% if type == allowed_types[type_idx] %}( {{entity}} ; instance of ; {{type}} ) {% endif %}{% endfor %}
\end{lstlisting}
\end{tcolorbox}

\begin{tcolorbox}[left=0mm,right=0mm,top=0mm,bottom=0mm,boxsep=1mm,arc=0mm,boxrule=0pt, frame empty]
\textbf{(Entity Typing, Prompt 1)}
\small
\begin{lstlisting}
List all "{{allowed_types[type_idx]}}" entities appeared in the following passage, joined by " | ":

{{text}} => ||| {{filter_type(zip(entities, entity_types), allowed_types[type_idx]) | join(" | ")}}
\end{lstlisting}
\end{tcolorbox}

\begin{tcolorbox}[left=0mm,right=0mm,top=0mm,bottom=0mm,boxsep=1mm,arc=0mm,boxrule=0pt, frame empty]
\textbf{(Entity Typing, Prompt 2)}
\small
\begin{lstlisting}
{% if entity_types.__len__() > 0 %}
Based on the list of potential entity types and ignore their order:

- {{shuffle(allowed_types) | join("\n- ")}}

the entity "{{entities[entity_idx]}}" marked with "[" and "]" in the following sentence:

{{text}}

belongs to ||| {{entity_types[entity_idx]}}
{% endif %}
\end{lstlisting}
\end{tcolorbox}

\subsubsection{Relation Classification}
Relation classification is a fundamental task in information extraction, which identifies the relationships from a list of candidates between two given entities.
The problem is a long standing one as it suffers from outrageous cost of data labeling, since manual labeling on knowledge-intensive tasks requires educated annotators that charges high.
A \textit{de facto} data creation method in relation extraction relies on distant supervision, which aligns existing knowledge triples in knowledge bases to text contents automatically, and assume that such alignments are correct in certain conditions.
Here we only include TacRED~\citep{zhang2017tacred} dataset and create several different tasks based on it.
\begin{itemize}[leftmargin=*,itemsep=0pt,parsep=0.2em,topsep=0.0em,partopsep=0.0em]
    \item \textbf{Relation Classification}: the most traditional task formulation. Given two entities from text and classify their relation from a list of candidates. The form can be either answering the relation directly or in the form of a triple (similar to relation extraction).
    \item \textbf{Knowledge Slot Filling}: change the task into given head entity and relation, to identify whether the tail entity exists in the input text. If not, generate nothing.
    \item \textbf{Yes or No Question}: turn the problem into a task similar to natural language inference. For example, given the sentence ``\textit{The series focuses on the life of Carnie Wilson, daughter of Brian Wilson, founder of the Beach Boys.}'', the model will be asked to judge the correctness of a triple such as \texttt{Carnie Wilson, father, Brian Wilson} by answering ``yes'' or ``no''.
\end{itemize}

\begin{tcolorbox}[left=0mm,right=0mm,top=0mm,bottom=0mm,boxsep=1mm,arc=0mm,boxrule=0pt, frame empty]
\textbf{(Relation Classification, Prompt 0)}
\small
\begin{lstlisting}
{% if entity_types.__len__() > 0 %}
Given the following categories of relations:

- {{shuffle(allowed_relations.values()) | join("\n- ")}}

predict the relation between "{{relations[0]['head']}}" and "{{relations[0]['tail']}}" in the following sentence:

{{text}}

The relation should be : ||| {{allowed_relations[relations[0]['relation']]}}
{% endif %}
\end{lstlisting}
\end{tcolorbox}

\begin{tcolorbox}[left=0mm,right=0mm,top=0mm,bottom=0mm,boxsep=1mm,arc=0mm,boxrule=0pt, frame empty]
\textbf{(Relation Classification, Prompt 1)}
\small
\begin{lstlisting}
1. (Relation Extraction) Answer the relation between entities in the form of "( X ; Y ; Z )":

{{text}}

The relation between "{{relations[0]['head']}}" and "{{relations[0]['tail']}}" is: ||| ( {{relations[0]['head']}} ; {{allowed_relations[relations[0]['relation']]}} ; {{relations[0]['tail']}} )
\end{lstlisting}
\end{tcolorbox}

\begin{tcolorbox}[left=0mm,right=0mm,top=0mm,bottom=0mm,boxsep=1mm,arc=0mm,boxrule=0pt, frame empty]
\textbf{(Knowledge Slot Filling, Prompt 0)}
\small
\begin{lstlisting}
Based on the sentence provided below, infer the missing argument asked by the question:

{{text}}

Question: What/Who/Where is "{{relations[0]['head']}}" {{allowed_relations[relations[0]['relation']]}} ?

Answer: ||| {{relations[0]['tail']}}
\end{lstlisting}
\end{tcolorbox}

\subsubsection{Semantic Role Labeling}
Semantic role labeling is a long-standing information task that wants to identify the semantic arguments related to a given predicate in a sentence.
For example, in the sentence ``\textit{Grant was employed at IBM for 21 years where she held several executive positions.}'' and the predicate ``\texttt{employed}'' in it, semantic role labeling identifies the \texttt{Grant} as the subject and \texttt{IBM} as the second object.

We create two different tasks based on semantic role labelling datasets CoNLL05~\citep{carreras-marquez-2005-introduction}, CoNLL12~\citep{pradhan-etal-2013-towards}, and PropBank~\citep{kingsbury2002treebank}.
\begin{itemize}[leftmargin=*,itemsep=0pt,parsep=0.2em,topsep=0.0em,partopsep=0.0em]
    \item \textbf{Semantic Role Labeling}: the traditional task form, where a verb (i.e., predicate) is annotated in text and the model is asked to generate related semantic roles.
    \item \textbf{Semantic Role Filling}: given a verb and and a potential semantic role, the model is asked to judge whether the role exists in the sentence and generate it.
    \item \textbf{Predicate Recognition}: given a segment of a sentence and its corresponding semantic role, identify which verb it is related to.
\end{itemize}

\begin{tcolorbox}[left=0mm,right=0mm,top=0mm,bottom=0mm,boxsep=1mm,arc=0mm,boxrule=0pt, frame empty]
\textbf{(Semantic Role Labeling, Prompt 0)}
\small
\begin{lstlisting}
Provided with the target verb "{{verb}}" marked with "[" and "]" in the following sentence, find out its "{{allowed_types[type_idx]}}":

{{text}} => ||| {% for entity, type in zip(entities, entity_types) %}{% if type == allowed_types[type_idx] %}{{entity}}{% endif %}{% endfor %}
\end{lstlisting}
\end{tcolorbox}

\begin{tcolorbox}[left=0mm,right=0mm,top=0mm,bottom=0mm,boxsep=1mm,arc=0mm,boxrule=0pt, frame empty]
\textbf{(Semantic Role Filling, Prompt 0)}
\small
\begin{lstlisting}
Given the following list of argument types:

Z = {{allowed_types | join(", ")}}

find out all arguments related to verb "{{verb}}" mentioned in the following sentence from left to right, in the form of "( X ; instance of ; Z )".

{{text}} => ||| {% for entity, type in zip(entities, entity_types) %}( {{entity}} ; argument type ; {{type}} ) {% endfor %}
\end{lstlisting}
\end{tcolorbox}

\begin{tcolorbox}[left=0mm,right=0mm,top=0mm,bottom=0mm,boxsep=1mm,arc=0mm,boxrule=0pt, frame empty]
\textbf{(Predicate Recognition, Prompt 0)}
\small
\begin{lstlisting}
FINAL EXAM

1. Based on the fact that "{{entities[entity_idx]}}" is a "{{entity_types[entity_idx]}}", which verb in the following sentence should it related to?

{{text}}

Answer: ||| {{verb}}
\end{lstlisting}
\end{tcolorbox}

\subsection{Result Sources for GPT-3, BLOOM-176B, and OPT-175B} \label{app:opt_sources}
Here we describe the result sources for GPT-3, BLOOM-176B, and OPT-175B.
Other LLMs we may compare are mostly completely closed-sourced; thus, their results are all taken from existing preprints, publications, or the results stored in BIG-bench repository\footnote{\url{https://github.com/google/BIG-bench}}.

For GPT-3, while most of its results in this paper are taken from existing literature if not specified, the rest were acquired via our own requesting OpenAI Danvici API are explicitly mentioned. 
For BLOOM-176B and OPT-175B, if without specific annotation, their results are:
\begin{itemize}[leftmargin=*,itemsep=0pt,parsep=0.2em,topsep=0.0em,partopsep=0.0em]
    \item Taken from the OPT paper~\citep{zhang2022opt}.
    \item Taken from the EAI-Eval BigScience Arch\&Scale - Google Sheet\footnote{\url{https://docs.google.com/spreadsheets/d/1CI8Q9RCblLRzUOPJ6ViqBmo284-8ojluQ-CmaEuhuv0}}.
    \item Taken from BigScience evaluation results repository in Huggingface Datasets\footnote{\url{https://huggingface.co/datasets/bigscience/evaluation-results/tree/main/bloom/bloomzeval/transformers/evaluation_val}}.
\end{itemize}

Specifically, we cannot evaluate OPT-175B by ourselves as we are still not officially granted the checkpoint, though we have sent several applications in the past few months.

\subsection{Pile Test-set Evaluation} \label{app:pile}
\begin{wraptable}{r}{6.5cm}
	\centering
	\footnotesize
	\vspace{-15mm}
    \renewcommand\tabcolsep{3pt}
	\caption{\glm and its similar-sized LLMs' BPB results on Pile test-set.}
	\vspace{-1mm}
	\scalebox{0.97}{
	\begin{tabular}{@{}lccc@{}}
    \toprule[1.2pt]
                      & Jurassic-1     & GPT-3          & \glm           \\ \midrule
    dm\_mathematics    & 1.040          & 1.370          & \textbf{0.786} \\
    ubuntu\_irc        & \textbf{0.857} & 0.946          & 0.977          \\
    opensubtitles      & \textbf{0.879} & 0.932          & 0.889          \\
    hackernews         & \textbf{0.869} & 0.975          & 0.873          \\
    books33            & 0.835          & \textbf{0.802} & 0.803          \\
    pile\_cc           & \textbf{0.669} & 0.698          & 0.771          \\
    philpapers         & 0.741          & \textbf{0.723} & 0.766          \\
    gutenberg\_pg\_19  & 0.890          & 1.160          & \textbf{0.821} \\
    arxiv              & 0.680          & 0.838          & \textbf{0.570} \\
    stackexchange      & 0.655          & 0.773          & \textbf{0.611} \\
    nih\_exporter      & \textbf{0.590} & 0.612          & 0.614          \\
    pubmed\_abstracts  & \textbf{0.587} & 0.625          & 0.610          \\
    uspto\_backgrounds & \textbf{0.537} & 0.566          & \textbf{0.537} \\
    pubmed\_central    & 0.579          & 0.690          & \textbf{0.510} \\
    freelaw            & 0.514          & 0.612          & \textbf{0.499} \\
    github             & 0.358          & 0.645          & \textbf{0.329} \\
    enron\_emails      & 0.621          & 0.958          & \textbf{0.604} \\
    youtube\_subtitles & 0.825          & 0.815          & \textbf{0.746} \\ \midrule
    Weighted Avg.      & 0.650          & 0.742          & \textbf{0.634} \\ \bottomrule[1.2pt]
    \end{tabular}
    }
    \vspace{-6mm}
    \label{tab:pile_full}
\end{wraptable}

Pile evalution~\citep{gao2020pile} is a comprehensive language modeling benchmark which originally includes 22 different text datasets from diverse domains.
We report our results over a part of 18 datasets with previously reported baseline results~\citep{lieber2021jurassic}.
Different from traditional language modeling benchmarks, Pile evaluation report the BPB (bits-per-byte) perplexity to avoid the mismatch comparison between models with different vocabularies.
Because in general, language models with a larger vocabulary will be favored in perplexity comparison if not restricted.
In the evaluation, we strictly follow the setting in~\citep{gao2020pile}, leveraging [gMASK] and a context-length of 1,024 with bidirectional attention, and the rest 1024 tokens to calculate BPB in an autoregressive manner.
The weighted average BPB are calculated based on each shared dataset's ratio in Pile training-set~\citep{gao2020pile}.

The detailed metrics on Pile test-set are reported in Table~\ref{tab:pile_full}. We observe that compared to GPT-3, \glm has a noticeable weaker performance on phil\_papers and pile\_cc, which is likely because of \glm's bilingual natural and lack of more diverse and high-quality private collected corpora.


\subsection{BIG-bench-lite Evaluation} \label{app:big-bench}
\begin{wrapfigure}{r}{5.5cm}
    \centering
    \vspace{-5mm}
    \includegraphics[width=\linewidth]{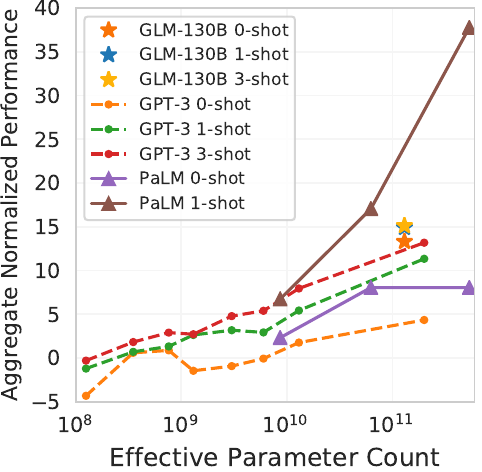}
    \caption{A full scope of BIG-bench-lite (24 tasks) evaluation.}
    \label{fig:big-bench-lite-full}
\end{wrapfigure}
Recent works~\citep{wei2022chain,wang2022rationale} reveal that LLMs are capable to do reasoning beyond conventional language tasks.
As a response, BIG-bench~\citep{srivastava2022beyond} is recently set up by crowdsourcing new types of tasks from global researchers to test LLMs unexplored abilities.
For economical consideration, we evaluate \glm on an official subset of original 150-task BIG-bench, the BIG-bench-lite with 24 tasks.
These tasks can be categorized into two types: one is based on multiple-choice question answering with answer options, and another is direct generation without options.
For the first category, we assess the probability of each option's full content and pick the largest one as the answer; for the second one, we generate the answer using greedy decoding.
All evaluations done in BIG-bench are based on [MASK], since answers here are usually short pieces of texts.
All results on 24 BIG-bench-lite~\citep{srivastava2022beyond} datasets of three LLMs are shown in Table~\ref{tab:big-bench-details} and Figure~\ref{fig:big-bench-lite-full}.
We just adopt the original prompts from BIG-bench and use the official implementation to generate priming examples for few-shot evaluation and to calculate the final scores.
\begin{table}[t]
\centering
\footnotesize
\vspace{2mm}
\renewcommand\tabcolsep{4.5pt}
\renewcommand\arraystretch{1.1}
\caption{Details results of \glm, GPT-3 175B~\citep{brown2020language}, and PaLM 540B~\citep{chowdhery2022palm} on BIG-bench-lite in 0, 1, and 3-shots. ``Normalized preferred metric'' is reported for each task. GPT-3 and PaLM's results are reported in BIG-bench's GitHub repository, and PaLM 540B's 3-shot results are not found.}
\begin{tabular}{@{}lcccccccc@{}}
\toprule[1.2pt]
                                        & \multicolumn{3}{c}{\glm} & \multicolumn{3}{c}{GPT-3 175B} & \multicolumn{2}{c}{PaLM 540B} \\ \cmidrule(l){2-4} \cmidrule(l){5-7} \cmidrule(l){8-9} 
                                        & 0      & 1      & 3      & 0        & 1        & 3        & 0             & 1             \\ \midrule
auto\_debugging                         & 11.76  & 20.59  & 23.53  & 0.00     & 0.00     & 0.00     & 0.00          & 38.23         \\
bbq\_lite\_json                         & 22.26  & 37.50  & 59.73  & -8.33    & 40.75    & 61.21    & -4.39         & 77.73         \\
code\_line\_description                 & 0.22   & 9.09   & -8.64  & 9.09     & 9.09     & 9.09     & 0.22          & 49.00         \\
conceptual\_combinations                & 37.51  & 31.33  & 27.86  & 2.37     & 3.70     & 14.33    & 45.68         & 73.36         \\
conlang\_translation                    & 34.72  & 38.01  & 33.88  & 46.82    & 47.07    & 51.60    & 36.88         & 61.92         \\
emoji\_movie                            & 1.25   & 4.88   & 3.75   & -10.00   & -2.49    & -1.24    & 17.50         & 88.75         \\
formal\_fallacies\_syllogisms\_negation & 0.83   & 1.46   & 0.35   & 1.00     & 6.80     & 5.60     & -0.20         & 4.40          \\
hindu\_knowledge                        & 32.23  & 37.56  & 34.52  & 10.15    & 40.61    & 44.42    & 41.37         & 93.15         \\
known\_unknowns                         & -4.35  & 0.00   & 4.35   & 21.74    & 4.35     & 0.00     & 13.04         & 34.78         \\
language\_identification                & 9.62   & 1.97   & 1.90   & 7.49     & 3.20     & 1.98     & 12.11         & 31.03         \\
linguistics\_puzzles                    & 0.00   & 0.00   & 0.00   & 0.00     & 0.00     & 0.00     & 0.00          & 0.10          \\
logic\_grid\_puzzle                     & 9.88   & 13.66  & 5.24   & 0.16     & 3.35     & 0.01     & 1.47          & 16.12         \\
logical\_deduction                      & 24.18  & 22.20  & 20.35  & 2.22     & 10.80    & 14.71    & 2.17          & 15.34         \\
misconceptions\_russian                 & -26.53 & -46.94 & -26.53 & -34.70   & -34.70   & -30.61   & -42.86        & -30.61        \\
novel\_concepts                         & 6.25   & 21.87  & 25.78  & 33.59    & 33.59    & 45.31    & 33.59         & 49.22         \\
operators                               & 14.76  & 18.10  & 18.10  & 30.0     & 34.29    & 33.33    & 30.48         & 56.19         \\
parsinlu\_reading\_comprehension        & 7.14   & 7.72   & 11.58  & 0.00     & 0.00     & 0.00     & 9.46          & 44.40         \\
play\_dialog\_same\_or\_different       & 2.88   & 5.33   & 3.80   & 8.00     & 0.80     & -5.40    & -33.0         & 0.10          \\
repeat\_copy\_logic                     & 0.00   & 0.00   & 0.00   & 0.00     & 0.00     & 0.00     & 0.00          & 37.5          \\
strange\_stories                        & 43.86  & 51.76  & 42.31  & 8.27     & 25.68    & 12.93    & 39.25         & 74.46         \\
strategyqa                              & 21.10  & 18.74  & 16.82  & 4.60     & 13.20    & 14.20    & 28.00         & 38.00         \\
symbol\_interpretation                  & 1.39   & 1.89   & 1.77   & 0.51     & -0.63    & 2.77     & 0.76          & 2.40          \\
vitaminc\_fact\_verification            & 71.87  & 60.72  & 56.55  & -31.55   & 22.15    & 29.05    & -28.85        & 55.60         \\
winowhy                                 & -3.49  & 5.38   & 3.0    & 3.0      & 10.60    & 13.00    & -5.0          & 31.80         \\ \bottomrule[1.2pt]
\end{tabular}
\label{tab:big-bench-details}
\end{table}

\subsection{MMLU Evaluation} \label{app:mmlu}
All results on 57 MMLU~\citep{hendrycks2021measuring} datasets of \glm and BLOOM 176B are shown in Table~\ref{tab:mmlu}.
In Section~\ref{sec:mmlu}, we report weighted average accuracy (i.e., accuracy average per sample, rather than by discipline) of \glm, GPT-3 175B, and BLOOM 176B.

\begin{table}[t]
\footnotesize
\centering
\caption{Detailed results of \glm and BLOOM 176B~\citep{scao2022what} on MMLU~\citep{hendrycks2021measuring}. We find that no existing literature has reported GPT-3 175B's numerical accuracy. BLOOM is evaluated using Huggingface Transformer implementation.}
\label{tab:mmlu}
\begin{tabular}{@{}clcc@{}}
\toprule[1.2pt]
                                 & Discipline                              & \glm  & BLOOM 176B \\ \midrule
\multirow{19}{*}{STEM}           & abstract\_algebra                       & 24.00 & 24.00      \\
                                 & anatomy                                 & 48.90 & 38.52      \\
                                 & astronomy                               & 48.03 & 34.87      \\
                                 & colledge\_biology                       & 47.22 & 37.50      \\
                                 & college\_chemistry                      & 34.00 & 19.00      \\
                                 & colledge\_computer\_science             & 44.00 & 1.00       \\
                                 & colledge\_mathematcis                   & 27.00 & 31.00      \\
                                 & colledge\_physics                       & 30.39 & 24.50      \\
                                 & computer\_security                      & 61.00 & 40.00      \\
                                 & conceptual\_physics                     & 38.72 & 31.49      \\
                                 & electrical\_engineering                 & 45.52 & 32.41      \\
                                 & elementary\_mathematics                 & 31.75 & 29.63      \\
                                 & high\_school\_biology                   & 51.29 & 27.42      \\
                                 & high\_school\_chemistry                 & 34.98 & 27.09      \\
                                 & high\_school\_computer\_science         & 53.00 & 30.00      \\
                                 & high\_school\_mathematics               & 28.15 & 25.93      \\
                                 & high\_school\_physics                   & 29.80 & 30.46      \\
                                 & high\_school\_statistics                & 38.43 & 26.39      \\
                                 & machine\_learning                       & 40.18 & 29.46      \\ \midrule
\multirow{12}{*}{Social Science} & econometrics                            & 26.32 & 26.32      \\
                                 & high\_school\_geography                 & 53.54 & 36.36      \\
                                 & high\_school\_government\_and\_politics & 62.18 & 40.41      \\
                                 & high\_school\_macroeconomics            & 42.56 & 30.77      \\
                                 & high\_school\_microeconomics            & 45.80 & 26.89      \\
                                 & high\_school\_psychology                & 54.13 & 39.27      \\
                                 & human\_sexuality                        & 51.15 & 35.11      \\
                                 & professional\_psychology                & 42.48 & 31.54      \\
                                 & public\_relations                       & 55.46 & 33.64      \\
                                 & security\_studies                       & 44.90 & 34.29      \\
                                 & sociology                               & 51.74 & 31.84      \\
                                 & us\_foreign\_policy                     & 61.00 & 46.00      \\ \midrule
\multirow{13}{*}{Humanities}     & formal\_logic                           & 27.78 & 23.02      \\
                                 & high\_school\_european\_history         & 58.18 & 35.76      \\
                                 & high\_school\_us\_history               & 58.33 & 40.69      \\
                                 & high\_school\_world\_history            & 67.09 & 32.07      \\
                                 & international\_law                      & 56.20 & 42.15      \\
                                 & jurisprudence                           & 43.52 & 35.19      \\
                                 & logical\_fallacies                      & 57.06 & 31.29      \\
                                 & moral\_disputes                         & 47.11 & 36.71      \\
                                 & moral\_scenarios                        & 24.25 & 24.36      \\
                                 & philosophy                              & 45.34 & 35.37      \\
                                 & prehistory                              & 50.93 & 40.43      \\
                                 & professional\_law                       & 37.94 & 29.53      \\
                                 & world\_religions                        & 55.56 & 42.11      \\ \midrule
\multirow{13}{*}{Other}          & business\_ethics                        & 51.00 & 34.00      \\
                                 & clinical\_knowledge                     & 48.68 & 35.85      \\
                                 & colledge\_medicine                      & 43.35 & 28.90      \\
                                 & glocal\_facts                           & 35.00 & 23.00      \\
                                 & human\_aging                            & 45.29 & 32.29      \\
                                 & management                              & 56.31 & 27.18      \\
                                 & marketing                               & 67.52 & 39.74      \\
                                 & medical\_genetics                       & 48.00 & 45.00      \\
                                 & miscellaneous                           & 61.18 & 40.23      \\
                                 & nutrition                               & 50.65 & 32.35      \\
                                 & professional\_accounting                & 35.46 & 28.72      \\
                                 & professional\_medicine                  & 43.38 & 18.01      \\
                                 & virology                                & 39.16 & 28.31      \\ \bottomrule[1.2pt]
\end{tabular}
\end{table}

Below is a prompted example with 1-shot priming. We predict the probability on \texttt{['A', 'B', 'C', 'D']} at the next token, and take the one with the maximal probability as the answer.

\begin{tcolorbox}[left=0mm,right=0mm,top=0mm,bottom=0mm,boxsep=1mm,arc=0mm,boxrule=0pt, frame empty]
\textbf{(MMLU 1-shot Example)}
\small
\begin{lstlisting}
The following are multiple choice questions about philosophy.

According to d'Holbach, people always act according to _____.
(A) free choices (B) dictates of the soul (C) necessary natural laws (D) undetermined will
Answer: (C) necessary natural laws

Epicurus holds that philosophy is:
(A) not suitable for the young. (B) not suitable for the old. (C) important, but unpleasant. (D) none of the above.
Answer: (
\end{lstlisting}
\end{tcolorbox}

\subsection{Chinese Language Understanding Evaluation} \label{app:clue}
Here we elaborate the prompts we use for CLUE~\citep{xu2020clue} and FewCLUE~\citep{xu2021fewclue} evaluation.
On Chinese datasets, prompting meets some challenges as Chinese texts are organized by single characters rather than words, leading to unequal length of verbalizers in many cases.
Albeit dataset-specific calibration~\citep{wang2021ernie,wu2021yuan} can help to mitigate the issue, the too specified technique can be complicated in implementation. 
Our evaluation in this paper adopts a more easy to solve method leveraging \glm's unique features.
As \glm is a bilingual LLM with English MIP, we adopt English prompts and verbalizers from similar tasks in~\citep{bach2022promptsource} for Chinese dataset evaluation and find such strategies to be quite effective.
In terms of evaluation metrics, except for DRCD and CMRC2018 two question answering datasets which reports EM, other datasets report accuracy.

\subsection{Natural Language Generation}
Natural language generation, or conditional natural language generation here, refers to tasks that require generating text based on the given information, such as tables and documents. 
We evaluate \glm on data-to-text and summarization tasks. The datasets include WebNLG 2020~\citep{Ferreira2020WebNLG}, Clean E2E NLG~\citep{dusek2019semantic} and WikiLingua~\citep{scialom2020mlsum} from GEM generation benchmark~\citep{gehrmann2021gem}. 
We select full WebNLG 2020 and the Clean E2E NLG in the test set and randomly select 5000 test examples from WikiLingua following the practice in~\citep{chowdhery2022palm}. 
Following the settings in PaLM, the prompt used for the Summarization tasks is ``Summarize the following article:'' and the prompt used for the Data-to-Text tasks is ``Verbalize:''.
An exception is E2E, where we process the data using the prompt ``generate-gramatically-correct-text from'' provided in promptsource for GLM-130B and GPT-3 175B (Davinci).
All evaluations are one-shot, and the demonstration samples are randomly sampled from the training set. 
We report the F-measure of ROUGE-2, ROUGE-L~\citep{lin2004rouge} and BLEURT-20~\citep{pu2021learning}. We compare our model with LaMDA, GPT-3 175B (Davinci), and PaLM, where the results of LaMDA and PaLM are reported by ~\citep{chowdhery2022palm}, and we evaluate GPT-3 175B (Davinci) through OpenAI API.\footnote{We use ROUGE implementation at \href{https://github.com/google-research/google-research/tree/master/rouge}{https://github.com/google-research/google-research/tree/master/rouge} and BLEURT-20 implementation at \href{https://github.com/google-research/google-research/tree/master/rouge}{https://github.com/google-research/google-research/tree/master/rouge}, whose checkpoint is available at \href{https://storage.googleapis.com/bleurt-oss-21/BLEURT-20.zip}{https://storage.googleapis.com/bleurt-oss-21/BLEURT-20.zip}}

Our results are presented in Table~\ref{tab:nlg}. It shows that \glm has better performances than LaMDA and GPT-3 (Davinci) on all tasks. 
In the Data-to-text task, \glm performs slightly worse than PaLM-540B, while in the summary task, \glm has even higher ROUGE results. 
We also ablate GLM-130B to unidirectional to demonstrate the advantage of bidirectional attention.
Unidirectional GLM-130B underperforms GPT-3 175B in all three datasets, but when it shifts to bidirectional attention, there is an instant boost, making GLM-130B even comparable to PaLM-540B in a few cases.
It indicates that bidirectional attention over the provided context (i.e., prefix) can also be beneficial for text generation missions.

\vspace{-2mm}
\begin{table}[H]
\centering
\caption{1-shot GEM English natural language generation tasks (WebNLG, E2E, and WikiLingua). We compare two versions of GLM-130B (uni: unidirectional attention, bi: bidirectional attention), showing that bidirectional attention can also improve conditional generation's performance.}
\vspace{-2mm}
\footnotesize
\begin{tabular}{@{}cllccccc@{}}
\toprule[1.2pt]
\multirow{2}{*}{Task}                                                     & \multicolumn{1}{c}{\multirow{2}{*}{Dataset}}    & \multicolumn{1}{c}{\multirow{2}{*}{Metric}} & \multirow{2}{*}{\begin{tabular}[c]{@{}c@{}}LaMDA\\ 137B\end{tabular}} & \multirow{2}{*}{\begin{tabular}[c]{@{}c@{}}GPT-3 175B\\ (Davinci)\end{tabular}} & \multicolumn{2}{c}{GLM-130B} & \multirow{2}{*}{PaLM-540B} \\ \cmidrule(lr){6-7}
                                                                          & \multicolumn{1}{c}{}                            & \multicolumn{1}{c}{}                        &                                                                       &                                                                                 & uni      & bi                &                            \\ \midrule
\multirow{6}{*}{\begin{tabular}[c]{@{}c@{}}Data\\ to\\ Text\end{tabular}} & \multirow{3}{*}{WebNLG}                         & ROUGE-2                                     & 30.5                                                                  & 29.9                                                                            & 25.3     & {\ul 38.5}        & \textbf{44.4}              \\
                                                                          &                                                 & ROUGE-L                                     & -                                                                     & 41.2                                                                            & 36.7     & {\ul 49.3}        & \textbf{53.8}              \\
                                                                          &                                                 & BLEURT-20                                   & -                                                                     & 59.0                                                                            & 53.2     & {\ul 67.7}        & \textbf{73.9}              \\ \cmidrule(l){2-8} 
                                                                          & \multirow{3}{*}{E2E}                            & ROUGE-2                                     & 29.2                                                                  & 30.3                                                                            & 30.9     & {\ul 33.9}        & \textbf{35.2}              \\
                                                                          &                                                 & ROUGE-L                                     & -                                                                     & 39.2                                                                            & 40.0     & {\ul 42.6}        & \textbf{43.9}              \\
                                                                          &                                                 & BLEURT-20                                   & -                                                                     & 64.5                                                                            & 65.0     & {\ul 68.1}        & \textbf{69.7}              \\ \midrule
\multicolumn{1}{l}{\multirow{3}{*}{Summary}}                              & \multicolumn{1}{c}{\multirow{3}{*}{WikiLingua}} & ROUGE-2                                     & 5.4                                                                   & 7.2                                                                             & 5.8      & \textbf{10.4}     & {\ul 9.9}                  \\
\multicolumn{1}{l}{}                                                      & \multicolumn{1}{c}{}                            & ROUGE-L                                     & -                                                                     & 18.9                                                                            & 16.4     & \textbf{23.4}     & {\ul 20.6}                 \\
\multicolumn{1}{l}{}                                                      & \multicolumn{1}{c}{}                            & BLEURT-20                                   & -                                                                     & 41.2                                                                            & 39.4     & {\ul 45.0}        & \textbf{47.7}              \\ \bottomrule[1.2pt]
\end{tabular}
\label{tab:nlg}
\vspace{-4mm}
\end{table}

\begin{tcolorbox}[left=0mm,right=0mm,top=0mm,bottom=0mm,boxsep=1mm,arc=0mm,boxrule=0pt, frame empty]
\textbf{(E2E Example, without demonstration sample)}
\small
\begin{lstlisting}
Aleksandr_Prudnikov , height , 185.0 (centimetres).
FC_Spartak_Moscow , ground , Otkrytiye_Arena.
Aleksandr_Prudnikov , club , FC_Spartak_Moscow. 
Verbalize:
\end{lstlisting}

\vspace{-0.1in}\tcbline

\textbf{Groundtruth:} 185 centimetre tall Aleksandr Prudnikov played for the Otkrytiye Arena based FC Spartak, Moscow.\vspace{0.07in}\\
\textbf{GPT-3 175B (Davinci):} Aleksandr Prudnikov is a midfielder for FC Spartak Moscow, a football (soccer) club based in Moscow, Russia.\vspace{0.07in}\\
\textbf{GLM-130B:} Aleksandr Prudnikov is 185.0 cm tall and plays for FC Spartak Moscow.

\end{tcolorbox}

\begin{tcolorbox}[left=0mm,right=0mm,top=0mm,bottom=0mm,boxsep=1mm,arc=0mm,boxrule=0pt, frame empty]
\textbf{(E2E Example, without demonstration sample)}
\small
\begin{lstlisting}
Combine all of the following data into a concise and grammatically correct text:
name : Blue Spice
eatType : coffee shop
area : riverside
\end{lstlisting}

\vspace{-0.1in}\tcbline

\textbf{Groundtruth:} At the riverside, there is a coffee shop called The Blue Spice.\vspace{0.07in}\\
\textbf{GPT-3 175B (Davinci):} Blue Spice is a riverside coffee shop which is located on the corner of River Street and Riverbank Street.\vspace{0.07in}\\
\textbf{GLM-130B:} There's a coffee shop that serves coffee in the riverside area, Blue Spice.

\end{tcolorbox}

\vspace{-3mm}

\begin{tcolorbox}[left=0mm,right=0mm,top=0mm,bottom=0mm,boxsep=1mm,arc=0mm,boxrule=0pt, frame empty]
\textbf{(WikiLingua Example, without demonstration sample)}
\small
\begin{lstlisting}
The majority of your customers will search for you online, so it's essential to have a user-friendly website. At the very least, your website should include information about your business and your history in the moving industry, details about the quoting process, contact information, and a description of the services you offer. If possible, allow customers to schedule quotes online, view your availability, or read testimonials from other customers. One of the easiest ways to start your business is by helping people you already know with their moves. You can be on the lookout for any announcements related to moving that your friends make on social media. Once you have provided good service to friends, they are likely to recommend you to others. In order to spread the word about your business, have some professional looking promotional materials printed and distribute them around your community.  You can distribute business cards at public events, tuck them into local bulletin boards, or even print them in directories, yearbooks, and other local print media. Flyers can be mailed, posted in public places, or distributed to businesses that might be able to refer customers to like you, such as furniture stores. Make sure you have a professional, recognizable logo that is consistent across all of your marketing materials. Another way to get your business's name out there is to make yourself visible. Whether it's by working with partners at local events, volunteering, or using your vehicle for an ad campaign, visibility is key for driving business. Build relationships with influential people in your community. Realtors are a great source of referrals to movers, as are the owners of local furniture stores or the office staff at a large apartment complex. You can use directory sites like Craigslist to advertise your services to people in your local community for free. Social media is also a great way to spread the word about your business. There are many options for advertising, depending on your budget and your target market. Consider options like PPC advertising, television and radio commercials, newspaper ads, direct mail flyers, or memberships with referral services. The best thing you can do to grow your business is to provide excellent service to your customers. Be sure to always be on time, be friendly, be respectful of your customers' belongings, and offer accurate price quotes.  Be sure to ask your happy customers to post reviews about you online. New customers will see these reviews and will have more faith in you as a reputable company if they see that others have had a good experience. You will inevitably end up having to deal with an unhappy customer at some point, but do your best to resolve the problem to the customer's satisfaction. The last thing you want is negative reviews about your business circulating the internet! Summarize the article: 
\end{lstlisting}

\vspace{-0.1in}\tcbline

\textbf{Groundtruth:} Create a website. Start with friends. Distribute business cards and flyers. Network for promotion. Seek out free advertising opportunities. Consider paid advertising. Gain a good reputation.\vspace{0.07in}\\
\textbf{GPT-3 175B (Davinci):} Market yourself online. Look for opportunities to market your services offline.\vspace{0.07in}\\
\textbf{GLM-130B:} Have a user-friendly website. Promote your business through social media. Build relationships with influential people in your community. Use directory sites like Craigslist to advertise your services.

\end{tcolorbox}

\begin{table}[H]
\caption{Winograd-style tasks evaluation (Winogender and Winograd273). All scores are accuracy. K refers to number of shots. $^*$PaLM 540B did not report the exact 0-shot Winogender result, so we have to estimate a value from its plotted diagram.}
\begin{tabular}{@{}lcccccccc@{}}
\toprule[1.2pt]
\multicolumn{1}{c}{}        & K & \begin{tabular}[c]{@{}c@{}}GPT-3\\ (Davinci)\end{tabular} & \begin{tabular}[c]{@{}c@{}}OPT\\ 175B\end{tabular} & \begin{tabular}[c]{@{}c@{}}BLOOM\\ 176B\end{tabular} & \begin{tabular}[c]{@{}c@{}}PaLM\\ 540B\end{tabular} & Chinchilla & \begin{tabular}[c]{@{}c@{}}Gopher\\ 280B\end{tabular} & GLM-130B \\ \midrule
\multirow{2}{*}{Winogender} & 0 & 64.2                                                      & 54.8                                               & 49.1                                                 & 75.0$^*$                                              & 78.3       & 71.4                                                  & 79.7     \\
                            & 1 & 62.6                                                      & -                                                  & 53.1                                                 & 79.4                                                & -          & -                                                     & 80.7     \\ \midrule
Winograd273                 & 0 & 88.3                                                      & 52.9                                               & 49.1                                                 & 90.1                                                & -          & -                                                     & 84.3     \\ \bottomrule[1.2pt]
\end{tabular}
\label{tab:winograd}
\end{table}
\begin{table}[H]
\centering
\caption{Closed-book question answering (Natural Questions, StrategyQA).}
\begin{tabular}{@{}lcccccc@{}}
\toprule[1.2pt]
\multicolumn{1}{c}{}   & \begin{tabular}[c]{@{}c@{}}GPT-3\\ (Davinci)\end{tabular} & \begin{tabular}[c]{@{}c@{}}BLOOM\\ 176B\end{tabular} & \begin{tabular}[c]{@{}c@{}}PaLM\\ 540B\end{tabular} & Chinchilla & \begin{tabular}[c]{@{}c@{}}Gopher\\ 280B\end{tabular} & GLM-130B \\ \midrule
Natural Questions (EM) & 14.6                                                      & 13.1                                                 & 21.2                                                & 16.6       & 10.1                                                  & 11.7     \\ \midrule
StrategyQA (Acc)       & 52.3                                                      & 49.8                                                 & 64.0                                                & -          & -                                                     & 60.6     \\ \bottomrule[1.2pt]
\end{tabular}
\label{tab:cbqa}
\end{table}
\begin{table}[H]
\centering
\caption{Commonsense reasoning (Commonsense QA, MC-TACO). K refers to number of shots.}
\begin{tabular}{@{}clcccc@{}}
\toprule[1.2pt]
                                      & K & GPT-3 (Davinci) & OPT 175B & BLOOM 176B & GLM-130B \\ \midrule
\multirow{2}{*}{Commonsense QA (Acc)} & 0 & 57.2            & -        & 42.8       & 61.6     \\
                                      & 1 & 61.2            & -        & -          & 62.2     \\ \midrule
\multicolumn{1}{l}{MC-TACO (EM)}      & 0 & -            & 12.4     & 13.1       & 13.6     \\ \bottomrule[1.2pt]
\end{tabular}
\label{tab:csqa}
\end{table}

\subsection{Winograd-Style Tasks}

We include the evaluation on Winograd-style tasks, which derives from the classical Winograd Schemas Challenge~\citep{levesque2012winograd} that aims to test coreference resolution in an ambiguous context for the machine to understand.
Since in MIP, we have included the Winogrande~\citep{sakaguchi2021winogrande} and SuperGLUE WSC~\citep{wangSuperGLUEStickierBenchmark2019}, here we test on Winogender~\citep{rudinger2018gender} and Winograd273~\citep{levesque2012winograd}.
For Winogender, GPT-3's results are acquired from OpenAI API, and BLOOM's 1-shot result is evaluated by ourselves.
For Winograd273, since existing works~\citep{brown2020language,chowdhery2022palm} show that 1-shot learning brings almost no improvement, we only test the zero-shot result.
Another thing to notice is that, despite GPT-style models (e.g., GPT-3, PaLM) adopting the ``partial evaluation'' described in~\citep{radford2019language}, we find the prompt ``\texttt{<sentence> The "<pronoun>" refers to [MASK]}'' is better for \glm and adopt it in the evaluation.

The results are presented in Table~\ref{tab:winograd}. \glm performs the best across all evaluated LLM on Winogender, and marginally poorer than GPT-3 and PaLM on Winograd273.

\subsection{Closed-book Question Answering}
Closed-book question answering (CBQA)~\citep{roberts2020much} is a widely adopted task to evaluate language models' memorization of factual knowledge, on contrary to the traditional ``open-book'' evaluation.
As we have included TriviaQA~\citep{joshi2017triviaqa} and WebQuestions~\citep{berant2013semantic} in the MIP training, here we choose Natural Questions~\citep{kwiatkowski2019natural} and StrategyQA~\citep{geva2021did} as the evaluation datasets for CBQA.

The results are presented in Table~\ref{tab:cbqa}.
\glm performs relatively poorer on Natural Questions and performs well on StrategyQA.
\glm's underperformance on Natural Questions, we speculate, potentially derives from the insufficiency fitting on English corpora, as it roughly only viewed 200B English tokens and thus does not memorize the detailed knowledge very well.
Since CBQA seems to be a task that especially stresses memorization, as is indicated by Chinchilla~\citep{hoffmann2022training}'s a strong performance, we think with sufficient training later, \glm can perform better.

\subsection{Commonsense Reasoning}
Here we evaluate \glm and some other LLMs on commonsense reasoning abilities.
As we have included PIQA~\citep{bisk2020piqa}, ARC~\citep{clark2018think}, and OpenbookQA~\citep{mihaylov2018can} in the MIP training, we select another two widely adopted commonsense reasoning datasets in our evaluation: Commonsense QA~\citep{talmor2019commonsenseqa} and Multiple-choice Temporal Commonsense (MC-TACO, \cite{zhou2019going}).
For Commonsense QA, we test the GPT-3 via OpenAI Davinci API, BLOOM-176B via its Huggingface Implementation, and GLM-130B using the prompt ``answer\_given\_question\_without\_options'' from promptsource~\citep{bach2022promptsource}.
For StrategyQA, we follow the EM computation method provided in~\citep{zhou2019going}.

The results are shown in Table~\ref{tab:csqa}.
As we can see, \glm performs the best on both Commonsense QA and MC-TACO across evaluated LLMs, demonstrating that \glm has a good grasp of commonsense knowledge.
OPT's results are not included due to the reason described in Appendix~\ref{app:opt_sources}.

\subsection{Fixed Label Datasets: A Case Study in Natural Language Inference}

As is discussed in Section~\ref{sec:results}, we adopt a rather strict criterion for selecting datasets for zero/few-shot learning in \glm's evaluation due to the use of MIP.
Nevertheless, the criterion significantly reduces the dataset we could currently evaluate, and especially some readers have doubted whether the restriction of not evaluating on MIP-seen fixed-label datasets is necessary (e.g., natural language inference (NLI)), and suggest that we may report them in an independent section to avoid confusion.

Frankly speaking, in such a setting GLM-130B’s zero/few-shot learning could be quite advantageous. Below, we take NLI as a typical example to show GLM-130B’s outperformance in the scenarios.
We include 6 widely-used NLI datasets--which are not incorporated in \glm's MIP training, as the benchmarks. 
The results are presented in Table~\ref{tab:nli}, which shows that GLM-130B’s “zero-shot” performance could be much better due to the seen task type.

\begin{table}[H]
\centering
\caption{``Zero-shot'' results of GLM-130B on 6 typical natural language inference (NLI) datasets. \textbf{$^*$DISCLAIMER: Despite the datasets are never seen, some other NLI datasets have been included in GLM-130B's MIP, making it different from the existing standard zero-shot setting.}}
\begin{tabular}{@{}lccc@{}}
\toprule[1.2pt]
                                               & BLOOM 176B & OPT 175B & GLM-130B$^*$ \\ \midrule
qnli (valid, median of 5 prompts)              & 50.9       & 55.4     & 86.7     \\
mnli (valid, median of 15 prompts)             & 35.5       & 36.0     & 85.7     \\
mnli\_mismatched (valid, median of 15 prompts) & 35.5       & 36.0     & 84.6     \\
wnli (valid, median of 5 prompts)              & 57.7       & 53.5     & 67.6     \\
glue/cola (valid, median of 5 prompts)         & 39.0       & 44.4     & 57.6     \\
glue/mrpc (valid, median of 5 prompts)         & 31.6       & 44.6     & 87.3     \\ \bottomrule[1.2pt]
\end{tabular}
\label{tab:nli}
\end{table}

\subsection{SuperGLUE}
\begin{figure}[t]
    \centering
    \includegraphics[width=\linewidth]{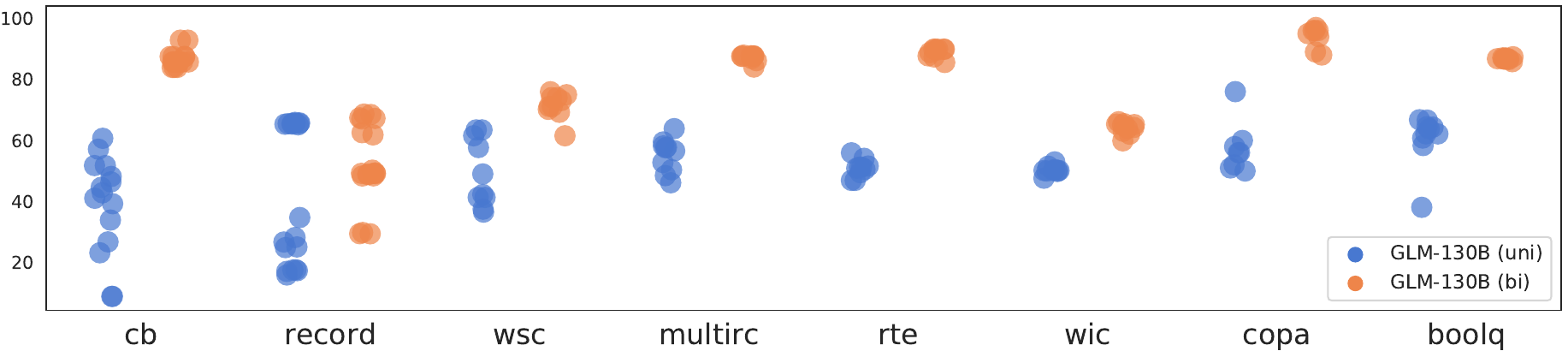}
    \caption{GLM-130B (uni and bi)'s untuned results on SuperGLUE development set, using promptsource~\citep{bach2022promptsource} prompts and task formulation. 
    \textbf{DISCLAIMER: Noted that some of the SuperGLUE training sets have been included in the MIP training. We report the results here only for readers' reference.}}
    \label{fig:superglue}
\end{figure}

We also report our evaluation of \glm on the SuperGLUE~\citep{wangSuperGLUEStickierBenchmark2019} benchmark, which consists 8 different natural language understanding challenges.
Noted that these results are neither zero/few-shot nor fine-tuned results, because 7 out of 8 tasks' training sets have been included in \glm's MIP training (except for ReCoRD) together with other 67 multi-task datasets; however, \glm is also not individually fine-tuned on any of them.
Therefore, these results are not for relative comparison for any other models', but only for readers' reference on \glm's absolute ability.

\begin{table}[H]
\centering
\begin{tabular}{lllllllll}
\toprule
         & BoolQ & CB    & COPA & MultiRC & ReCoRD & RTE   & WiC   & WSC  \\
\midrule
GLM-130B & 89.69 & 98.21 & 100  & 89.32   & 92.11  & 94.22 & 76.96 & 88.5 \\
\bottomrule
\end{tabular}
\caption{The results of \glm on the SuperGLUE dataset obtained using the P-tuning v2~\citep{liu2022p}. We report the Accuracy metric for all datasets except for MultiRC (F1a) and ReCoRD (F1).}
\label{tab:superglue-p-tuning-v2}
\end{table}

The results are presented in Figure~\ref{fig:superglue}.
We ablate the unidirectional and bidirectional \glm to justify the usefulness of GLM objective in boosting LLMs' ability to understand.
Each point in the figure refers to a prompt-specific result, for which the prompt is from the promptsource~\citep{bach2022promptsource} repository.
We adopt the task formulation from promptsource, too.
As we can observe, GLM (bi) has much fewer variances and higher performances on all tasks.
For some of the tasks (such as CB, MultiRC, RTE, COPA, and BoolQ), \glm can even achieve over 80\% accuracy.

We also attempted to fine-tune \glm on the SuperGLUE dataset. However, we encountered the issue of rapid overfitting within a single epoch when we used full parameter fine-tuning on downstream tasks. This resulted in poor performance on the validation set. To address this issue, we explored the use of efficient parameter fine-tuning methods, which tune only a small number of parameters and are less prone to overfitting. After experimenting with several methods, we use P-Tuning v2~\citep{liu2022p}, which demonstrated comparable results to full parameter fine-tuning in GLM-130B, but with only 0.1\% to 3\% of tuned parameters. The results of our experiments with P-Tuning v2 are presented in Table~\ref{tab:superglue-p-tuning-v2}.

\subsection{Chain-of-Thought Prompting}
\begin{figure}[t]
    \centering
    \includegraphics[width=0.8\linewidth]{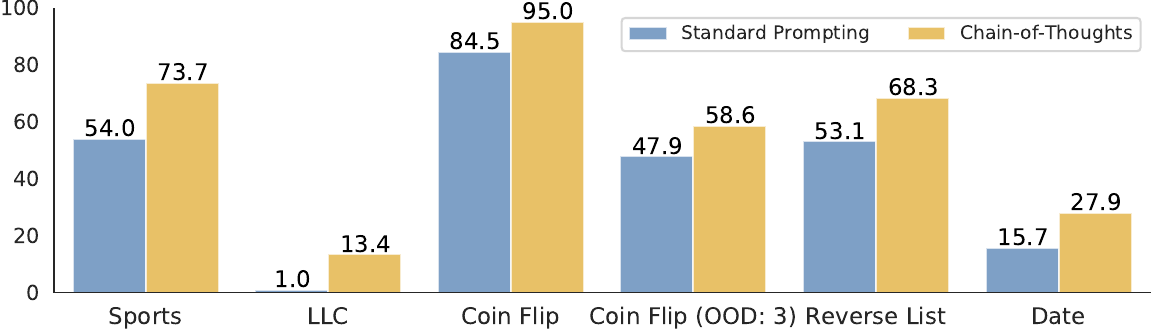}
    \caption{Chain-of-thought prompting can also improve \glm's performance on reasoning tasks compared to standard prompting.}
    \label{fig:CoT}
\end{figure}

We evaluate the chain-of-thought prompting performance on \textbf{Last letter concatenation} (LLC), \textbf{Coin Flip}, \textbf{Reverse List}, and two tasks from BIG-bench~\cite{srivastava2022beyond} \textbf{Sports} understanding, and \textbf{Date} understanding, following the setting in \cite{wei2022chain}. The results are shown in Figure~\ref{fig:superglue}. We find that chain-of-thought prompting can improve GLM-130B's performance on symbolic reasoning and commonsense reasoning.

\textbf{Last letter concatenation (LLC)}. The task asks the model to concatenate the last letters of words in a name (e.g., "Elon Musk" -> "nk"). We generate full names by randomly concatenating the top 1000 first and last names from name census data\footnote{\url{https://namecensus.com}}.

\textbf{Coin flip}. This task asks the model to answer whether a coin is still heads up after people either flip or don't flip it beginning from being heads up. (e.g., "A coin is heads up. Phoebe flips the coin. Osvaldo does not ﬂip the coin. Is the coin still heads up?" -> "no"). We additionally evaluate on the scenario where the number of people in the query examples is larger than that in the in-context examples, i.e. the out-of-distribution (OOD) setting.

\textbf{Reverse List}. This task asks the model to reverse the order of a list of everyday objects (e.g., "cigar, umbrella, key, gum, alarm" -> "alarm, gum, key, umbrella, cigar"). We generate the lists by randomly sampling from the vocabulary of everyday objects\footnote{\url{https://www.vocabulary.com/lists/189583}}.

\textbf{Sports}. This task asks the model to judge the truthfulness of a statement about a sports player (e.g., "Joao Moutinho caught the screen pass in the NFC championship" -> "false").

\textbf{Date}. This task asks the model to infer the data from a given context (e.g., "2015 is coming in 36 hours. What is the date one week from today in MM/DD/YYYY?" -> "01/05/2015").

We use the same examples and chains as \cite{wei2022chain}. For each task, we try two different formats of prompts and both unidirectional and bidirectional attention mechanism and report the best performance. The first format is "Question: \{context\} Answer: \{target\}". The second one is to add serial numbers before examples in the first format of prompts.
The results are presented in Figure~\ref{fig:CoT}.

\begin{figure}[t]
    \centering
    \includegraphics[width=\linewidth]{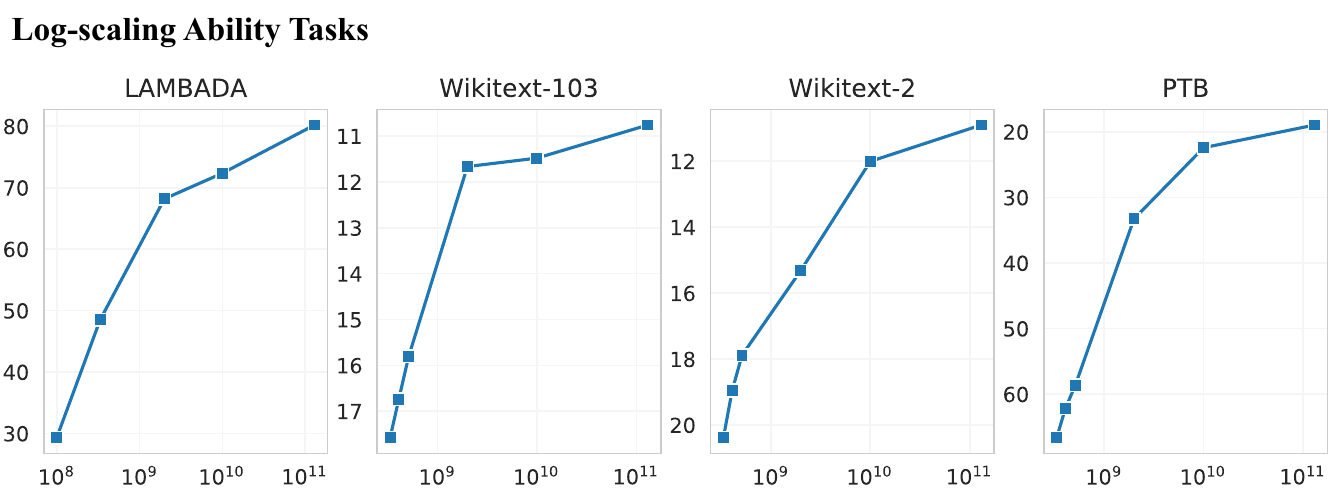}
    \caption{Log-scaling ability tasks of \glm. These tasks' performance grows logarithmically with the amount of GLM parameters. Most of traditional NLP tasks fall into the same pattern.}
    \label{fig:log_scaling}
\end{figure}

\begin{figure}[t]
    \centering
    \includegraphics[width=\linewidth]{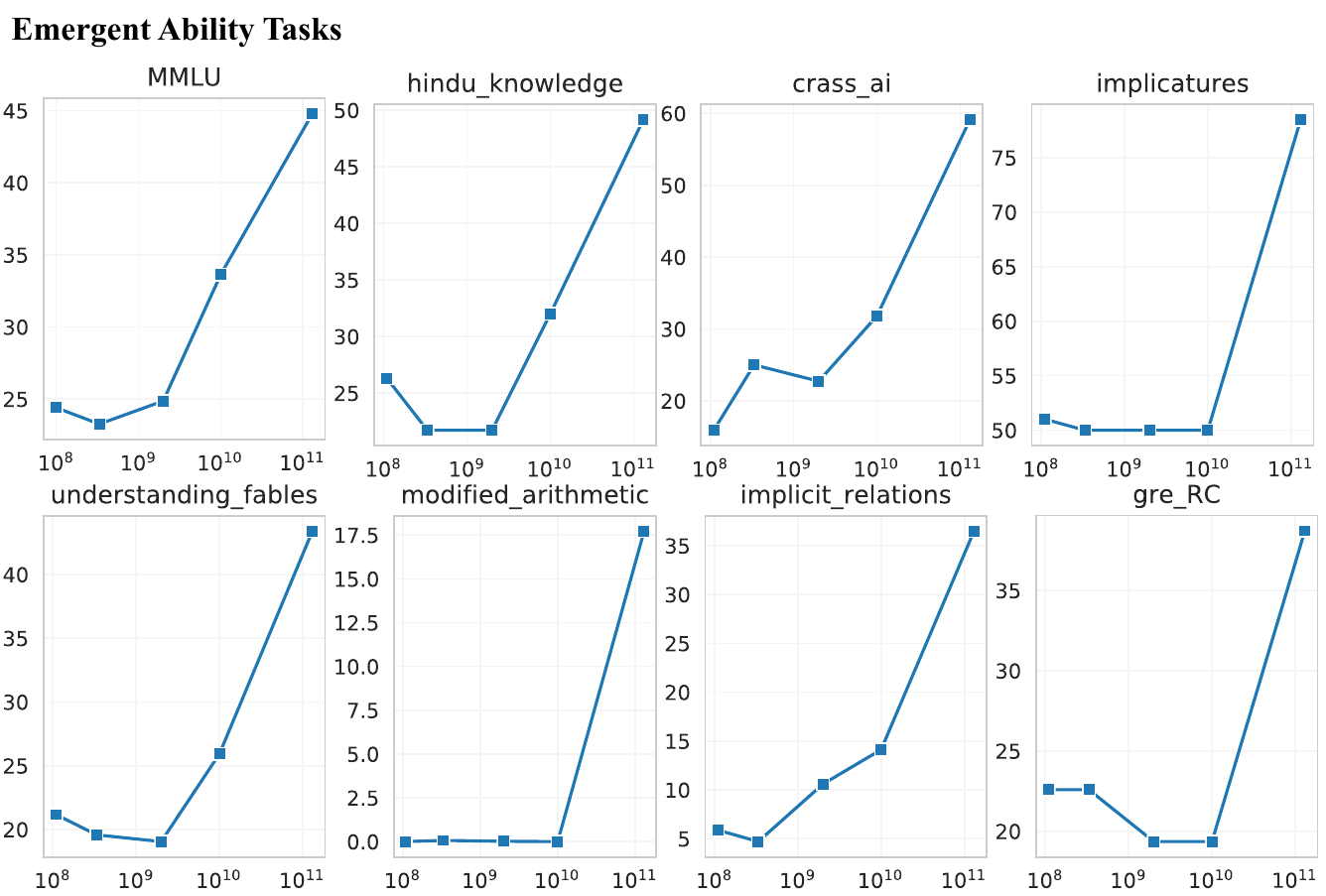}
    \caption{Emergent ability tasks of \glm. These tasks' performance does not grow much until the model size reaches a certain threshold (e.g., 100B or 10B). After reaching the threshold, the model performance soars up quickly. The BIG-bench~\citep{srivastava2022beyond} benchmark collects many of these challenges.}
    \label{fig:emergent}
\end{figure}

\section{Scaling and Emergent Abilities in GLM-130B}

Scaling up pre-trained language models has been proven to boost downstream performance on a wide range of tasks continually. His, emergent abilities which are unpredictable from smaller scales. To illustrate this, we conducted extensive experiments to explore the scaling property and emergent abilities. Following prior literature~\citep{wei2022emergent}, we categorize the NLP tasks into two types based on our observations.

\begin{itemize}[leftmargin=*,itemsep=0pt,parsep=0.2em,topsep=0.0em,partopsep=0.0em]
    \item \textbf{Log-scaling Ability Tasks (Cf. Figure~\ref{fig:log_scaling})}: where the task performance grows logarithmically with the number of model parameters. Typical tasks and datasets include LAMBADA, Wikitext-103, Wikitext-2, Penn Tree Bank.
    \item \textbf{Emergent Ability Tasks (Cf. Figure~\ref{fig:emergent})}: where the task performance only soars up when the amount of model parameters reaches a certain threshold. Typical tasks and datasets include: MMLU, hindu\_knowledge, crass\_ai, implicatures, understanding\_fables, modified\_arithmetic, implicit\_relations, and gre\_reading\_comprehension from BIG-bench~\citep{srivastava2022beyond}.
\end{itemize}

In line with the observation in~\citep{wei2022emergent}, we show that \glm also presents the two similar scaling behaviors to other LLMs such as GPT-3, LaMDA, and PaLM.
Though why and how LLMs present these intriguing properties remain unclear, \glm provides open opportunities for all researchers to test and understand the reason behind them.

\clearpage
\clearpage
\section{Contributions} \label{app:contribution}


The GLM-130B project was conceived in Dec. 2021 with its pre-training part completed in July 3rd, 2022 and its evaluation and applications still ongoing. 
Over the course, we have experienced various technical and engineering challenges (Cf. Appendix \ref{sec:appendix-history} and Figure~\ref{fig:timeline} for details). 
It would not be possible to reach its current status if without the collaboration of multiple teams---the Knowledge Engineering Group (KEG), Parallel Architecture \& Compiler technology of Mobile, Accelerated, and Networked systems Group (PACMAN), and Natural Language Processing Group (THUNLP) at Tsinghua University, 
as well as Zhipu.AI.
The detailed contributions are listed below. 

\subsection{Preparation}

\begin{itemize}[leftmargin=*,itemsep=0pt,parsep=0.4em,topsep=0.0em,partopsep=0.0em]
\item \textbf{Model Implementation:} Aohan Zeng, Zhengxiao Du
\item \textbf{Self-Supervised Data Processing:} Ming Ding, Wendi Zheng
\item \textbf{Multitask Data Processing:} Xiao Liu, Xiao Xia
\item \textbf{Model Architecture:} Aohan Zeng, Xiao Liu, Zhengxiao Du, Hanyu Lai
\item \textbf{Training Stability:} Aohan Zeng, Xiao Liu, Ming Ding
\item \textbf{3D-Parallelism and Training Efficiency:} Aohan Zeng, Zixuan Ma, Jiaao He, Zhenbo Sun
\end{itemize}

\subsection{Model Training}

\begin{itemize}[leftmargin=*,itemsep=0pt,parsep=0.4em,topsep=0.0em,partopsep=0.0em]
\item \textbf{Large-Scale Training \& Monitoring:} Aohan Zeng, Xiao Liu
\item \textbf{Model Performance Validation:} Aohan Zeng
\end{itemize}

\subsection{Post Training}

\begin{itemize}[leftmargin=*,itemsep=0pt,parsep=0.4em,topsep=0.0em,partopsep=0.0em]
\item \textbf{Evaluation Framework:} Aohan Zeng, Zhengxiao Du
\item \textbf{Language Modeling Evaluation:} Aohan Zeng
\item \textbf{MMLU \& BIG-Bench Evaluation:} Aohan Zeng
\item \textbf{CLUE  \&  FewCLUE Evaluation:} Xiao Liu, Aohan Zeng
\item \textbf{Ethical Evaluation:} Yifan Xu, Aohan Zeng, Xiao Liu, Zihan Wang
\item \textbf{Baseline Evaluation:} Xiao Liu, Jifan Yu, Weng Lam Tam
\item \textbf{INT4 Quantization:} Aohan Zeng, Zihan Wang, Xiao Liu, Hanyu Lai
\item \textbf{Inference Acceleration:} Zihan Wang, Aohan Zeng
\item \textbf{Low-Resource Inference:} Gouyang Zeng, Xu Han, Weilin Zhao, Zhiyuan Liu
\item \textbf{Demo and API:} Hanyu Lai, Jifan Yu, Xiaohan Zhang, Yufei Xue, Shan Wang, Jiecai Shan, Haohan Jiang, Zhengang Guo
\item \textbf{Manuscript Writing:} Xiao Liu, Yuxiao Dong, and Jie Tang wrote the main paper, and Xiao Liu, Aohan Zeng, and Zhengxiao Du wrote the Appendix. 
\end{itemize}


\subsection{Project Management}

\begin{itemize}[leftmargin=*,itemsep=0pt,parsep=0.4em,topsep=0.0em,partopsep=0.0em]
\item \textbf{Student Leaders:} Aohan Zeng, Xiao Liu
\item \textbf{Technical Advisors:} Yuxiao Dong, Jidong Zhai, Wenguang Chen, Zhiyuan Liu, Peng Zhang, Jie Tang
\item \textbf{Project Leader:} Jie Tang
\end{itemize}
 
 \subsection{Computation Sponsor}
 \begin{itemize}[leftmargin=*,itemsep=0pt,parsep=0.4em,topsep=0.0em,partopsep=0.0em]
\item \textbf{GPU Sponsor:} Zhipu.AI

 \end{itemize}
 
\clearpage

\section{A Brief History  of \glm}
\label{sec:appendix-history}

\begin{figure}[hbtp]
    \vspace{-10mm}
    \centering
    \includegraphics[width=\textwidth]{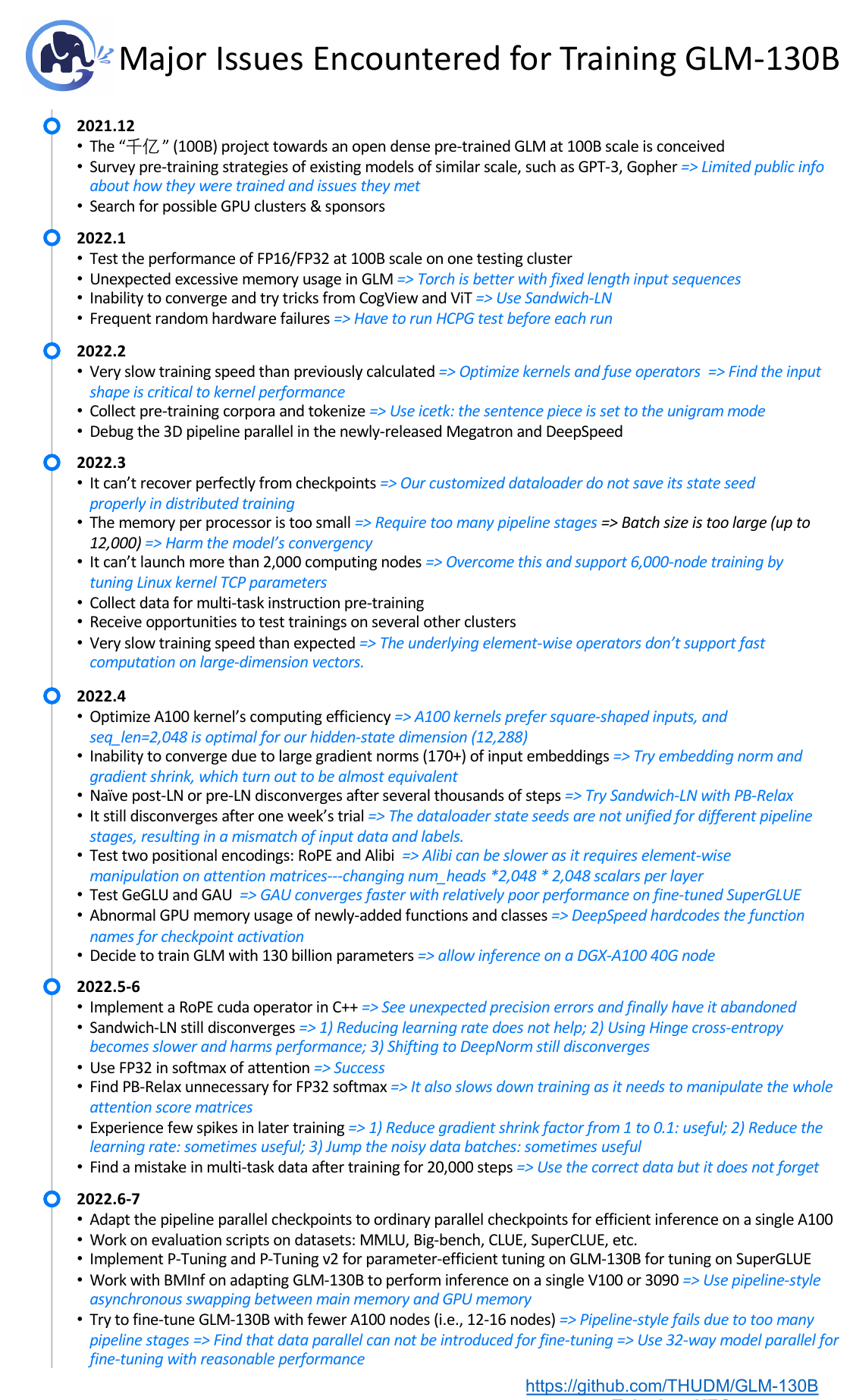}
    \caption{The timeline of major issues that training \glm encountered and addressed, as of July 31st, 2022.}
    \label{fig:timeline}
\end{figure}

The \glm project\footnote{This section is largely extracted and updated from the blog introduction of \glm at  \url{http://keg.cs.tsinghua.edu.cn/glm-130b/} (Posted date: August 4, 2022).}
was conceived in Dec. 2021 in a brainstorming meeting at Tsinghua KEG. 
We firmly believe that it is of value to pre-train a highly accurate language model, in particular for both Chinese and English. 
Though GPT-3~\citep{brown2020language} is the pioneer for this effort, it is not available to most people in the world. 
In addition, it supports English only. 
We therefore decide to initialize the project \glm.
Please note that the WuDao 1.75T model we built last year is a sparse model with 480 mixture-of-experts (MoE), rather than a dense one as GPT-3. 
Our goal then is to train a bilingual pre-trained dense model with high accuracy on downstream tasks, and to make it open to everyone in the world-anyone, anywhere can download it and use it on a single server with appropriate GPUs.

The ambitious project soon faced several important challenges:

\begin{itemize}[leftmargin=*,itemsep=0pt,parsep=0.2em,topsep=0.0em,partopsep=0.0em]
    \item \textbf{Lack of computational resources}: No organization is willing to sponsor such a big project and freely make it public.
    \item \textbf{Lack of a robust pre-training algorithm}: Despite GPT-3’s success on English corpus, it is unclear how to train a high-accurate bilingual model for both English and Chinese.
    \item \textbf{Lack of fast inference solutions}: Since the goal is to have the model public to everyone, we need to design fast inference solutions with low resource requirements to run the model.
\end{itemize}

For the pre-training algorithm, we finally chose GLM \citep{du2022glm} due to its high performance in practice. 
We eventually decided to train a GLM model of 130 billion parameters after several rounds of discussions and exploration, because such a size makes it possible to run the inference on a single A100 (40G * 8) server.

Our first attempt at training the model was in January 2022, shortly after we received a small sponsor of GPUs for test running. 
However, we soon realized that we had significantly underestimated the technical difficulties of pre-training a model at such a scale (>100B). 
It seems that pre-training a highly accurate 100B-scale model is quite different from training a 10B-scale one. Due to frequent random hardware failures, model gradients exploding, unexpected excessive memory usage in the algorithm, debug for the 3D pipeline in the new Megatron and DeepSpeed frameworks, inability to recover from optimizer states, blocked TCP responses between processes, and many many unexpected ``bugs'', the project was delayed for many times. 
The Tsinghua PACMAN team gave us a hand at this difficult time and together we successfully fixed most of the ``bugs''.

By March, we were still short on computational resources, but fortunately got a chance to try test runs on several other platforms, including Ascend 910, Hygon DCU, NVIDIA, and Sunway. 
The immediate challenge was for us to adapt our training code to these different platforms, as the underlying operators are quite different. 
Also, it introduced many new issues: the element-wise operators not supporting fast computation for large-dimension vectors, various issues that hindered convergence—the large gradient norms of input embeddings, native Post-LN, Pre-LN, and Sandwich-LN, dataloader state seeds, and computation precision choices in Softmax and Attention — as well as numerous mistakes we ourselves made. 
With tremendous help from all of our generous partners, we finally succeeded in making our pre-training algorithms runnable across all the platforms—frankly, a surprising achievement for this project. 
The timeline of \glm in Figure~\ref{fig:timeline} covers most of the issues we have encountered and addressed as of this writing.

On April 26th, we received a generous computing sponsorship from Zhipu.AI — an AI startup that aims to teach machines to think like humans. 
After another week of testing, we finally kicked off the training of the \glm model on its 96 A100 (40G * 8) servers on May 6th. 
Additionally, Zhipu.AI also sent a team to help evaluate the pre-trained model and build a demonstration website.

The training period spanned two months, during which we began developing a toolkit to allow \glm's inference in low-resource setting with swapping technique and quantization. 
Though it is already the most accessible model of its scale, 
together with our partner from Tsinghua NLP, 
we have been exploring the limit of popularized hardware platforms, which would truly make the 100B-scale model accessible to as many people as possible. 
To date, we managed to reach the INT4 weight quantization for \glm.  
Importantly, the INT4 version of \glm without post training faces negligible performance degradation compared to its uncompressed original, while it consumes only 25\% of the GPU memory  required by the uncompressed version, thus supporting its effective inference on 4 $\times$ RTX 3090 Ti (24G) or 8 $\times$ RTX 2080 Ti (11G). 
We will attempt to further reduce the resource requirements and keep the community updated on this important working item.
\section{Broader Impact}
This paper introduces an open bilingual pre-trained language model with 130 billion parameters. Currently most pre-trained language models with over 100 billion parameters are privately owned by governments and large corporations~\citep{brown2020language,thoppilan2022lamda,rae2021scaling,chowdhery2022palm,wang2021ernie}. A few of them~\citep{brown2020language,lieber2021jurassic} provide limited inference APIs with fees. In contrast, the weights and code of \glm are open to anyone who is interested in LLMs. Moreover, we significantly lower the hardware requirements for inference by speed-up implementation and INT4 quantization. The paper can have a broader impact on the research community, individual developers and small companies, and society.

\subsection{Impact on AI Research}
Most research institutions cannot afford the substantial cost of pretraining large language models. As a result, most researchers, except employees of governments and large corporations, only have access to the limited inference APIs with fees. With the inference APIs, researchers can only analyze the outputs of models as black boxes, which limits the scope of potential work. With \glm, researchers can analyze the model parameters and internal states corresponding to specific inputs, leading to in-depth studies of LLMs' theory, capacity, and flaws. Researchers can also modify the model architecture and weights, to validate the proposed algorithms to improve LLMs~\cite{Zhu2020Modifying,Cao2021Editing,Hase2021Belief,Mitchell2022Editing}.

With INT4 quantization, \glm can perform inference on popularized GPUs such as 4 $\times$ RTX 3090 or 8 $\times$ RTX 2080 Ti, which can be easily accessed from cloud service. As a result, researchers who cannot afford powerful data-center GPU servers like DGX-A100 can also utilize \glm.

\subsection{Impact on Individual Developers and Small Companies}
Currently, individual developers and small companies who want to integrate LLMs into their business can only choose paid inference APIs. The increased cost can hinder their attempts. Instead, \glm can be deployed on popularized hardware that they own or can access via cloud service to reduce the cost. Furthermore, they can utilize distillation techniques~\cite{sanh2019distilbert,jiao2020tinybert} to obtain smaller models that preserve comparable performance on their specific tasks. While some developers may lack the ability to complete deployment and distillation on their own, we believe with \glm and more open LLMs in the future, the corresponding toolkits and service providers will become more available.

We also note that currently most applications of LLMs are based on prompt engineering, partly due to the limitation of inference APIs. In downstream scenarios such as online customer service, the companies accumulate huge amounts of human-generated data that contain domain knowledge. With the open-source weights and code, developers can finetune \glm on their own data to mitigate the gap of domain knowledge.

\subsection{Social Impact}
Large language models, together with other machine learning models in different modalities (e.g., Image~\citep{ramesh2021zero,ding2021cogview,sahariaphotorealistic} and Video~\citep{hong2022cogvideo}), could be used to generate synthetic text for harmful applications, such as telemarketing fraud, political propaganda, and personal harassment as is discussed in~\citep{weidinger2021ethical,Sheng2021SocietalBI,Dev2021HarmsOG}. 
We do not anticipate any hazardous outputs, especially towards vulnerable and historically disadvantaged groups of people, after using the model.

While some people think that restricting access to LLMs can prevent such harmful applications, we argue that promoting LLM inclusivity can lead to better defense against potential harm caused by LLMs. 
Currently, only governments and large corporations can afford the considerable costs of pre-training LLMs. 
There is no guarantee that organizations having the substantial financial resources to pretrain an LLM will not do harm with it. 
Without access to such LLMs, individuals cannot even realize the role of LLMs in harm. Conversely, releasing an open LLM can provide access and transparency to all the researchers and promote the research to reduce the potential harm of LLMs, like algorithms to identify the synthetic text~\cite{gehrmann2019gltr} or detect fake news~\cite{Li2021ExploringTI}. 

Also, it is known that LLMs can suffer from problems in fairness, bias, privacy, and truthfulness~\cite{abs-2112-12938,lin2022truthfulqa,Liang2021SocialBias,Bender2021Danger}.
An open LLM can reveal the model parameters and internal states corresponding to specific inputs instead of providing APIs to black-box models. 
In conclusion, researchers can conduct analysis of LLMs' flaws in depth and propose improved algorithms to solve the problems. 

\section{Environmental Impact}
One of the major concerns about large language models is their huge energy usage and associated carbon emissions~\cite{Strubell2019Energy,Lacoste2019Carbon,Patterson2021Carbon,Bender2021Danger}. GPT-3 was estimated to use 500 tons of carbon emissions footprint (CO2eq)~\cite{Patterson2021Carbon}. We consumed a total of 442.4MWh of electricity over the 60-day course of training. Given the 0.5810 kg/kWh carbon efficiency of local power grid, the pre-training released 257.01 metric tons of CO$_2$. This is around half of GPT-3's carbon footprint, probably due to the efficient parallel strategies and NVIDIA's hardware improvements. The carbon emission is roughly the equivalent of the yearly emissions of 18 average Americans. However, we believe that with \glm released, more carbon emissions for reproducing 100B-scale LLMs can be saved.

\end{document}